\documentclass[10pt,twocolumn,letterpaper]{article}

\usepackage{iccv}
\usepackage{times}
\usepackage{epsfig}
\usepackage{graphicx}
\usepackage{amsmath}
\usepackage{amssymb}
\usepackage{booktabs}

\usepackage{tabularx}
\usepackage{stmaryrd}
\usepackage{multirow}
\usepackage[table,xcdraw,dvipsnames]{xcolor}
\usepackage{enumitem}
\usepackage{pifont}
\usepackage{bm}
\usepackage[ruled,vlined]{algorithm2e}
\usepackage{wrapfig}
\usepackage{float}
\usepackage{graphbox} 
\makeatletter
\@namedef{ver@everyshi.sty}{}
\makeatother
\usepackage{tikz}
\usepackage{adjustbox}
\usepackage{nicefrac}
\usepackage[nocompress]{cite}

\usepackage[pagebackref,breaklinks,colorlinks,citecolor=NavyBlue,hypertexnames=false]{hyperref}

\iccvfinalcopy %

\makeatletter
\DeclareRobustCommand\onedot{\futurelet\@let@token\@onedot}
\def\@onedot{\ifx\@let@token.\else.\null\fi\xspace}
\def\eg{\emph{e.g}\onedot} 
\def\ie{\emph{i.e}\onedot} 
\def\RR{\mathbb{R}}

\newcommand{\stimes}{\hspace{-0.15em}\times\hspace{-0.15em}}
\newcolumntype{C}[1]{>{\centering\arraybackslash}p{#1}}	
\newcolumntype{M}[1]{>{\centering\arraybackslash}m{#1}}
\newcolumntype{L}[1]{>{\arraybackslash}p{#1}}
\newcommand{\citesupp}{\cite}

\makeatletter
\renewcommand{\paragraph}{%
  \@startsection{paragraph}{4}%
  {\z@}{1ex \@plus 1ex \@minus .2ex}{-0.5em}%
  {\normalfont\normalsize\bfseries}%
}
\makeatother

\newcommand{\myparagraph}[1]{\paragraph{#1}}
\newcommand{\myquote}[1]{``\emph{#1}''}
\newlist{todolist}{itemize}{2}
\setlist[todolist]{label=$\square$}

\newcommand{\app}[1]{Appendix~\ref{#1}} %
\newcommand{\proj}{project webpage} %

\newcommand{\remove}[1]{\ignorespaces}
\newcommand{\new}[1]{\textcolor{black}{#1}}
\newcommand{\blue}[1]{\textcolor{black}{#1}}
\newcommand{\change}[1]{\textcolor{black}{#1}}
\newcommand{\revision}[1]{\textcolor{black}{#1}}
\newcommand{\corgrid}[1]{\textcolor{black}{#1}}

\newcommand{\croppedgraphicstaichi}[1]{\begin{tikzpicture} \draw (0, 0) node[inner sep=0] {\adjincludegraphics[width=0.0975\linewidth,trim={{.\width} {.\height} {.\width} {.\height}},clip]{#1}}; \draw (-0.3, -0.15) node[opacity=0.75]{}; \end{tikzpicture}}
\newcommand{\croppedgraphicsucf}[1]{\begin{tikzpicture} \draw (0, 0) node[inner sep=0] {\adjincludegraphics[width=0.097\linewidth,trim={{.15\width} {.05\height} {.05\width} {.15\height}},clip]{#1}}; \draw (-0.3, -0.15) node[opacity=0.75]{}; \end{tikzpicture}}
\newcommand{\croppedgraphicsucfz}[1]{\begin{tikzpicture} \draw (0, 0) node[inner sep=0] {\adjincludegraphics[width=0.097\linewidth,trim={{.3\width} {.2\height} {.1\width} {.2\height}},clip]{#1}}; \draw (-0.3, -0.15) node[opacity=0.75]{}; \end{tikzpicture}}
\newcommand{\croppedgraphichmth}[1]{\begin{tikzpicture} \draw (0, 0) node[inner sep=0] {\adjincludegraphics[width=0.097\linewidth,trim={{.35\width} {.2\height} {.0\width} {.15\height}},clip]{#1}}; \draw (-0.3, -0.15) node[opacity=0.75]{}; \end{tikzpicture}}
\newcommand{\croppedgraphichmtw}[1]{\begin{tikzpicture} \draw (0, 0) node[inner sep=0] {\adjincludegraphics[width=0.097\linewidth,trim={{.2\width} {.2\height} {.2\width} {.2\height}},clip]{#1}}; \draw (-0.3, -0.15) node[opacity=0.75]{}; \end{tikzpicture}}
\newcommand{\croppedgraphichmon}[1]{\begin{tikzpicture} \draw (0, 0) node[inner sep=0] {\adjincludegraphics[width=0.097\linewidth,trim={{.15\width} {.2\height} {.15\width} {.1\height}},clip]{#1}}; \draw (-0.3, -0.15) node[opacity=0.75]{}; \end{tikzpicture}}
\newcommand{\croppedinp}[1]{\begin{tikzpicture} \draw (0, 0) node[inner sep=0] {\adjincludegraphics[width=0.45\linewidth,trim={{.\width} {.2\height} {.\width} {.\height}},clip]{#1}}; \draw (-0.3, -0.15) node[opacity=0.75]{}; \end{tikzpicture}}

\def\iccvPaperID{6972} %

\begin{document}

\title{WALDO: Future Video Synthesis using Object Layer Decomposition and Parametric Flow Prediction}

\author{Guillaume Le Moing\textsuperscript{1, 2, }\thanks{corresponding author: guillaume.le-moing@inria.fr} \hspace{0.8cm} Jean Ponce\textsuperscript{2, 3} \hspace{0.8cm} Cordelia Schmid\textsuperscript{1, 2}\\[0.5em]
\begin{tabular}{@{}cccc@{}}
\textsuperscript{1}Inria & ~~~~~~\textsuperscript{2}D\'epartement d’informatique de~~~~~~ & \textsuperscript{3}Center for Data Science\\
& l’ENS (CNRS, ENS-PSL, Inria) & New York University\\[-0.3em]
\end{tabular}}

\maketitle

\begin{abstract}
   This paper presents WALDO (WArping Layer-Decomposed Objects), a novel approach to the prediction of future video frames from past ones. Individual images are decomposed into multiple layers combining object masks and a small set of control points. The layer structure is shared across all frames in each video to build dense inter-frame connections. Complex scene motions are modeled by combining parametric geometric transformations associated with individual layers, and video synthesis is broken down into discovering the layers associated with past frames, predicting the corresponding transformations for upcoming ones and warping the associated object regions accordingly, and filling in the remaining image parts. \blue{Extensive experiments on multiple benchmarks including urban videos (Cityscapes and KITTI) and videos featuring nonrigid motions (UCF-Sports and H3.6M), show that our method consistently outperforms the state of the art by a significant margin in every case. Code, pretrained models, and video samples synthesized by our approach can be found in the project webpage.\footnote{url: \url{https://16lemoing.github.io/waldo}}} %
\end{abstract}

\section{Introduction}
\label{sec:intro}

Predicting the future from a video stream is an important tool to make autonomous agents more robust and safe. %
In this paper, we are interested in the case where future frames are synthesized from a fixed number of past ones. 
One possibility is to build on \revision{advanced} image synthesis models~\cite{menick2019generating, esser2020taming, karras2021alias, dhariwal2021diffusion} %
and adapt them to predict new frames conditioned on past ones~\cite{tian2021good}. Extending these already memory- and compute-intensive methods to our task may, however, lead to prohibitive costs due to the extra temporal dimension.  Hence, the resolution of videos predicted by this approach is often limited~\cite{weissenborn2020scaling, ho2022video, wu2021greedy}. %
Other works resort to compression~\cite{vandenoord2017neural, nash2021generating} to reduce computations~\cite{yan2021videogpt, wu2021nuwa, nash2022transframer}, %
\revision{at the potential cost of poorer temporal consistency~\cite{lemoing2021ccvs}.} %

\begin{figure}
    \centering
    \includegraphics[width=0.88\linewidth]{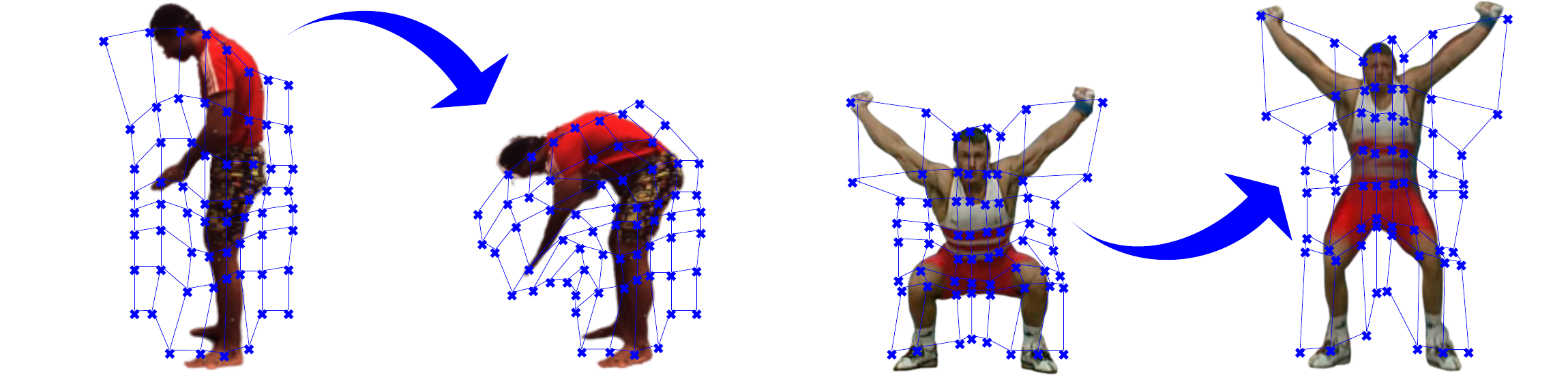} %
    \begin{tabular}{C{1.43cm}C{1.43cm}C{1.43cm}C{1.43cm}}
     {\scriptsize $T$} & {\scriptsize $T{+}10$} & {\scriptsize $T$} & {\scriptsize $T{+}10$} \\
    \end{tabular}
	\caption{\new{\textbf{WALDO} synthesizes future frames by deforming grids of control points, automatically associated with different objects.}} %
	\label{fig:teaser}
\end{figure}

\begin{figure*}
    \centering
    \includegraphics[width=.93\linewidth]{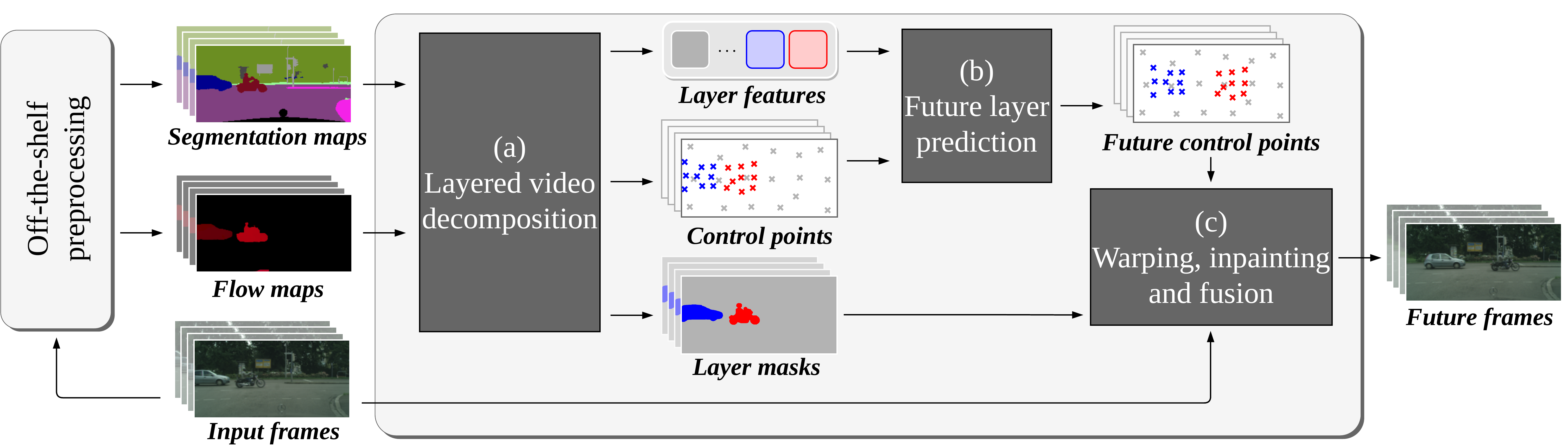} %
	\caption{\textbf{Overview of WALDO.} Given an input video sequence and associated semantic segmentation and optical flow maps \change{preprocessed by off-the-shelf models~\cite{teed2020raft, chen2018encoder}}, our approach breaks down video synthesis into {\bf (a) layered video decomposition:} using semantic and motion cues to decompose the sequence into layers represented by both object masks and features, with spatial information in the form of a small set of control points, {\bf (b) future layer prediction:} predicting the new position of these points in the target output frames, and {\bf (c) warping, inpainting and fusion:} using the corresponding offset and a thin-plate spline deformation model to warp the input frames and object masks, merge the corresponding regions, and fill in the empty image parts. \revision{Our model is trained, without explicit annotations, on a set of videos of $T{+}K$ frames by using the first $T$} to predict the next $K$. \revision{At inference, a (potentially greater) number $K$ of frames is predicted from $T$ input ones by repeating (a), (b) and (c) in an autoregressive fashion if needed.} %
	}
	\label{fig:overview}
\end{figure*} 

Our work relies instead on semantic and motion cues extracted from the past to model complex dynamics in high resolution and predict the future. %
Wu~\etal~\cite{wu2020future} decompose scenes into objects/background, predict affine transformations for the objects and non-affine ones for the background, and warp the last input frame to produce new ones. %
Bei~\etal~\cite{bei2021learning} %
\revision{predict dense flow maps for individual semantic regions}. %
Geng~\etal~\cite{geng2022comparing} augment classical frame reconstruction losses with flow-based correspondences between pairs of frames. %
\revision{Wu~\etal~\cite{wu2022optimizing} build on a pretrained video frame interpolation model which they adapt to future video prediction.} %
Contrary to~\cite{wu2020future}, our model \revision{automatically discovers the object decomposition} without explicit supervision. %
Moreover, rather than directly predicting optical flow at each pixel~\cite{wu2020future, bei2021learning, wu2022optimizing}, we use thin-plate splines~(TPS)~\cite{bookstein1989principal} as a parametric model of per layer flow for any pair of frames. %
This improves robustness since we predict future frames using all past ones as opposed to~\cite{wu2020future, bei2021learning, geng2022comparing, wu2022optimizing}. %
In addition, TPS provide optical flow at any resolution, and allow using a lower resolution for fast training while retaining good performance with high-resolution inputs at inference. %
Lastly, a few time-dependent control points associated with each object and the background are sufficient to parameterize TPS.
This compact yet expressive representation of motion allows us to break down video synthesis into: (a) discovering the layers associated with past frames, (b) predicting the corresponding transformations for upcoming ones \revision{while handling complex motions~\cite{chang2022strpm} and modeling future uncertainty if needed~\cite{akan2021slamp, akan2022stochastic}}, and (c) warping the associated object regions accordingly and filling in the remaining image parts. %
By combining these three components, our approach 
(Figure~\ref{fig:overview}) to predict future frames by {WA}rping {L}ayer-{D}ecomposed {O}bjects~(WALDO) from past ones sets a new state of the art on diverse benchmarks including urban scenes (Cityscapes~\cite{cordts2016cityscapes} and KITTI~\cite{geiger2013vision}), %
and scenes featuring nonrigid motions (UCF-Sports~\cite{rodriguez2008action} and H3.6M~\cite{ionescu2013human3}). %
Our main contributions \remove{can be summarized as follows} \new{are twofold}: %

\noindent {\bf $\bullet$ Much broader operating assumptions:} Previous approaches to video prediction that decompose individual frames into layers assume prior foreground/background knowledge~\cite{wu2020future} or reliable keypoint detection~\cite{walker2017pose, ye2019compositional}, and they do not allow the recovery of dense scene flow at arbitrary resolutions for arbitrary pairs of frames~\cite{wu2020future, bei2021learning, geng2022comparing, wu2022optimizing, akan2021slamp, akan2022stochastic, chang2022strpm}.
Our approach overcomes these limitations.
Unlike~\cite{wu2020future, bei2021learning, geng2022comparing, wu2022optimizing, chang2022strpm}, it also allows multiple predictions. %

\noindent {\bf $\bullet$ Novelty:} The main novelties of our approach to video prediction are (a) a layer decomposition algorithm that leverages a transformer architecture to exploit long-range dependencies between semantic and motion cues; %
(b) a low-parameter deformation model that allows the long-term prediction of sharp frames from a small set of adjustable control points;
and (c) the effective fusion of multiple predictions from past frames using state-of-the-art inpainting %
to handle (dis)occlusion. %
Some of these elements have been used separately in the past (e.g.,~\cite{yang2021selfsupervised, ye2019compositional, lemoing2021ccvs}), but never, to the best of our knowledge, in an integrated setting.

\remove{\noindent {\bf (3) An extensive validation:} Computer vision is an experimental engineering field. 
Designing and assembling the different modules that make up our approach
is a contribution on its own, but
validating it on multiple benchmarks and settings is the ultimate test.
We present experiments on urban datasets (Cityscapes and KITTI),
as well as scenes featuring nonrigid motions (UCF-Sports, H3.6M), and show that our method outperforms the state of the art by a significant margin in {\em every case}.} %
\new{These contributions are validated through extensive experiments on urban datasets (Cityscapes and KITTI),
as well as scenes featuring nonrigid motions (UCF-Sports, H3.6M), where our method outperforms the state of the art by a significant margin in {\em every case}}

\section{Related work}
\myparagraph{Video prediction} %
ranges from unconditional synthesis~\cite{tulyakov2018mocogan, denton2018stochastic, vondrick2016generating, franceschi2020stochastic, clark2019adversarial, saito2020train, tian2021good, finn2016unsupervised, babaeizadeh2018stochastic, kumar2020video} to multi-modal and controlled prediction tasks~\cite{hao2018controllable, lemoing2021ccvs, wu2021nuwa, han2022show, hu2022make}. %
Here, we intend to exploit the temporal redundancy of videos by tracking the trajectories of different objects.
One may infer future frames by extrapolating the position of keypoints associated to target objects %
\cite{walker2017pose, ye2019compositional}, but this requires manual labeling. %
Some works~\cite{chang2016compositional, hsieh2018learning, minderer2019unsupervised, henderson20neurips, ehrhardt2020relate} propose instead to let structured object-related information naturally emerge from the videos themselves.
\revision{We use a similar strategy} but also rely on off-the-shelf models to extract semantic and motion cues in the hope of better capturing the scene dynamics~\cite{wang2018video,bei2021learning, schmeckpeper2021object, wu2020future, geng2022comparing, wu2022optimizing}. 
Without access to ground-truth objects, we discover them via layered video decomposition.

\myparagraph{Layered video decompositions,} introduced in~\cite{wang1994representing}, have been applied to optical flow estimation~\cite{sun2013fully, wulff2015efficient}, motion segmentation~\cite{pawan2008learning, yang2021selfsupervised}, and video editing~\cite{jojic2001learning,lu2021omnimatte, ye2022sprites}. %
They are connected to object-centric representation learning~\cite{burgess2019monet,engelcke2019genesis,greff2019multi,locatello2020object, bao2022discovering, kipf2022conditional}, where the compositional structure of scenes is also essential. Like~\cite{yang2021selfsupervised}, we use motion in the form of optical flow maps to decompose videos into objects and background. %
We go beyond the single-object scenarios they tackle, and propose a decomposition scheme which works well on real-world scenarios like urban scenes, with multiple objects and complex motions. %
In addition, we associate with every layer a geometric transformation allowing the recovery of the flow between past and future frames. %

\myparagraph{Spatial warps,} \revision{as implemented in~\cite{jaderberg2015spatial}}, %
have proven useful for various tasks, \eg, automatic image rectification for text recognition~\cite{shi2016robust}, semantic segmentation~\cite{ganeshan2021warp}, and the contextual synthesis of images~\cite{zhou2016view, monnier2021unsupervised, park2017transformation} or videos~\cite{vondrick2017generating, wu2020future,bei2021learning,luc2020transformation,gao2019disentangling,hao2018controllable, bar2020compositional, li2018flow, akan2021slamp, akan2022stochastic, wu2022optimizing}.
\new{Inspired by D'Arcy Thompson's pioneering work in biology~\cite{thompson1992} as well as the shape contexts of Belongie et al.~\cite{belongie2002shape}}
\remove{Here}, we parameterize the warp with thin-plate splines~(TPS)~\cite{bookstein1989principal}, whose parameters are motion vectors sampled at a small set of control points. %
TPS allow optical flow recovery by finding the transformations of minimal bending energy which complies with points motion. %
This has several advantages: The flow is differentiable with respect to motion vectors; deformations are more general than affine ones; and the number of control points allow to trade off deformation expressivity for parameter size. %

\section{Proposed method}
\label{sec:method}
\myparagraph{Notation.} \revision{We use a subscript $t$ in $\llbracket 1, T{+}K \rrbracket$ for {\em time}, with frames $1$ to $T{+}K$ available at train time, and the last $K$ predicted from the first $T$ at inference time. We use a superscript $i$ in $\llbracket 0, N \rrbracket$ for {\em image layer}, where $i{=}0$ represents the background and $i{>}0$ represents an object.} %
For example, we denote by $p_t^i$ the control points associated with layer $i$ in frame $t$, by $p_t$ those associated with all layers at time $t$, and by $p^i$ those associated with layer $i$ over all time steps.

\myparagraph{Overview.} We consider a video $X$ consisting of \revision{$T{+}K$ RGB frames $x_1$ to $x_{T+K}$} with spatial resolution $H \stimes W$.
Each frame is decomposed into $N{+}1$ layers tracking objects and background motions over time. Layers are represented by pairs $(m_t^i,p_t^i)$, where $m_t^i$ is a (soft) mask indicating for every pixel the presence of object $i$ (or background if $i{=}0$), and $p_t^i$ is the set of control points associated with layer $i$ at time $t$. %
Our approach (Figure~\ref{fig:overview}) \revision{consists of: (a)} decomposing \revision{frames $1$ to $T$ of} video $X$ into layers (Sec.~\ref{sec:dec});
\revision{(b)} predicting the decompositions up to time $T{+}K$ (Sec.~\ref{sec:for});
and \revision{(c)} using them to warp each of the \revision{first} $T$ frames of $X$ into the $K$ future time steps, and finally predicting frames $T{+}1$ to $T{+}K$ by fusing the $T$ views for each future time step and filling in empty regions (Sec.~\ref{sec:inp}). %
\revision{The three corresponding modules are trained separately, starting with (a), and then using (a) to supervise (b) and (c).}

\myparagraph{Off-the-shelf preprocessing.} Similar to~\cite{yang2021selfsupervised, bao2022discovering}, we adopt a motion-driven definition of \emph{objectness} where an object is defined as a spatially-coherent region which follows a smooth deformation over time, such that discontinuities in scene motion occur at layer boundaries.
We compute for the first $T$ frames of the video the corresponding backward flow maps \revision{$F=[f_1,\ldots,f_{T}]$} with an off-the-shelf method~\cite{teed2020raft}, %
where $f_t$ \revision{is the translation map associating with every pixel of frame $x_t$ the vector of $\RR^2$} pointing to the matching location in $x_{t-1}$. %
We also suppose that each object has a unique semantic class out of $C$ and compute for the first $T$ frames the corresponding semantic segmentation maps \revision{$S=[s_1,\ldots,s_{T}]$} again with an off-the-shelf %
method~\cite{chen2018encoder}, %
where $s_t$ assigns to every pixel a label (\eg, car, road, buildings, sky) represented by its index in $\llbracket 1, C \rrbracket$. %

\subsection{Layered video decomposition}
\label{sec:dec}

\begin{figure}
    \centering
    \includegraphics[width=.90\linewidth]{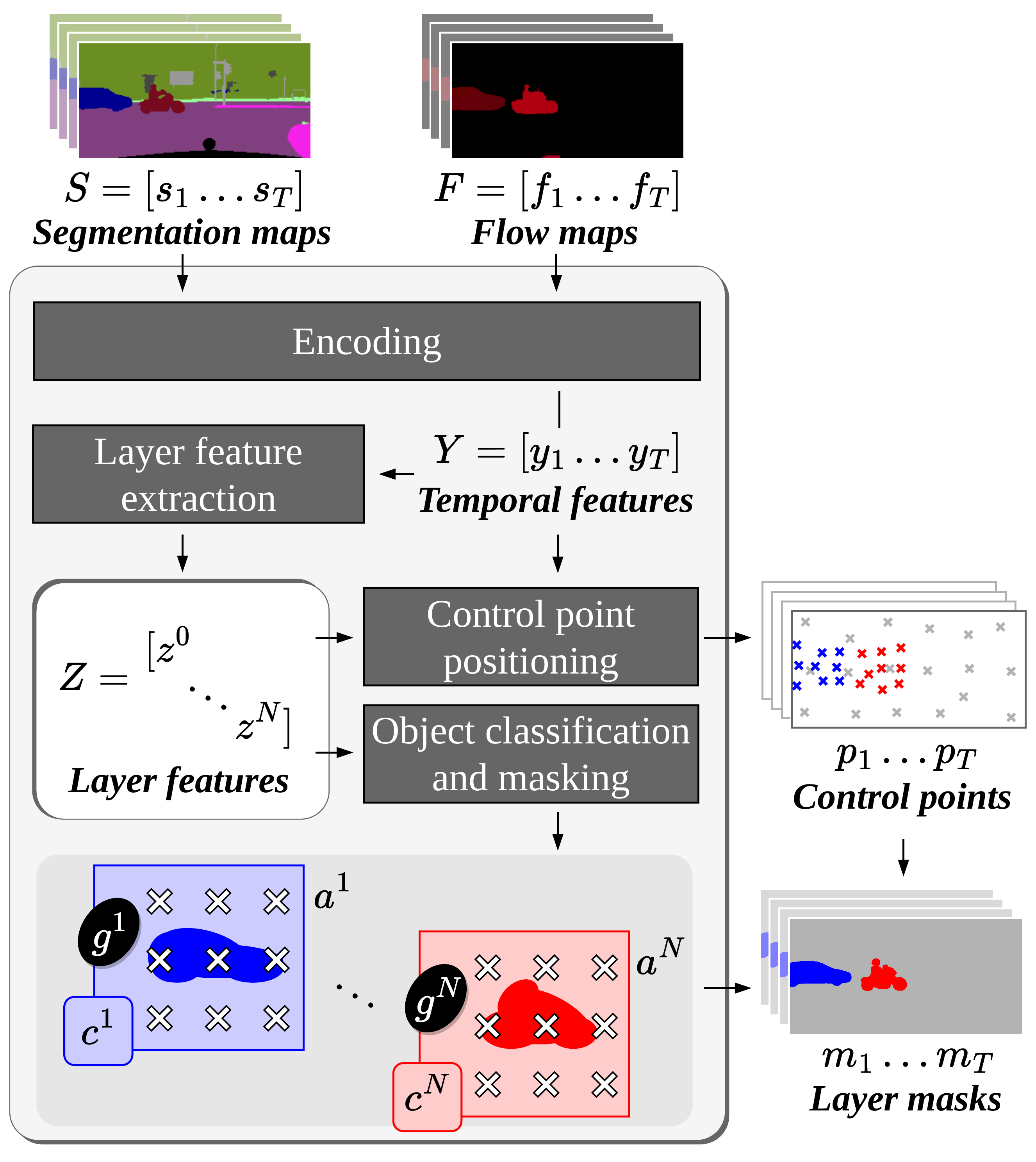} %
	\caption{\textbf{Layered video decomposition.} Semantic and motion cues are mapped to temporal features with a time-independent encoder. These are combined to form layer features representing the background and the objects for the whole video. The position of individual layers at each time step is determined by a sparse set of control points, predicted from temporal features and layer features. We further associate with the layers a soft mask and a semantic class which are combined with the information of position to produce layer masks, segmenting the objects from the background.} %
	\label{fig:decomposition}
\end{figure}

We associate with each layer $i$ in $\llbracket 0, N \rrbracket$ a (soft) object ($i{>}0$) or background ($i{=}0$) mask $a^i$ of size $H^i\stimes W^i$ over which is overlaid a coarse $h^i\stimes w^i$ regular grid $g^i$ of control points.\footnote{In practice we use larger values of $H^i$, $W^i$ for the background (same as $H$, $W$) than for object layers, \change{and we fix the ratio $H^i/h^i$ to $16$.}} Deformed {\em layer masks} $m_t^i$ are obtained by mapping the points in $g^i$ onto their positions $p_t^i$ at time step $t$ and applying the corresponding TPS transformation to $a^i$.
The decomposition module %
(Figure~\ref{fig:decomposition}) maps the segmentation and flow maps $S$ and $F$ onto the positioned control points $p_1$ to $p_T$ and the deformed layer masks $m_1$ to $m_T$. %

\subsubsection{Architecture}
\myparagraph{Input encoding.}
Input flow maps and segmentation maps $F$ and $S$
are fed to a time-independent encoder, implemented by a convolutional neural network (CNN) which outputs temporal feature maps $Y=[y_1,\ldots,y_T]$. Each map $y_t$ lies in $\RR^{d \times h \times w}$ with feature dimension $d$ and downscaled spatial resolution $h\stimes w$ such that $hw \ll HW$. %
We denote by $l=hw$ the latent feature size, and reshape feature $y_t$ to an $l \stimes d$ matrix through raster-scan reordering. %

\myparagraph{Layer feature extraction.}
We form layer features $Z=[z^0,\ldots,z^N]$ from temporal ones $Y$, where, for each layer, $z^i$ is an $l^i \stimes d$ matrix with $l^i=h^iw^i$. %
We implement this with a transformer~\cite{vaswani2017attention}, %
where self-attention is replaced \revision{by a binding mechanism, discovering object-centric features by iteratively grouping intra-object pixels together}, as in~\cite{locatello2020object}. %

\myparagraph{Control point positioning.}  We add a third dimension to the control points $p_t^i$ positioned in the frame $x_t$ to record the corresponding layer ``depth'' ordering $o_t^i{\ge}0$, with $o_t^i{=}0$ for the background. In practice, the 3D vectors associated with the points $p_t$ are once again predicted by a transformer, from the set $Z$ of layer features and the temporal feature $y_t$, thus accounting for possible interactions between layers. \remove{We note that} Control points are typically close to the associated objects, but it is still fine if some end up outside an object mask. Their role is to recover a dense deformation field through TPS interpolation, and what matters is only that the part of this field within the object mask is correct. 
\new{Figure~\ref{fig:teaser} shows grids of control points extracted by our approach.
Note how their structure automatically adapts to fit precise motions.}

\myparagraph{Layer mask prediction.}
A CNN maps layer features $z^i$ onto soft masks $a^i$ with values in $[0, 1]$ corresponding to opaqueness %
\corgrid{and defined in their own ``intrinsic'' coordinate systems, in which points associated with $p_t^i$ lie on a regular grid $g^i$.}
The background is fully opaque ($a^0{=}\mathbf{1}$). 
At time $t$, $a^i$ is warped onto the corresponding layer mask $m_t^i$
using the TPS transformation $w_t^i$ mapping the grid points of $g^i$
onto their positions $p_t^i$ in frame $x_t$. %
We then improve object contours by semantic refinement using segmentation maps $S$. %
Concretely, given a mask $m_t^i$ corresponding to an object layer ($i{>}0$), a soft class assignment $c^i$ in $[0,1]^C$ is obtained from $z^i$ with a fully-connected layer, then used to update $m_t^i$ by comparing $c^i$ to the actual class in $s_t$. %
We finally use the ordering scores $o_t$ and a classical occlusion model~\cite{monnier2021unsupervised, jojic2001learning} to filter non-visible layer parts. More details about this process are in Appendix \ref{sec:tps-warp}-\ref{sec:occ}.

\subsubsection{Training procedure}
\label{sec:dec-train}
\revision{The goal of the decomposition module
is to discover layers whose associated masks and control points best reconstruct scene motion. We achieve this by minimizing the objective:}
\begin{equation}
\label{eq:dec}
    \mathcal{L}_{{d}}=\lambda_{\text{o}}\mathcal{L}_{{o}}+\lambda_{\text{f}}\mathcal{L}_{{f}}+\lambda_{\text{r}}\mathcal{L}_{{r}},
\end{equation}
with an object discovery loss $\mathcal{L}_{{o}}$ to encourage objects from different layers to occupy moving foreground regions; %
a flow reconstruction loss $\mathcal{L}_{{f}}$ to ensure the temporal consistency of the learned decompositions; and a regularization loss $\mathcal{L}_{{r}}$.
\revision{At train time, we extract layer features from the first $T$ frames, but also position layers in the $K$ subsequent ones to have a supervision signal for later stages (Secs.~\ref{sec:for} and \ref{sec:inp}). At inference, only the first $T$ frames are used.} %

\myparagraph{Object discovery.} Without ground-truth objects for training, we \revision{discover reasonable candidates using} semantic and motion cues, $f_t$ and $s_t$, \revision{by positioning objects in $\mathcal{M}(s_t,f_t)$, a binary mask indicating \emph{foreground} regions (\eg, cars) with significant motion compared to \emph{stuff} regions (\eg, road), written $\mathcal{S}(s_t)$. We get the mask $\mathcal{M}(s_t,f_t)$ by extrapolating the full background flow from $f_t$ in $\mathcal{S}(s_t)$, and by thresholding with a constant $\tau_{m}$ the $L_1$ distance between background and foreground flows. } %
The object discovery loss is:
\begin{equation}
    \mathcal{L}_{o}=\sum_{t} (k_{s}\mathcal{S}(s_t)-k_{m}\mathcal{M}(s_t,f_t))\odot(\max\limits_{i>0}m_t^i),
\end{equation} %
\change{where $k_{m}$ and $k_{s}$ are positive scalars, weighting the attraction of \revision{discovered objects $m_{t}^{i>0}$} towards moving \emph{foreground} regions ($\mathcal{M}$) and the repulsion from \emph{stuff} regions ($\mathcal{S}$)}.

\myparagraph{Flow reconstruction.} We reconstruct the backward flow $w_{{t_2}\leftarrow{t_1}}$ between consecutive time steps $t_1$ and $t_2$, by considering the individual layer warps, denoted $w_{{t_2}\leftarrow{t_1}}^i$ %
and computed as $w_{t_2}^i\circ w_{t_1}^i{}^{-1}$ where $w_{t_1}^i$ and $w_{t_2}^i$ are obtained through the TPS transformation associated with $p_{t_1}^i$ and $p_{t_2}^i$. %
We recover $w_{{t_2}\leftarrow{t_1}}$ by compositing layer warps, \ie, $w_{{t_2}\leftarrow{t_1}}=\sum_i{m_{t_2}^i\odot{w^i_{{t_2}\leftarrow{t_1}}}}$, where the mask $m_{t_2}^i$ determines layer transparency. The flow reconstruction loss is: %
\begin{equation}
    \label{eq:flow}
    \mathcal{L}_{{f}}=\sum_t\|f_t-\hat{f}_t\|_1, \,\, \text{with} \,\,\,\, \hat{f}_t=w_{t\leftarrow{t-1}}.
\end{equation}

\myparagraph{Regularization.}
The last objective $\mathcal{L}_{{r}}$ %
is composed of: an entropy term applied to layer mask $m_t$ to ensure that %
a single layer prevails
for every pixel, and an object initialization term which is the $L_2$ distance from regions of interest $\mathcal{M}(s_t,f_t)$ which are still empty (as per $m_t^{i>0}$) to the control points $p_t^{i}$ of the nearest object (${i{>}0}$). %

\subsection{Future layer prediction}
\label{sec:for}

\begin{figure}
    \centering
    \includegraphics[width=.90\linewidth]{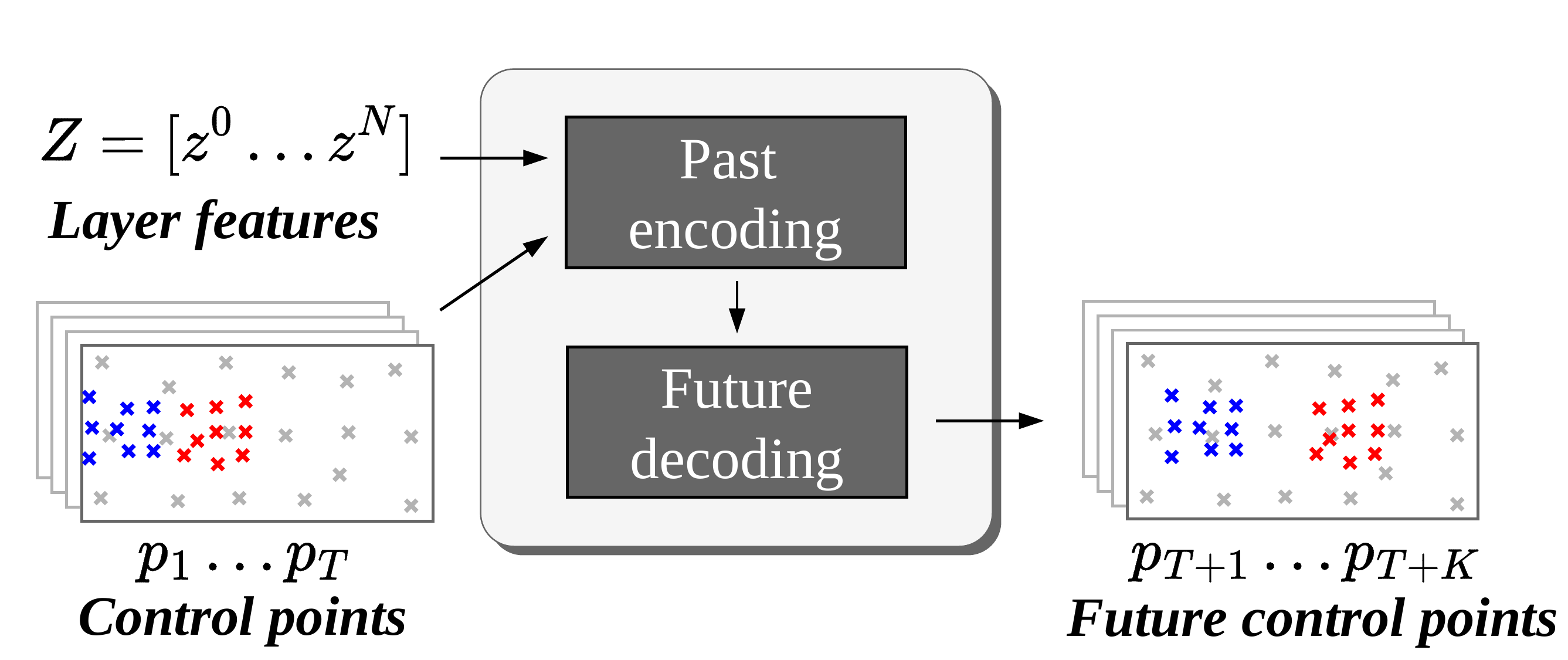} %
	\caption{\textbf{Future layer prediction.} We encode past control point positions and \revision{object-specific knowledge (\eg, shape and semantics) in the form of} layer features to produce future positions.}
	\label{fig:future}
\end{figure}

Thanks to our layered video decomposition, the prediction of future layers (Figure~\ref{fig:future}) is reduced to inferring, from past control points  $[p_1,\ldots,p_T]$, the position of future ones. %

\myparagraph{Architecture.}
For each time step $t{\leq}T$ and each layer $i$, $p_t^i$ is mapped onto a vector in $\RR^d$ by a linear layer.
Likewise, each $z^i$ is also mapped onto a vector in $\RR^d$.
We concatenate and feed these vectors %
to a two-stage transformer
to construct representations for the $K$ future time steps by combining self-attention modules applied to intermediate future representations and cross-attention modules between past and future ones. %
\change{The prediction of the control points $p_t^i$ associated with each layer in the $K$ future time steps is done with a linear layer,} which does not output their position directly, but rather their displacement with respect to $p_T$, with $T$ being the last known time step from the past context.

\myparagraph{Training procedure.}
We extract training data for past and future time steps by decomposing videos of length $T{+}K$ as described in Sec.~\ref{sec:dec}. %
We then mask the control points corresponding to the last $K$ time steps and train the future prediction module to reconstruct them \revision{by minimizing $\mathcal{L}_{{p}}$, the $L_1$ distance between extracted and reconstructed control points for time steps $T{+}1$ to $T{+}K$.}
Under uncertainty, $\mathcal{L}_{{p}}$ makes points converge towards an \emph{average} future trajectory. When interested in predicting multiple futures, we add noise \revision{(as input and in attention modules)} and an adversarial term~\cite{goodfellow2014generative} to $\mathcal{L}_{{p}}$. More details are in \app{sec:sto}. %

\subsection{Warping, fusion and inpainting}
\label{sec:inp}

\begin{figure}
    \centering
    \includegraphics[width=.90\linewidth]{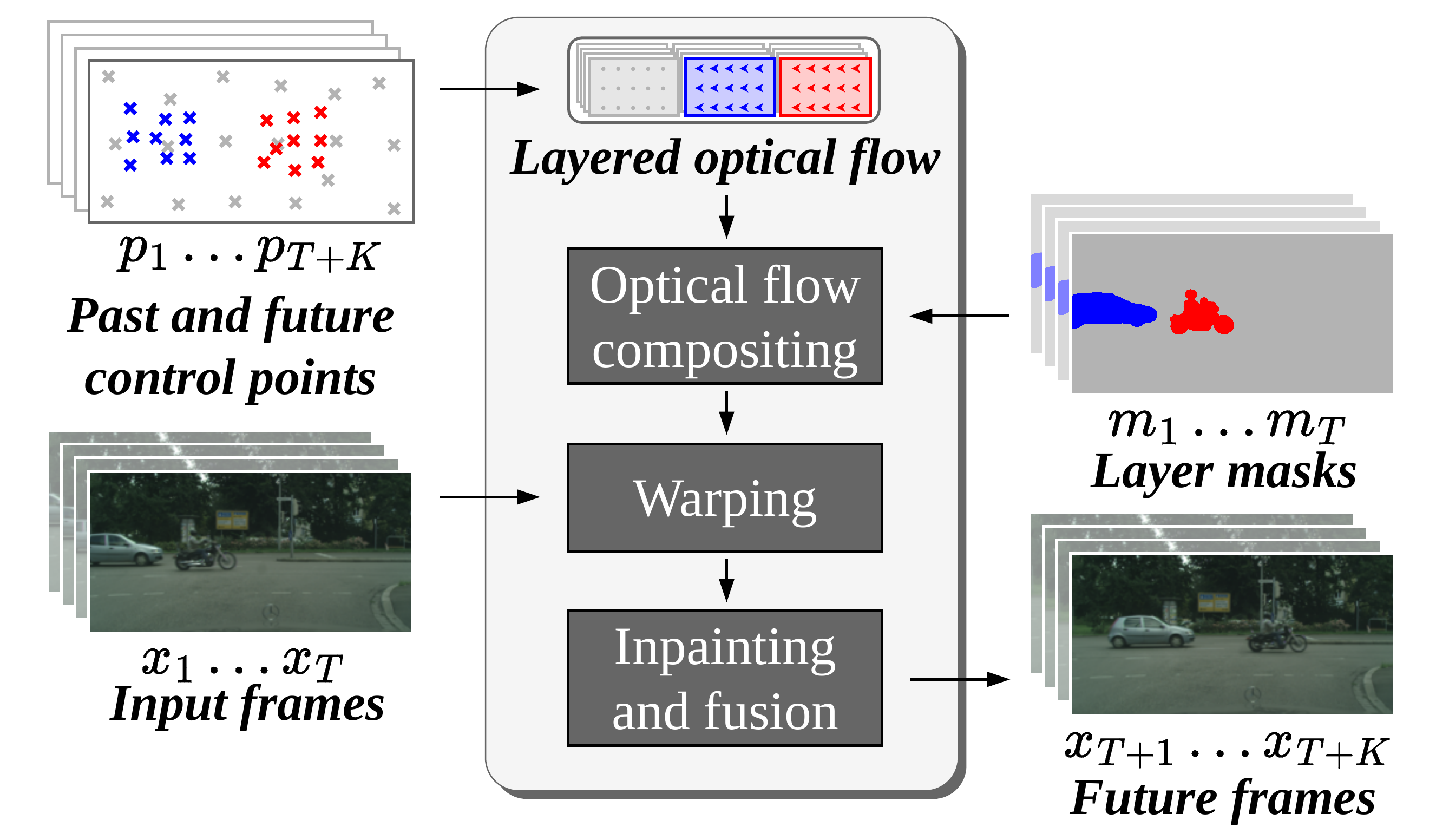} %
	\caption{\textbf{Warping, fusion and inpainting.} From past and future control point positions we compute the warps associated with each layer and composite them to recover the dense scene flow by leveraging past layer masks for setting their transparency. \change{We then warp frames from the past to form multiple views of future ones, merge these together, and fill in empty regions to produce final frames}.}
 \label{fig:warping} 
\end{figure}

Last comes the actual synthesis of future frames from past ones (Figure~\ref{fig:warping}), using the layer decompositions extracted from the past, and the ones predicted into the future. %

\myparagraph{Architecture.}
Given the predicted control points, we compute, for every layer $i$ and pair of time steps $(t_1,t_2)$ in $\llbracket 1, T \rrbracket\stimes\llbracket T+1, T+K \rrbracket$, the layer warp $w^i_{{t_2}\leftarrow{t_1}}$ from $t_1$ to $t_2$ as described in Sec.~\ref{sec:dec}.
We construct $m_{t_2}^i$ by warping $m_{t_1}^i$ with $w^i_{{t_2}\leftarrow{t_1}}$. %
We compute $w_{{t_2}\leftarrow{t_1}}$ from layer warps and future masks by $w_{{t_2}\leftarrow{t_1}}=\sum_i{m_{t_2}^i\odot{w^i_{{t_2}\leftarrow{t_1}}}}$ as before, and warp past frames to produce multiple views of future ones, one for each pair of time steps $(t_1,t_2)$.
We obtain $T$ views for each future frame, with different missing regions due to disocclusion.
\change{A fusion network, implemented by a U-Net~\cite{ronneberger2015unet}, merges these views together according to pixel-level scores predicted for each of them.
Regions which remain empty are then filled with an off-the-shelf inpainting network~\cite{li2022mat}, to obtain future frame predictions $\hat{x}_{T+1}$ to $\hat{x}_{T+K}$. We ensure temporal consistency by filling frames one at a time and by using predicted motions to propagate the newly filled content onto the next frames.}

\myparagraph{Training procedure.}
\revision{The inpainting model~\cite{li2022mat}, trained on $8$M images from the Places dataset~\cite{zhou2017places}, is kept frozen}
and the objective to train the fusion U-Net is the weighted $L_1$ distance in pixel space and between features $\mathcal{F}$ 
extracted using the $\text{VGG}$~\cite{simonyan2015very} classification network trained on~\cite{deng2009imagenet}:
\begin{equation}
\label{eq:fusion}
    \mathcal{L}_{{u}}=\sum_{t=T+1}^{T+K}\lambda_{p}\|x_t-\hat{x}_t\|_1+\lambda_{v}\|\mathcal{F}(x_t)-\mathcal{F}(\hat{x}_t)\|_1,
\end{equation}

\section{Experiments}
\label{sec:exp}

\myparagraph{Datasets.} We train WALDO on two urban datasets, Cityscapes~\cite{cordts2016cityscapes}, which contains $2975$ $30$-frame video sequences for training and $500$ for testing captured at $17$ FPS, and KITTI~\cite{geiger2013vision}, with a total of $156$ longer video sequences (${\sim}340$ frames each) including $4$ for testing. We use suitable resolutions and train / test splits for fair comparisons with prior works (setup from~\cite{bei2021learning} in Table~\ref{tab:sota} and from \cite{akan2021slamp} in Table~\ref{tab:sota2}). %
We also train on nonrigid scenes from UCF-Sports~\cite{rodriguez2008action} (resp. H3.6M~\cite{ionescu2013human3}) using the splits from~\cite{chang2022strpm} consisting of $6288$ sequences (resp. $73404$) for training and $752$ (resp. $8582$) for testing with roughly $10$ frames per sequence. %
We extract semantic and motion cues using pretrained models, namely, DeepLabV3~\cite{chen2018encoder} %
and  RAFT~\cite{teed2020raft}. %

\myparagraph{Evaluation metrics.}
We evaluate the different methods with the multi-scale structure similarity index measure (SSIM)~\cite{wang2004image}, the learned perceptual image patch similarity (LPIPS)~\cite{zhang2018the}, and the peak signal-to-noise ratio (PSNR), all standard image reconstruction metrics for evaluating video predictions.
We also use the Fréchet video distance (FVD)~\cite{unterthiner2019fvd} to estimate the gap between real and synthetic video distributions. %
We report in Tables~\ref{tab:sota}-\ref{tab:sota3} the mean and standard deviation for $3$ randomly-seeded training sessions. 

\begin{table*}
\setlength\heavyrulewidth{.25ex}
\aboverulesep=0ex
\belowrulesep=.3ex
\centering
\caption{Comparison to state-of-the-art deterministic methods on Cityscapes and KITTI test sets. We compute multi-scale SSIM ({\tiny$\times 10^3$}) and LPIPS ({\tiny$\times 10^3$}) for the $k^\text{th}$ future frame and average for $k$ in $\llbracket 1, K \rrbracket$. We indicate if methods use semantic or flow ground truths for training.}
\small
\begin{tabular}{@{}c@{\hskip 0.65em}c@{\hskip 0.65em}c@{}}

\begin{tabular}{@{}L{1.7cm}@{}C{0.8cm}@{}C{0.8cm}@{}}
~ \\
\toprule
\multirow{2}{*}{\normalsize Method} & \multirow{2}{*}{Sem.} & \multirow{2}{*}{Flow} \\[1.4pt]
\\
\cmidrule(){1-3}
PredNet~\cite{lotter2017deep} & & \\
MCNet~\cite{villegas2017decomposing} & & \\
VFlow~\cite{liu2017video} & & \\
VEST~\cite{zhang2022video} & & \\
VPVFI~\cite{wu2022optimizing} & & \checkmark \\
VPCL~\cite{geng2022comparing} & & \checkmark \\
Vid2vid~\cite{wang2018video} & \checkmark & \checkmark \\
OMP~\cite{wu2020future} & \checkmark & \checkmark \\
SADM~\cite{bei2021learning} & \checkmark & \checkmark \\
\cmidrule(){1-3}
WALDO & \checkmark & \checkmark \\
\bottomrule 
\end{tabular} &

\begin{tabular}{@{}L{0.90cm}@{}L{1.00cm}@{}L{0.90cm}@{}L{1.00cm}@{}L{0.90cm}@{}L{1.00cm}@{}L{0.90cm}@{}}
\multicolumn{7}{@{}c@{}}{(a) Cityscapes ({\tiny $512\stimes1024$})} \\
\toprule
\multicolumn{2}{@{}c@{}}{$K=1$} & \multicolumn{2}{@{}c@{}}{$K=5$} & \multicolumn{3}{@{}c@{}}{$K=10$} \\
\cmidrule(r){1-2}\cmidrule(r){3-4}\cmidrule(){5-7}
{\footnotesize SSIM $\uparrow$} & {\footnotesize LPIPS $\downarrow$} & {\footnotesize SSIM $\uparrow$} & {\footnotesize LPIPS $\downarrow$} & {\footnotesize SSIM $\uparrow$} & {\footnotesize LPIPS $\downarrow$} & {\footnotesize FVD $\downarrow$} \\
\cmidrule(){1-7}
~~840 & ~~260 & ~~752 & ~~360 & ~~663 & ~~522 & ~~~~-\\
~~897 & ~~189 & ~~706 & ~~373 & ~~597 & ~~451 & ~~~~-\\
~~839 & ~~174 & ~~711 & ~~288 & ~~634 & ~~366& ~~~~-\\
~~~~- & ~~~~- & ~~~~- & ~~~~- & ~~~~- & ~~~~- & ~~~~-\\
~~945 & ~~064 & ~~804 & ~~178 & ~~700 & ~~278 & ~~159\\
~~928 & ~~085 & ~~\underline{839} & ~~150 & ~~\underline{751} & ~~\underline{217} & ~~129\\
~~882 & ~~106 & ~~751 & ~~201 & ~~669 & ~~271 & ~~~~-\\
~~891 & ~~085 & ~~757 & ~~165 & ~~674 & ~~233 & ~~\underline{113}\\
~~\textbf{959} & ~~\underline{076} & ~~835 & ~~\underline{149} & ~~~~- & ~~~~- & ~~~~-\\
\cmidrule(){1-7}
~~$\underline{957}{}_{\scriptscriptstyle\pm2}$ & ~~$\textbf{049}_{\scriptscriptstyle\pm1}$ & ~~$\textbf{854}_{\scriptscriptstyle\pm1}$ & ~~$\textbf{105}_{\scriptscriptstyle\pm1}$ & ~~$\textbf{771}_{\scriptscriptstyle\pm1}$ & ~~$\textbf{158}_{\scriptscriptstyle\pm1}$ & ~~$\textbf{055}_{\scriptscriptstyle\pm1}$ \\
\bottomrule 
\end{tabular} &  

\begin{tabular}{@{}L{0.90cm}@{}L{1.00cm}@{}L{0.90cm}@{}L{1.00cm}@{}L{0.90cm}@{}L{1.00cm}@{}L{0.90cm}@{}}
\multicolumn{7}{@{}c@{}}{(b) KITTI ({\tiny $256\stimes832$})}\\
\toprule
\multicolumn{2}{@{}c@{}}{$K=1$} & \multicolumn{2}{@{}c@{}}{$K=3$} & \multicolumn{3}{@{}c@{}}{$K=5$} \\
\cmidrule(r){1-2}\cmidrule(r){3-4}\cmidrule(){5-7}
{\footnotesize SSIM $\uparrow$} & {\footnotesize LPIPS $\downarrow$} & {\footnotesize SSIM $\uparrow$} & {\footnotesize LPIPS $\downarrow$} & {\footnotesize SSIM $\uparrow$} & {\footnotesize LPIPS $\downarrow$} & {\footnotesize FVD $\downarrow$} \\
\cmidrule(){1-7}
~~563 & ~~553 & ~~514 & ~~586 & ~~475 & ~~629 & ~~~~- \\
~~753 & ~~240 & ~~635 & ~~317 & ~~554 & ~~373 & ~~~~- \\
~~539 & ~~324 & ~~469 & ~~374 & ~~426 & ~~415 & ~~~~- \\
~~~~- & ~~156 & ~~~~- & ~~344 & ~~~~- & ~~447 & ~~~~- \\
~~827 & ~~\underline{123} & ~~695 & ~~\underline{203} & ~~611 & ~~264 & ~~050\\
~~{820} & ~~172 & ~~\underline{730} & ~~{220} & ~~\underline{667} & ~~\underline{259} & ~~075 \\
~~~~- & ~~~~- & ~~~~- & ~~~~- & ~~~~- & ~~~~- & ~~~~- \\
~~792 & ~~185 & ~~676 & ~~246 & ~~607 & ~~304 & ~~\underline{047} \\
~~\underline{831} & ~~{144} & ~~{724} & ~~246 & ~~{647} & ~~312 & ~~~~-\\
\cmidrule(){1-7}
~~$\textbf{867}_{\scriptscriptstyle\pm1}$ & ~~$\textbf{108}_{\scriptscriptstyle\pm1}$ & ~~$\textbf{766}_{\scriptscriptstyle\pm4}$ & ~~$\textbf{163}_{\scriptscriptstyle\pm2}$ & ~~$\textbf{702}_{\scriptscriptstyle\pm6}$ & ~~$\textbf{206}_{\scriptscriptstyle\pm3}$ & ~~$\textbf{042}_{\scriptscriptstyle\pm2}$ \\
\bottomrule 
\end{tabular}\\

\end{tabular} 
\label{tab:sota}
\end{table*}

\begin{figure*}
	\setlength\tabcolsep{1.5pt}
	\renewcommand{\arraystretch}{0.5}
	\small
    \centering
	\begin{tabular}{ccccc}
        \cline{1-5}
        \multicolumn{1}{@{}|c|@{}}{\raisebox{-0.5em}[0pt][0pt]{OMP~\cite{wu2020future}}} & \multicolumn{1}{@{}c|@{}}{\raisebox{-0.5em}[0pt][0pt]{VPCL~\cite{geng2022comparing}}} & \multicolumn{1}{@{}c|@{}}{\raisebox{-0.5em}[0pt][0pt]{VPVFI~\cite{wu2022optimizing}}} & \multicolumn{1}{@{}c|@{}}{\raisebox{-0.5em}[0pt][0pt]{WALDO}} & \multicolumn{1}{@{}c|@{}}{\raisebox{-0.5em}[0pt][0pt]{Ground truth}} \\
 		\\

 		\includegraphics[width=.190\linewidth]{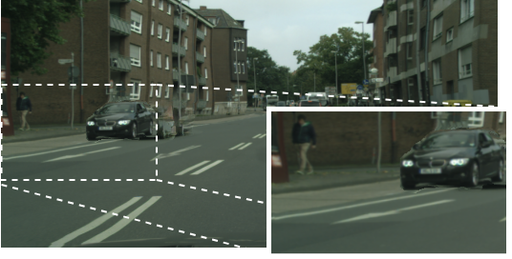} &
        \includegraphics[width=.190\linewidth]{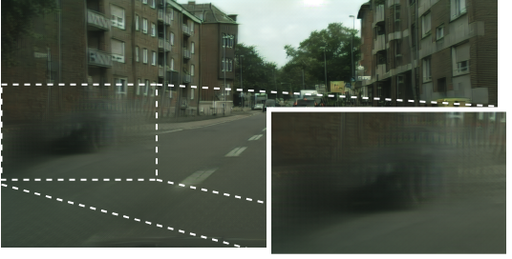} &
        \includegraphics[width=.190\linewidth]{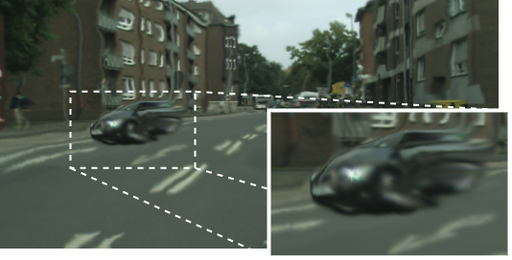} &
        \includegraphics[width=.190\linewidth]{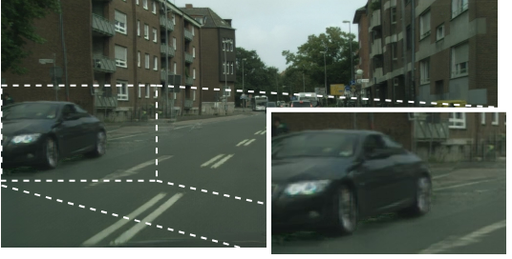} &
        \includegraphics[width=.190\linewidth]{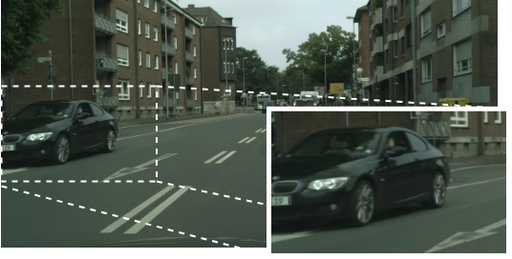} \\
   
 		\includegraphics[width=.190\linewidth]{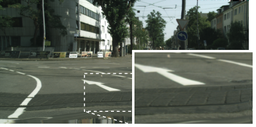} &
        \includegraphics[width=.190\linewidth]{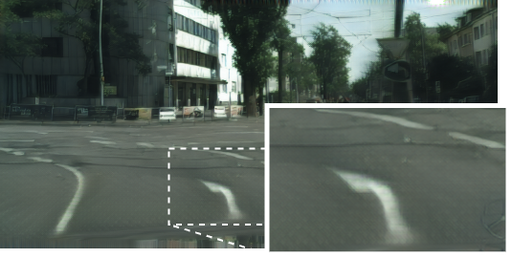} &
        \includegraphics[width=.190\linewidth]{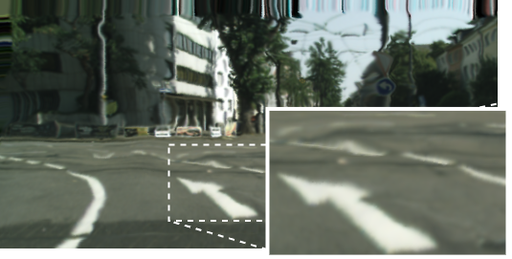} &
        \includegraphics[width=.190\linewidth]{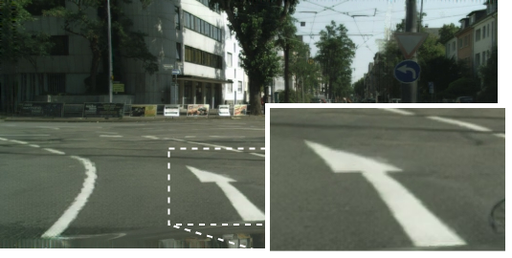} &
        \includegraphics[width=.190\linewidth]{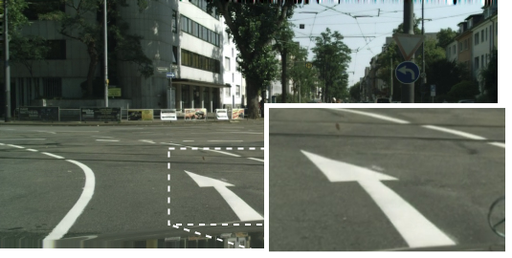} \\
   
 		\scalebox{-1}[1]{\includegraphics[width=.190\linewidth]{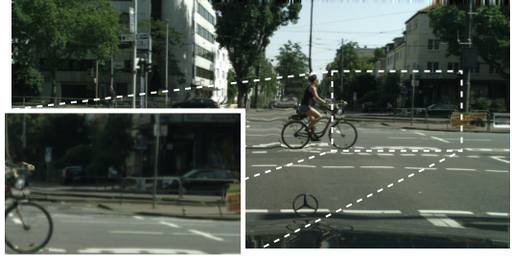}} &
        \scalebox{-1}[1]{\includegraphics[width=.190\linewidth]{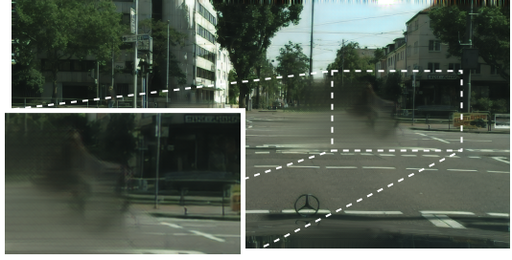}} &
        \scalebox{-1}[1]{\includegraphics[width=.190\linewidth]{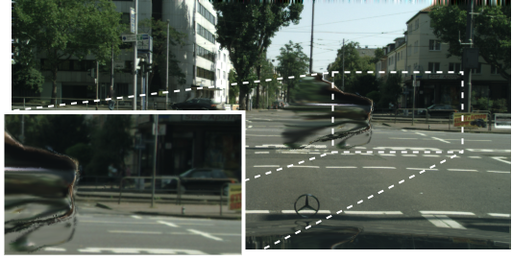}} &
        \scalebox{-1}[1]{\includegraphics[width=.190\linewidth]{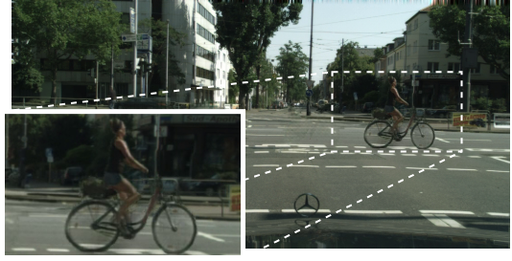}} &
        \scalebox{-1}[1]{\includegraphics[width=.190\linewidth]{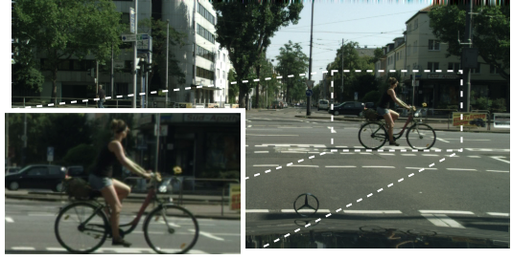}} \\

	\end{tabular}
	\caption{\change{Comparison with~\cite{wu2020future, geng2022comparing, wu2022optimizing} on the Cityscapes test set at time step $T{+}10$. WALDO better extracts objects from the background, better predicts their motion, and is more robust to occlusion.} We strongly encourage readers to watch videos in the \proj{}.} %
	\label{fig:qual}
\end{figure*}
\begin{table}
\setlength\heavyrulewidth{.25ex}
\aboverulesep=0ex
\belowrulesep=.3ex
\centering
\caption{\change{Comparison to  state-of-the-art \revision{stochastic methods} (\myquote{\textdagger}) on $20$-frame future prediction on Cityscapes and KITTI. We report the best SSIM ({\tiny$\times 10^3$}), PSNR ({\tiny$\times 10$}), and LPIPS ({\tiny$\times 10^3$}) out of $100$ trajectories sampled from each test sequence \revision{(except for our deterministic variant)}. We use $4$ frames as input but others use $10$}.} %
\small

\begin{tabular}{@{}c@{\hskip 0.40em}c@{\hskip 0.40em}c@{}}

\begin{tabular}{@{}L{2.1cm}@{}}
~ \\
\toprule
{\normalsize Method} \\ %
\cmidrule(){1-1}
SVG\textsuperscript{\textdagger}~\cite{denton2018stochastic} \\
SRVP\textsuperscript{\textdagger}~\cite{franceschi2020stochastic} \\
HierVRNN\textsuperscript{\textdagger}~\cite{castrejon2019improved} \\
SLAMP\textsuperscript{\textdagger}~\cite{akan2021slamp} \\
SLAMP-3D\textsuperscript{\textdagger}~\cite{akan2022stochastic} \\
\cmidrule(){1-1}
WALDO \\
WALDO\textsuperscript{\textdagger} \\
\bottomrule 
\end{tabular} &

\begin{tabular}{@{}C{0.90cm}@{}C{1.1cm}@{}C{0.95cm}@{}}
\multicolumn{3}{@{}c@{}}{(a) Cityscapes ({\tiny $128\stimes256$})} \\
\toprule
{\footnotesize SSIM $\uparrow$} & {\footnotesize PSNR $\uparrow$} & {\footnotesize LPIPS $\downarrow$} \\
\cmidrule(){1-3}
$606_{\scriptscriptstyle~~~~}$ & $204_{\scriptscriptstyle~~~~}$ & $340_{\scriptscriptstyle~~~~}$ \\
$603_{\scriptscriptstyle~~~~}$ & $210_{\scriptscriptstyle~~~~}$ & $447_{\scriptscriptstyle~~~~}$ \\
$618_{\scriptscriptstyle~~~~}$ & $214_{\scriptscriptstyle~~~~}$ & ${260}_{\scriptscriptstyle~~~~}$ \\
$\underline{649}_{\scriptscriptstyle~~~~}$ & ${217}_{\scriptscriptstyle~~~~}$ & $294_{\scriptscriptstyle~~~~}$ \\
$643_{\scriptscriptstyle~~~~}$ & $214_{\scriptscriptstyle~~~~}$ & $306_{\scriptscriptstyle~~~~}$ \\
\cmidrule(){1-3}
${638}_{\scriptscriptstyle\pm1}$ & $\underline{220}{}_{\scriptscriptstyle\pm1}$ & $\underline{158}{}_{\scriptscriptstyle\pm1}$ \\
$\textbf{653}_{\scriptscriptstyle\pm1}$ & $\textbf{224}_{\scriptscriptstyle\pm1}$ & $\textbf{147}_{\scriptscriptstyle\pm1}$ \\
\bottomrule 
\end{tabular} &  

\begin{tabular}{@{}C{0.90cm}@{}C{1.cm}@{}C{0.95cm}@{}}
\multicolumn{3}{@{}c@{}}{(b) KITTI ({\tiny $92\stimes310$})} \\
\toprule
{\footnotesize SSIM $\uparrow$} & {\footnotesize PSNR $\uparrow$} & {\footnotesize LPIPS $\downarrow$} \\
\cmidrule(){1-3}
$329_{\scriptscriptstyle~~~~}$ & $127_{\scriptscriptstyle~~~~}$ & $594_{\scriptscriptstyle~~~~}$ \\
$336_{\scriptscriptstyle~~~~}$ & $134_{\scriptscriptstyle~~~~}$ & $635_{\scriptscriptstyle~~~~}$ \\
${379}_{\scriptscriptstyle~~~~}$ & ${142}_{\scriptscriptstyle~~~~}$ & ${372}_{\scriptscriptstyle~~~~}$ \\
${337}_{\scriptscriptstyle~~~~}$ & ${135}_{\scriptscriptstyle~~~~}$ & $537_{\scriptscriptstyle~~~~}$ \\
${383}_{\scriptscriptstyle~~~~}$ & ${143}_{\scriptscriptstyle~~~~}$ & $501_{\scriptscriptstyle~~~~}$ \\
\cmidrule(){1-3}
$\underline{410}{}_{\scriptscriptstyle\pm1}$ & $\underline{145}{}_{\scriptscriptstyle\pm1}$ & $\underline{348}{}_{\scriptscriptstyle\pm1}$ \\
$\textbf{418}_{\scriptscriptstyle\pm2}$ & $\textbf{147}_{\scriptscriptstyle\pm1}$ & $\textbf{340}_{\scriptscriptstyle\pm2}$ \\
\bottomrule 
\end{tabular}\\

\end{tabular}

\label{tab:sota2}
\end{table}

\begin{figure}
	\setlength\tabcolsep{2pt}
	\renewcommand{\arraystretch}{0.8}
	\small
    \centering
	\begin{tabular}{C{0.25cm}cc}
 		
 	      \raisebox{2.1\normalbaselineskip}[0pt][0pt]{\rotatebox[origin=c]{90}{SLAMP~\cite{akan2021slamp}}} &
 		\includegraphics[width=.45\linewidth]{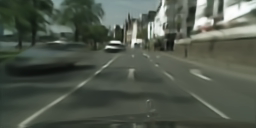} &
 		\includegraphics[width=.45\linewidth]{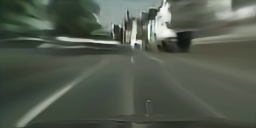} \\

        \raisebox{2.1\normalbaselineskip}[0pt][0pt]{\rotatebox[origin=c]{90}{WALDO}} &
 		\includegraphics[width=.45\linewidth]{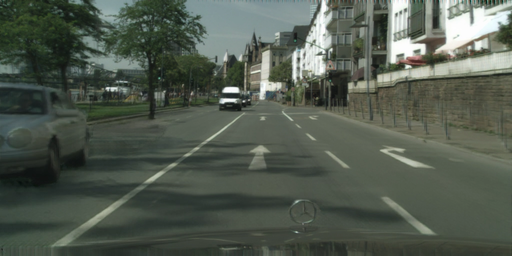} &
 		\includegraphics[width=.45\linewidth]{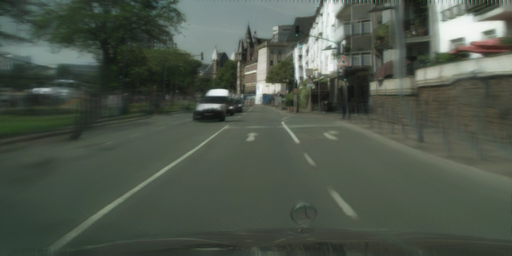} \\[-0.1em]
 		
 		& $T+10$ & $T+50$ \\ %
		
	\end{tabular}
	\caption{\change{Comparison to SLAMP~\cite{akan2021slamp} on 50-frame prediction on the Cityscapes test set.} See \proj{} for videos.} %
	\label{fig:long}
\end{figure}

\begin{figure*}
\small
\setlength\tabcolsep{0.7pt}
\renewcommand{\arraystretch}{0.5}
\begin{tabular}{@{}cccccC{0.2em}ccccc@{}}
\multicolumn{5}{@{}c@{}}{(a) UCF-Sports ({\tiny $512\stimes512$})} & & \multicolumn{5}{@{}c@{}}{(a) H3.6M ({\tiny $1024\stimes1024$})} \\ \cline{1-5}\cline{7-11}
\multicolumn{2}{@{}|c|@{}}{\raisebox{-0.5em}[0pt][0pt]{STRPM~\cite{chang2022strpm}}} & \multicolumn{3}{@{}c|@{}}{\raisebox{-0.5em}[0pt][0pt]{WALDO}} & & \multicolumn{2}{@{}|c|@{}}{\raisebox{-0.5em}[0pt][0pt]{STRPM~\cite{chang2022strpm}}} & \multicolumn{3}{@{}c|@{}}{\raisebox{-0.5em}[0pt][0pt]{WALDO}} \\
\\ %
\croppedgraphicsucf{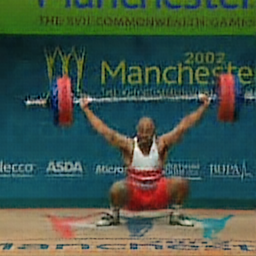} & \croppedgraphicsucf{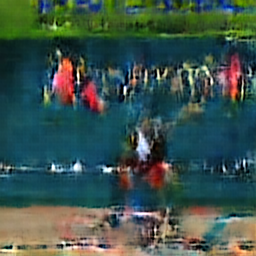} & \croppedgraphicsucf{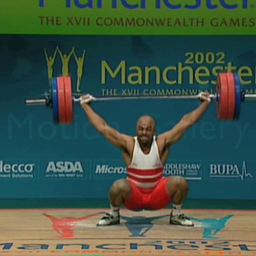} & \croppedgraphicsucf{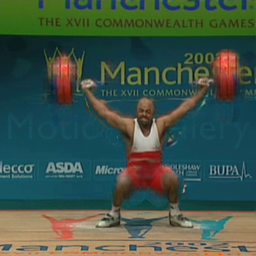} & \croppedgraphicsucf{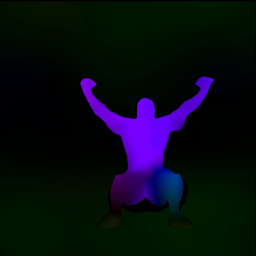} & 

& \croppedgraphichmon{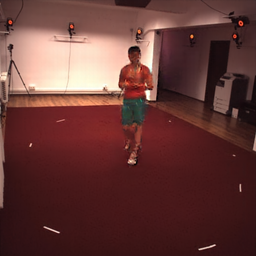} & \croppedgraphichmon{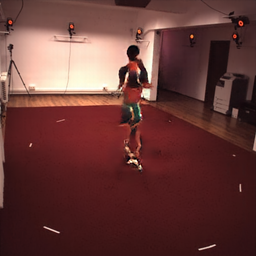} & \croppedgraphichmon{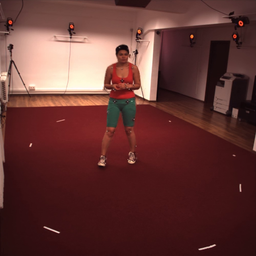} & \croppedgraphichmon{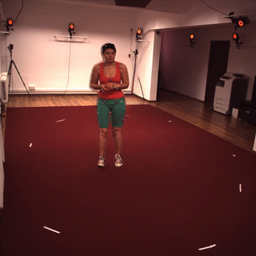} & \croppedgraphichmon{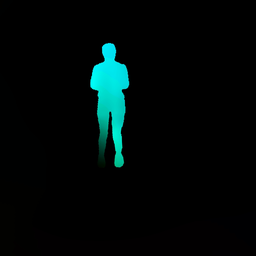} \\

\croppedgraphicsucf{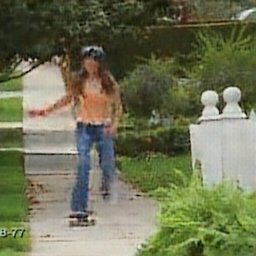} & \croppedgraphicsucf{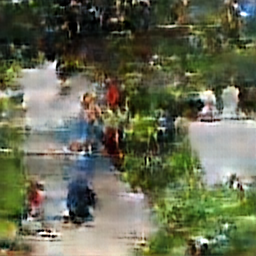} & \croppedgraphicsucf{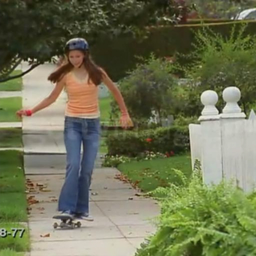} & \croppedgraphicsucf{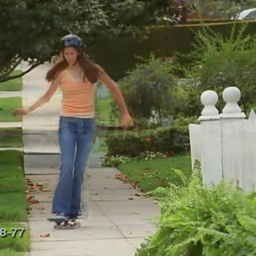} & \croppedgraphicsucf{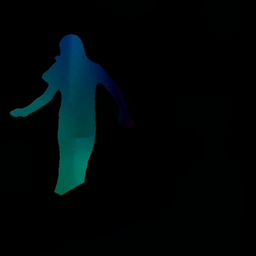} & 

& \croppedgraphichmtw{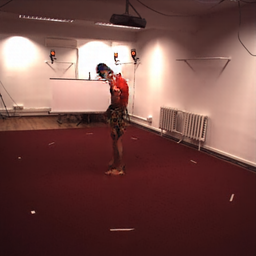} & \croppedgraphichmtw{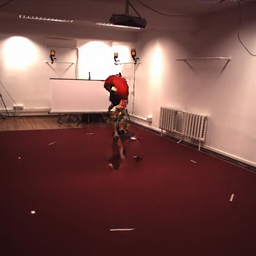} & \croppedgraphichmtw{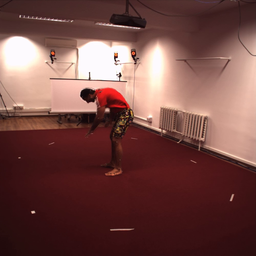} & \croppedgraphichmtw{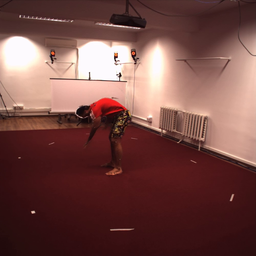} & \croppedgraphichmtw{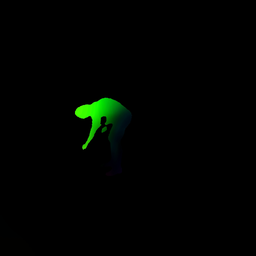} \\

\croppedgraphicsucfz{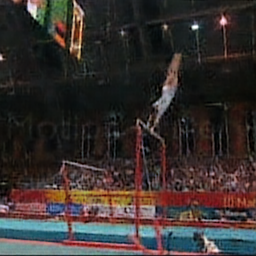} & \croppedgraphicsucfz{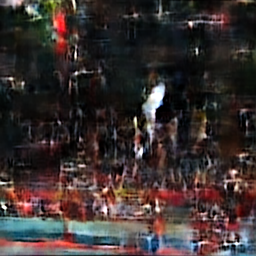} & \croppedgraphicsucfz{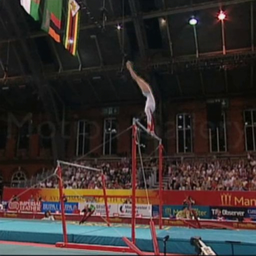} & \croppedgraphicsucfz{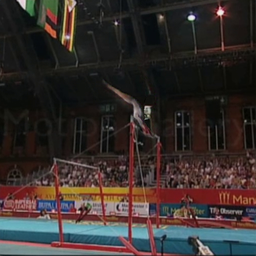} & \croppedgraphicsucfz{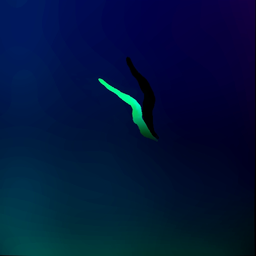} & 

& \croppedgraphichmth{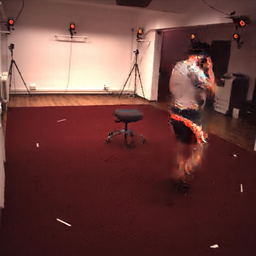} & \croppedgraphichmth{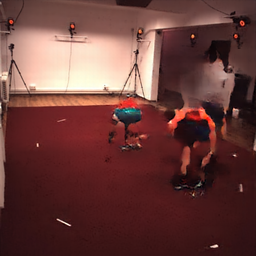} & \croppedgraphichmth{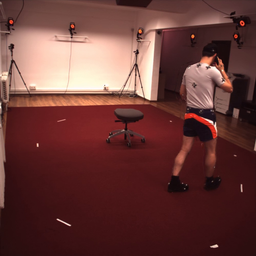} & \croppedgraphichmth{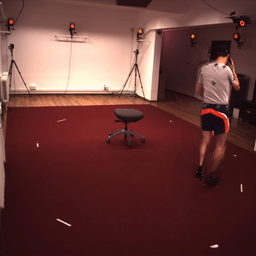} & \croppedgraphichmth{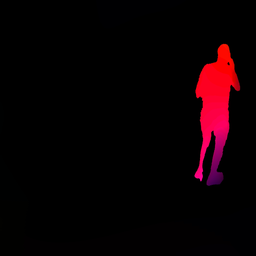} \\

{$T{+}1$} & {$T{+}10$} & {$T{+}1$} & {$T{+}10$} & {Warp} & & {$T{+}1$} & {$T{+}10$} & {$T{+}1$} & {$T{+}10$} & {Warp} \\

\end{tabular}   

\caption{\blue{Future prediction comparisons to STRPM~\cite{chang2022strpm} from $T{=}4$ frames on (a) UCF-Sports and (b) H3.6M. In our case, nonrigid motions can be visualized by the associated warps, predicted from the control points between $T$ and $T{+}10$ (colors represent different directions).}} %
\label{fig:nonrigid}
\end{figure*}

\myparagraph{Implementation details.} %
For reproducibility, code and pretrained models are available on our project webpage.\footnote{\url{https://16lemoing.github.io/waldo}} %
WALDO is trained with the ADAM optimizer~\cite{kingma2014adam} and a learning rate of $10^{-4}$ on $4$ NVIDIA V100 GPUs for about a week. %
We set $(\lambda_o,\lambda_f,\lambda_r,\lambda_p,\lambda_v){=}(1, 100, 1, 1, 1)$, $(k_s,k_m){=}(0.25,1)$ and $\tau_m{=}0.005$ by validating the performance on a random held-out subset of the training data. 
For example, we use spatial resolutions of $128 \stimes 256$ %
for training (a) the layered video decomposition, and (b) the future layer prediction module; and $512 \stimes 1024$ %
for training (c) the compositing/inpainting module on Cityscapes. %
As in prior works~\cite{geng2022comparing, bei2021learning, wu2020future, wang2018video}, we use a context of $T{=}4$ past frames.
We set the feature vector size to $d=512$. %
In (a), the encoder extracts one such vector for each $16\stimes 16$ patch from the input. %
The grids $g^i$ of resolution $4\stimes 4$ for each of the $N{=}16$ object layers and $8 \stimes 16$ for the background add up to a total of $N_c{=}384$ control points.
More details are in \app{sec:arch}.%

\subsection{Evaluation with the state of the art}

\myparagraph{Deterministic prediction.}
WALDO sets a new state of the art on two urban datasets. %
On Cityscapes (Table~\ref{tab:sota}{\small (}a{\small )}), it yields a better than 27\% relative gain for LPIPS across all predicted time windows, 
and a significant margin for $K{>}1$ with SSIM and FVD. %
On KITTI (Table~\ref{tab:sota}{\small (}b{\small )}), WALDO outperforms prior methods in a 
quite challenging setting, with a frame rate half that of Cityscapes and a quality of precomputed semantic and motion cues poorer on KITTI. %
Comparisons with latest methods~\cite{wu2020future, geng2022comparing, wu2022optimizing}\footnote{\label{fn:note1}best performing method(s) in the corresponding benchmark(s) with predicted samples available, or pretrained checkpoints to synthesize them.} on Cityscapes (Figure~\ref{fig:qual}) %
show that WALDO successfully models complex object (\eg, cars, bikes) and background (\eg, road markings) motions with realistic and temporally coherent outputs, whereas others %
produce stationary or blurry videos.

\begin{table}
\setlength\heavyrulewidth{.25ex}
\aboverulesep=0ex
\belowrulesep=.3ex
\centering
\caption{Comparison with methods designed for nonrigid motions. We compute PSNR ({\tiny$\times 10$}), LPIPS ({\tiny$\times 10^3$}) for the $k\textsuperscript{th}$ future frames synthesized from 4 past ones on UCF-Sports and H3.6M test sets.} %
\small

\begin{tabular}{@{}c@{\hskip 0.40em}c@{\hskip 0.40em}c@{}}

\begin{tabular}{@{}L{1.8cm}@{}}
~ \\
\toprule
\multirow{2}{*}{\normalsize Method} \\[1.4pt]
\\
\cmidrule(){1-1}
BMSE~\cite{mathieu2016deep} \\
PRNN~\cite{wang2017predrnn} \\
PRNN++~\cite{wang2018predrnn} \\
SAVP~\cite{lee2018stochastic} \\
SV2P~\cite{babaeizadeh2018stochastic} \\
HFVP~\cite{villegas2019high} \\
ELSTM~\cite{wang2019eidetic} \\
CGAN~\cite{kwon2019predicting} \\
CrevNet~\cite{yu2020efficient} \\
MRNN~\cite{wu2021motionrnn} \\
STRPM~\cite{chang2022strpm} \\
\midrule
WALDO \\
\bottomrule 
\end{tabular} &

\begin{tabular}{@{}L{0.8cm}@{}L{0.8cm}@{}L{0.8cm}@{}L{0.7cm}@{}}
\multicolumn{4}{@{}c@{}}{(a) UCF-Sports ({\tiny $512\stimes512$})} \\
\toprule
\multicolumn{2}{@{}c@{}}{PSNR $\uparrow$} & \multicolumn{2}{@{}c@{}}{LPIPS $\downarrow$} \\
\cmidrule(r){1-2}\cmidrule(){3-4}
{\footnotesize $k{=}1$} & {\footnotesize $k{=}6$} & {\footnotesize $k{=}1$} & {\footnotesize $k{=}6$}\\
\midrule
${264}$ &  ${185}$ & 
${290}$ &
${553}$ \\
${272}$ &
${197}$ &
${281}$ &  ${553}$ \\
${273}$ &
${197}$ &
${268}$ &  ${568}$ \\
${274}$ &
${199}$ &
${255}$ &  ${499}$ \\
${274}$ &
${200}$ &
${259}$ &  ${513}$ \\
~~- & ~~- & ~~- & ~~- \\  
${280}$ &
${203}$ &
${251}$ & ${478}$ \\
${280}$ & 
${200}$ &
${229}$ &  ${449}$ \\
${282}$ &
${203}$ &
${239}$ &  ${481}$ \\
${277}$ & 
${200}$ & ${242}$ & ${492}$ \\
$\underline{285}$ &
$\underline{206}$ &
$\underline{207}$ &  $\underline{411}$ \\
\midrule
$\textbf{292}_{\scriptscriptstyle\pm2}$ & $\textbf{235}_{\scriptscriptstyle\pm1}$ & $\textbf{090}_{\scriptscriptstyle\pm1}$ & $\textbf{183}_{\scriptscriptstyle\pm1}$ \\
\bottomrule 
\end{tabular} &  

\begin{tabular}{@{}L{0.775cm}@{}L{0.775cm}@{}L{0.775cm}@{}L{0.775cm}@{}}
\multicolumn{4}{@{}c@{}}{(b) H3.6M ({\tiny $1024\stimes1024$})} \\
\toprule
\multicolumn{2}{@{}c@{}}{PSNR $\uparrow$} & \multicolumn{2}{@{}c@{}}{LPIPS $\downarrow$} \\
\cmidrule(r){1-2}\cmidrule(){3-4}
{\footnotesize $k{=}1$} & {\footnotesize $k{=}4$} & {\footnotesize $k{=}1$} & {\footnotesize $k{=}4$}\\
\midrule
~~- & ~~- & ~~- & ~~- \\
$319$ & $257$ & $126$ & $140$ \\
$321$ & $275$ & $138$ & $150$ \\
~~- & ~~- & ~~- & ~~- \\
$319$ & $273$ & $139$ & $150$ \\
$321$ & $273$ & $134$ & $145$ \\
$324$ & $277$ & $131$ & $139$ \\
$328$ & $283$ & $102$ & $110$ \\
$332$ & $283$ & $115$ & $124$ \\
$322$ & $280$ & $121$ & $133$ \\
$\underline{333}$ & $\underline{290}$ & $\underline{097}$ & $\underline{104}$ \\
\midrule
$\textbf{363}_{\scriptscriptstyle\pm3}$ & $\textbf{314}_{\scriptscriptstyle\pm1}$ & $\textbf{058}_{\scriptscriptstyle\pm2}$ & $\textbf{071}_{\scriptscriptstyle\pm1}$ \\
\bottomrule 
\end{tabular} 

\end{tabular}

\label{tab:sota3}
\end{table}

\myparagraph{Stochastic prediction.}
We adapt WALDO to the prediction of multiple futures by injecting noise inputs to the layer prediction module and using (at train time only) a discriminator to capture
multiple modes of the distribution of synthetic trajectories.
This simple procedure results in significant improvements in the stochastic setting (Table~\ref{tab:sota2}). 
In particular, WALDO outperforms other stochastic methods on the same datasets by a large margin.
Interestingly, even our deterministic variant compares favourably to these approaches.
Minimizing the $L_1$ reconstruction error (in this variant) make predictions average over all possible futures.
Because we predict positions and not pixel values directly, and since averaging still yields valid positions, we obtain sharp images.
On the other hand, averaging over pixels inevitably blur them out.
This critical distinction is illustrated in Table~\ref{tab:abl}(e). Although the loss function introduced in VPCL~\cite{geng2022comparing} aims at sharpening pixel predictions, the examples in Figure~\ref{fig:qual} show that WALDO is more successful. %

\myparagraph{Long-term prediction.}
WALDO produces arbitrary long videos, without additional training, when used in an autoregressive mode.
In $20$-frame prediction (Table~\ref{tab:sota2}), %
it significantly outperforms prior works, \revision{without using its full potential since a lower resolution is used to match those of other methods}.
Visual comparisons with SLAMP~\cite{akan2021slamp}\textsuperscript{\ref{fn:note1}} (Figure~\ref{fig:long}) \revision{at full resolution} on 50-frame prediction \revision{(longer than the 30-frame videos in Cityscapes)} are also striking. %

\begin{figure*}
    \centering
    \begin{tabular}{@{}cc@{}}
         \includegraphics[width=0.48\linewidth]{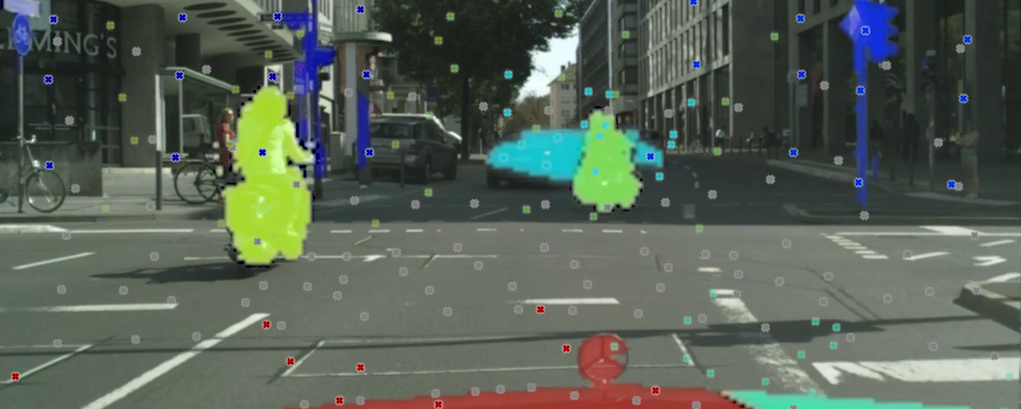} & \includegraphics[width=0.48\linewidth]{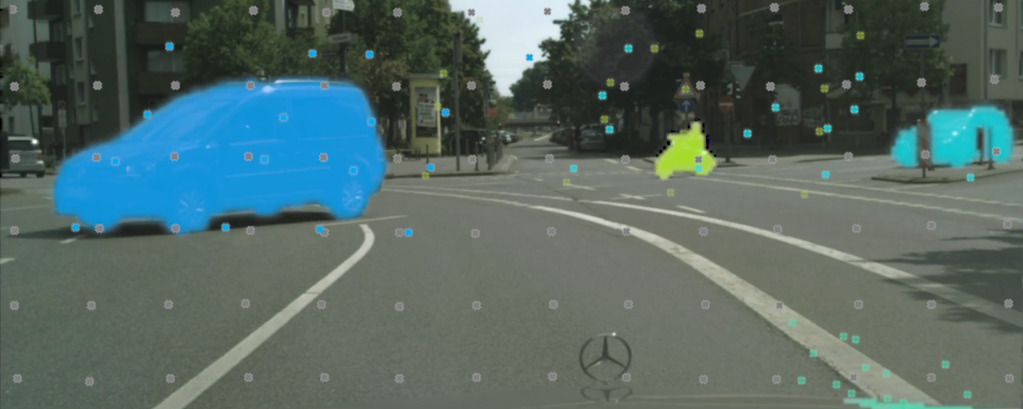} \\
         \includegraphics[width=0.48\linewidth]{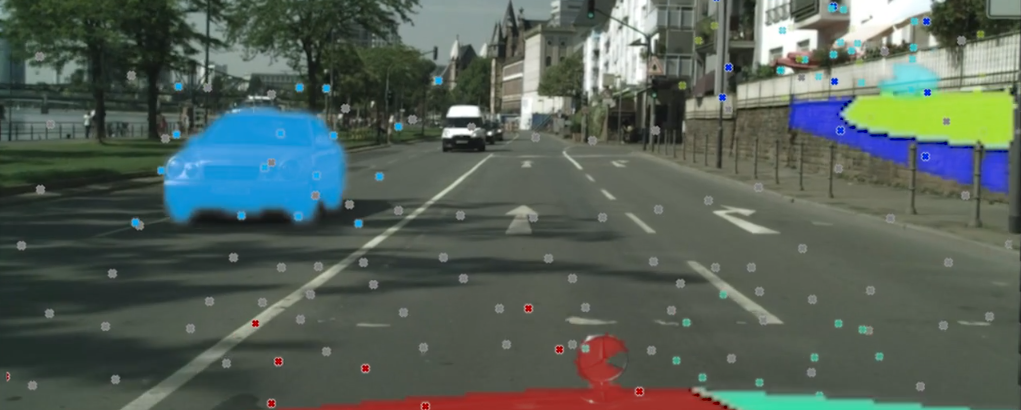} & 
         \includegraphics[width=0.48\linewidth]{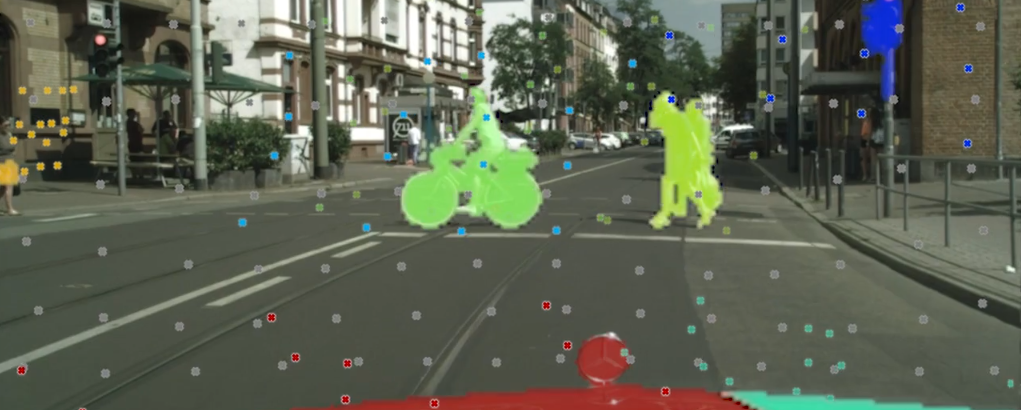} \\
    \end{tabular}
    \caption{\new{Visualization of control points and layer masks with different colors for each layer. See project webpage for videos.}}
    \label{fig:dec}
\end{figure*}
\begin{table}
\setlength\heavyrulewidth{.25ex}
\aboverulesep=0ex
\belowrulesep=.3ex
\centering
\caption{Ablation study on the Cityscapes test set. (a)~{\bf Layered video decomposition}: We evaluate decompositions in terms of flow reconstruction and object discovery as captured by $\mathcal{L}_{f}$ and $\mathcal{L}_{o}$. (b)~{\bf Future layer prediction}: We measure the accuracy of predicted control points. (c)~{\bf Video synthesis}: We evaluate the image reconstruction quality. \revision{(d)~An example without (left) and with (right) semantic refinement. (e)~$L_1$ reconstructions of an image in pixel space (left) and using control points (right) assuming Gaussian uncertainty
over the horizontal position of the object. 
The former is blurry due to the plurality of potential positions. The latter singles out one position and preserves the object appearance.}} %
\small

\begin{tabular}{@{}c@{\quad}c@{}}

\begin{tabular}{@{}c@{}}
\begin{tabular}{@{}C{.40cm}@{}C{.90cm}@{}C{.60cm}@{}C{.60cm}@{}C{0.90cm}@{}C{0.90cm}@{}}
\multicolumn{6}{@{}c@{}}{(a) Layered video decomposition.} \\
\toprule
$N$ & Input & Ref. & $N_c$ & $\mathcal{L}_{f}$ $\downarrow$ & $\mathcal{L}_{o}$ $\downarrow$ \\
\midrule
0 & $X$ & & 128 & 4.42 & ~0.00 \\
1 & $X$ & & 144 & 6.06 & -3.69 \\
8 &  $X$ & & 256 & 4.47 & -5.96 \\
16 &  $X$ & & 384 & 3.97 & -5.79 \\
16 &  $S$ & & 384 & 3.90 & -7.59 \\
16 &  $S{+}F$ & & 384 & \underline{2.69} & {-7.56} \\
16 &  $S{+}F$ & \checkmark & 24 & {7.44} & {~0.00} \\
16 &  $S{+}F$ & \checkmark & 72 & {7.01} & {-3.26} \\
16 &  $S{+}F$ & \checkmark & 176 & {4.17} & {-4.78} \\
16 &  $S{+}F$ & \checkmark & 288 & {3.16} & \underline{-7.93} \\
\midrule
16 &  $S{+}F$ & \checkmark & 384 & \textbf{2.59} & \textbf{-8.16} \\
\bottomrule
\end{tabular} \\

\end{tabular} &

\begin{tabular}{@{}c@{}}
\begin{tabular}{@{}C{.90cm}@{}C{.90cm}@{}C{0.60cm}@{}C{1.10cm}@{}}
\multicolumn{4}{@{}c@{}}{(b) Future layer prediction.} \\
\toprule
Arch. & Input & $\Delta p_T$ & $\mathcal{L}_{p}$ $\downarrow$ \\
\midrule
MLP & $P$ & & .514 \\
T & $P$ & & .178 \\
T & $P$ & \checkmark & \underline{.150}  \\
\midrule
T & $P{+}Z$ & \checkmark & \textbf{.144} \\
\bottomrule
\end{tabular} \\

~ \\[-0.75em]

\begin{tabular}{@{}C{.75cm}@{}C{.75cm}@{}C{.65cm}@{}C{1.25cm}@{}}
\multicolumn{4}{@{}c@{}}{(c) Video synthesis.} \\
\toprule
VGG & HR & Ctxt. & SSIM $\uparrow$ \\
\midrule
& & 1 & 812.1 \\
\checkmark & & 1 & 815.7 \\
\checkmark & \checkmark & 1 & \underline{847.2} \\
\midrule
\checkmark & \checkmark & 4 & \textbf{848.0} \\
\bottomrule
\end{tabular} \\

\end{tabular} \\

\end{tabular}
\begin{tabular}{@{}C{2.20cm}@{}C{2.20cm}@{}C{1.90cm}@{}C{1.90cm}@{}}
~ \\[-0.85em]
\multicolumn{2}{@{}c@{}}{(d) Semantic refinement.} & \multicolumn{2}{@{}c@{}}{(e) Prediction strategy.} \\
\includegraphics[width=0.12\textwidth]{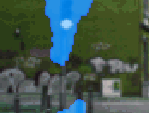} &
\includegraphics[width=0.12\textwidth]{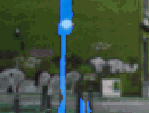} & ~~~~\includegraphics[width=0.07\textwidth]{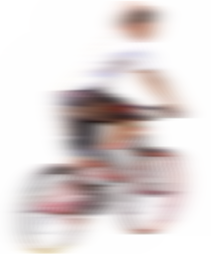} &
\includegraphics[width=0.07\textwidth]{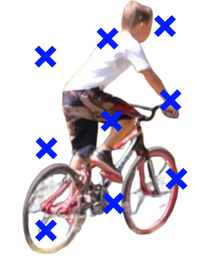}~~~~ \\
\end{tabular}
\label{tab:abl}
\end{table}

\myparagraph{Nonrigid prediction.}
To show that WALDO can handle complex motions, we retrain it on data with significant nonrigid objects, namely UCF-Sports and H3.6M datasets. 
Since our approach relies on TPS transformations, it can represent arbitrarily nonrigid motions when the number of control points is set accordingly~\cite{bookstein1989principal}. %
By increasing the total number of per object points from $16$ to $64$, WALDO produces realistic videos even for deformable bodies such as human beings (Figure~\ref{fig:nonrigid}).
As highlighted by the predicted warps, it successfully handles fine-grained motions covering a variety of human activities such as weight-lifting, doing gymnastics, walking sideways or leaning forward.
Moreover, WALDO yields much more realistic outputs than STRPM~\cite{chang2022strpm}\textsuperscript{\ref{fn:note1}} %
by producing high quality frames longer into the future and with much less synthesis artifacts.
This is confirmed by quantitative evaluations on both datasets (Table~\ref{tab:sota3}), with large relative improvements over the state of the art ranging from $1$ to $3$dB in PSNR and 31 to 56\% in LPIPS.

\begin{figure*}[h]
	\setlength\tabcolsep{1.0pt}
	\renewcommand{\arraystretch}{1.0}
	\small
	\begin{tabular}{L{0.6cm}cccccc}
	& $T+1$ & $T+10$ & $T+1$ & $T+10$ \\

 	\raisebox{2.3\normalbaselineskip}[0pt][0pt]{\rotatebox[origin=c]{90}{\parbox{4cm}{\centering Real \\ \vspace{-0.3em}frame}}} &
 	\includegraphics[width=.235\linewidth]{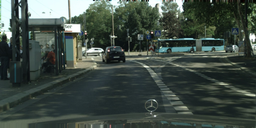} &
 	\includegraphics[width=.235\linewidth]{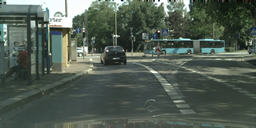} &
    \includegraphics[width=.235\linewidth]{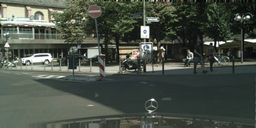} &
 	\includegraphics[width=.235\linewidth]{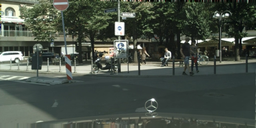} \\
 	
 	\raisebox{2.3\normalbaselineskip}[0pt][0pt]{\rotatebox[origin=c]{90}{\parbox{4cm}{\centering Reconstructed \\ \vspace{-0.3em}motion}}} &
 	\includegraphics[width=.235\linewidth]{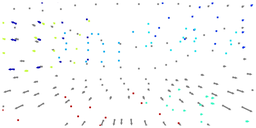} &
 	\includegraphics[width=.235\linewidth]{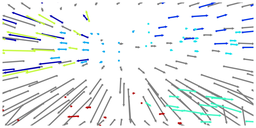} &
    \includegraphics[width=.235\linewidth]{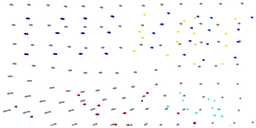} &
 	\includegraphics[width=.235\linewidth]{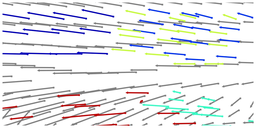} \\
 	
 	\raisebox{2.3\normalbaselineskip}[0pt][0pt]{\rotatebox[origin=c]{90}{\parbox{4cm}{\centering Predicted \\ \vspace{-0.3em}motion}}} &
 	\includegraphics[width=.235\linewidth]{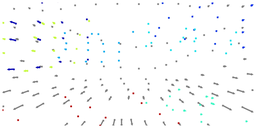} &
 	\includegraphics[width=.235\linewidth]{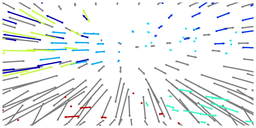} &
    \includegraphics[width=.235\linewidth]{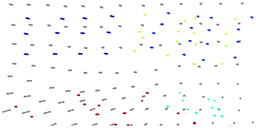} &
 	\includegraphics[width=.235\linewidth]{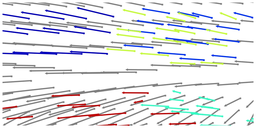} \\
    
	\end{tabular}
	\vspace{-0.5em}
	\caption{\new{Visualization of future layer prediction. We use control points from the layered video decomposition as supervision. We compare motion vectors reconstructed from these points to the ones predicted for up to time step $T+10$ from a context of $T{=}4$ past frames. The motion vectors are computed between time step $T$ and time step $t$ in $\{T+1,T+10\}$. Different colors correspond to different layers.}}
	\label{fig:flp}
\end{figure*}
\begin{figure}
	\setlength\tabcolsep{1.5pt}
	\renewcommand{\arraystretch}{1.5}
	\small
    \centering
	\begin{tabular}{C{0.5cm}cc}

        \raisebox{3.6\normalbaselineskip}[0pt][0pt]{\rotatebox[origin=c]{90}{\parbox{4cm}{\centering Warped regions}}} &
 		\croppedinp{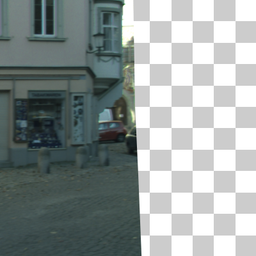} &
 		\croppedinp{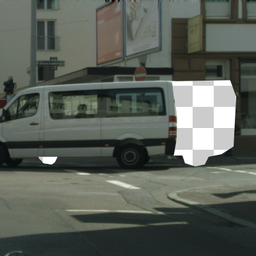} \\[-0.12cm]

        \raisebox{3.6\normalbaselineskip}[0pt][0pt]{\rotatebox[origin=c]{90}{\parbox{4cm}{\centering Inpainted regions}}} &
 		\croppedinp{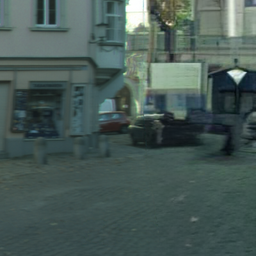} &
 		\croppedinp{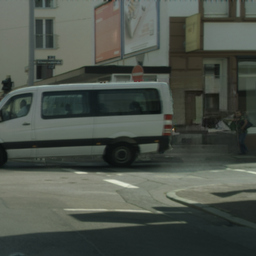} \\

	\end{tabular}
	\caption{\new{Visualization of the synthesis process: warped visible regions and inpainted disoccluded ones between $T$ and $T{+}10$.}}
	\label{fig:inpaint}
\end{figure}

\subsection{Ablation studies}
\label{sec:abl}

We conduct a detailed analysis on the Cityscapes test set to compare and highlight the key novelties of WALDO. %
The set of hyperparameters validated through this study have proven to work well on other datasets without any tuning.

\myparagraph{Layered video decomposition} 
\new{is illustrated in Figure~\ref{fig:dec}} and evaluated in Table~\ref{tab:abl}(a) through the lens of our object discovery criterion ($\mathcal{L}_{o}$) and flow reconstruction ($\mathcal{L}_{f}$).
We observe that the more object layers ($N$) the better, %
except for $N{=}1$ where fitting multiple objects with a single layer is suboptimal. %
Comparing inputs, we find using segmentation maps ($S$) better than RGB frames~($X$) alone for object discovery, and that using flow maps~($F$) helps motion reconstruction.
Semantic refinement (Ref.) yields further gains, especially to segment thin objects like traffic lights, see Table~\ref{tab:abl}(d). %
\revision{Finally, increasing the number of control points ($N_c$) allows us to capture finer motions, so we use as much as fits into memory (384 per time step). 
Our motion representation is scalable, with parameter size reduced from a quadratic ($TK$) to a linear ($T{+}K$) dependency on time compared to methods relying on optical flow directly.}%

\myparagraph{Future layer prediction}
is illustrated in Figure~\ref{fig:flp} and evaluated in Table~\ref{tab:abl}(b) %
in terms of trajectory reconstruction.
Our baseline is a multi-layer perceptron~(MLP), which maps past control points $P{=}\{p_t\}_{t=1}^T$ to future ones. %
Our actual architecture relies on a transformer~(T)~\cite{vaswani2017attention}, which performs much better. 
Further gains are obtained by \revision{not predicting control points directly but rather} their relative position~($\Delta p_T$) with respect to time step $T$, 
and by using layer features $Z{=}\{z^i\}_{i=0}^N$ as input. %

\myparagraph{Warping, fusion and inpainting} is illustrated in Figure~\ref{fig:inpaint} and evaluated in Table~\ref{tab:abl}(c) in terms of SSIM. %
Control points, obtained by the layer decomposition module, %
are used to warp, fuse and inpaint different views of the past frames to reconstruct future ones.
We find using the feature distance (VGG) to be slightly better than the pixel one alone. %
The flexibility of WALDO, which trains at low resolution ($128\stimes256$) but produces dense motions well suited for high resolution (HR) inputs ($512\stimes1024$), results in higher SSIM \revision{than keeping the same resolution than for training during inference}. 
We further improve SSIM by warping not only one but all of the four past context frames (Ctxt.).%

\myparagraph{Off-the-shelf models.}
We compare standard approaches in \app{sec:setup}, and find that WALDO is robust to the choice of the pretrained segmentation~\cite{chen2018encoder, sandler2018mobilenetv2, chen2022vitadapter} and optical flow models~\cite{teed2020raft, sun2018pwc}. %
\revision{We also show in \app{sec:inpabl} that although we use an inpainting method~\cite{li2022mat} pretrained on external data~\cite{zhou2017places} to produce realistic outputs in filled-in regions, it does not provide quantitative advantage to WALDO with at best marginal improvements in SSIM, LPIPS and FVD.} %

\section{Conclusion}
\label{sec:conclusion}

\blue{We have introduced WALDO, an approach to video synthesis which automatically decomposes frames into layers and relies on a compact representation of motion to predict their future deformations. Our method outperforms the state of the art for video prediction on various datasets. Future work includes exploring extensions of WALDO for applications from motion segmentation to video compression.}

\myparagraph{Limitations.} Our performance depends on the accuracy of the layer decomposition. Failure cases include objects moving in different directions but merged into the same layer, or segmentation failures, where parts of an object are missed.

\section*{Acknowledgements}

We thank Daniel Geng and Xinzhu Bei for clarifications on the evaluation process, and Pauline Sert for helpful feedback.
This work was granted access to the HPC resources of IDRIS under the allocation 2021-AD011012227R1 made by GENCI. It was funded in part by the French government under management of Agence Nationale de la Recherche as part of the ``Investissements d’avenir'' program, reference ANR-19-P3IA-0001 (PRAIRIE 3IA Institute), and the ANR project VideoPredict, reference ANR-21-FAI1-0002-01.
JP was supported in part by the Louis Vuitton/ENS chair in artificial intelligence and a Global Distinguished Professorship at the Courant Institute of Mathematical Sciences and the Center for Data Science at New York University.

{\small
\bibliographystyle{ieee_fullname}
\bibliography{egbib}
}

\appendix
\onecolumn

\noindent We present the following items in the Appendix: %
\begin{itemize}[itemsep=-0.15em]
    \item The thin-plate splines warp computation (Section~\ref{sec:tps-warp}) and its inverse (Section~\ref{sec:inverse-warp})
    \item The detailed formulation for the semantics-aware refinement step (Section~\ref{sec:sem-ref})
    \item The layer occlusion model (Section~\ref{sec:occ})
    \item A stochastic extension of our method (Section~\ref{sec:sto})
    \item Detailed architectural choices for each of WALDO's modules (Section~\ref{sec:arch})
    \item More qualitative samples on nonrigid scenes (Section~\ref{sec:taichi})
    \item The influence of the choice of the pretrained segmentation and optical flow models (Section~\ref{sec:setup})
    \item An ablation study of our inpainting strategy (Section~\ref{sec:inpabl})
    \item Further information about our implementation and the overall training process (Section~\ref{sec:imp})
    \item A statement about the societal impact of this project (Section~\ref{sec:state})
    \item A qualitative study of our approach (Section~\ref{sec:qual-abl})
\end{itemize}

\section{Thin-plate splines warp computation}
\label{sec:tps-warp}
\setcounter{figure}{0}
\renewcommand{\thefigure}{A\arabic{figure}}

\begin{wrapfigure}[11]{r}{8.7cm}
    \centering
    \vspace{-0.75em}
    \includegraphics[width=\linewidth]{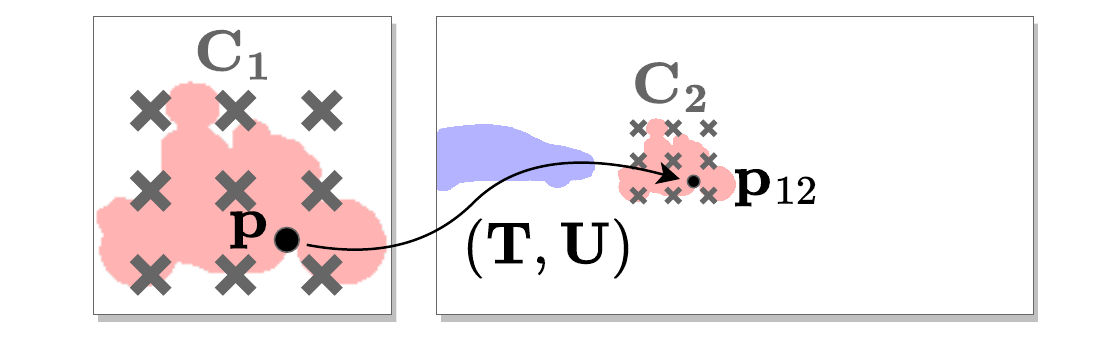}
	\caption{Thin-plate splines transformation mapping points from $\mathbf{C_1}$ onto $\mathbf{C_2}$ applied to an arbitrary point $\mathbf{p}$ using parameters ($\mathbf{T}$, $\mathbf{U}$).} %
	\label{fig:tps}
\end{wrapfigure}

For clarity, vectors (or points) are denoted by bold face lower case letters, and matrices are denoted by bold face upper case letters throughout this presentation.
Let $\mathbf{p}=(x,y,1)^\intercal$ be a point in homogeneous coordinates, and $\mathbf{C_1}$ and $\mathbf{C_2}$ be two sets of such points in $\RR^{3\times L}$, which we refer to as \emph{control points}, with $L$ the (fixed) number of points in each set.
The thin-plate splines (TPS)~\citesupp{bookstein1989principal} transformation which maps $\mathbf{p}$ onto $\mathbf{p_{12}}$ writes: %
\begin{equation}
\label{eq:tps}
    \mathbf{p_{12}}=\mathbf{T}\mathbf{p}+\mathbf{U}\bm{\phi}(\mathbf{p},\mathbf{C_1}),
\end{equation}
where $\bm{\phi}(\mathbf{p},\mathbf{C_1})=[k(\mathbf{p}, \mathbf{p_1})]_{\mathbf{p_1}\in \mathbf{C_1}}$ is a $L$-dimensional vector, and ($\mathbf{T}$, $\mathbf{U}$) are TPS parameters in the form of matrices of dimension $3\stimes 3$ and $3 \stimes L$ respectively.
The transformation is decomposed into a global affine one (through $\mathbf{T}$) and a local non-affine one (through $\mathbf{U}$ and $\bm{\phi}$).
The kernel function $k$ is defined by $k:(\mathbf{p},\mathbf{q})\rightarrow\|\mathbf{p}-\mathbf{q}\|_2^2\log\|\mathbf{p}-\mathbf{q}\|_2$ and materializes the fact that the TPS transformation minimizes the bending energy.
The $3(3+L)$ TPS parameters are found using the constraint that points from $\mathbf{C_1}$ should be mapped onto points from $\mathbf{C_2}$ using (\ref{eq:tps}). This yields a system of $3L$ equations to which Bookstein adds $9$ extra ones, as explained in~\citesupp{bookstein1989principal}, which we formulate as:
\begin{equation}
\label{eq:extra}
    \mathbf{C_1}\mathbf{U}^\intercal=\mathbf{0_{3\times 3}}.
\end{equation}
By applying (\ref{eq:tps}) on pairs of points from $\mathbf{C_1}$ and $\mathbf{C_2}$, and using extra constraints~(\ref{eq:extra}), we obtain the TPS parameters ($\mathbf{T}$, $\mathbf{U}$):
\begin{equation}
\label{eq:param}
\begin{bmatrix}
\mathbf{T}^\intercal \\
\mathbf{U}^\intercal
\end{bmatrix}=
\bm{\Delta}(\mathbf{C_1})^{-1}
\begin{bmatrix}
\mathbf{C_2}^\intercal \\
\mathbf{0_{3\times3}}
\end{bmatrix}, \,\,\,\,\,\,\,\,\,\, \bm{\Delta}(\mathbf{C_1})=
\begin{bmatrix}
\mathbf{C_1}^\intercal & \bm{\Phi}(\mathbf{C_1},\mathbf{C_1}) \\
\mathbf{0_{3\times3}} & \mathbf{C_1}
\end{bmatrix},
\end{equation}
where $\bm{\Phi}(\mathbf{C_1},\mathbf{C_1})$ is a $L \stimes L$ matrix obtained by stacking $\bm{\phi}(\mathbf{p_1},\mathbf{C_1})$ for all points $\mathbf{p_1}$ in $\mathbf{C_1}$.

Hence, computing the TPS parameters amounts to inverting a $(L+3)\stimes(L+3)$ matrix, $\bm{\Delta}(\mathbf{C_1})$.
Since doing that for each new transformation is impractical, we use a fixed grid of control points associated with each layer for $\mathbf{C_1}$, as a proxy to compute the flow between pair of frames.
By fixing the value of $\mathbf{C_1}$, the corresponding matrix inversion is done only once for the whole training; and we are still able to model different deformations by setting $\mathbf{C_2}$ to different values.
We note that, once $\bm{\Delta}(\mathbf{C_1})^{-1}$ has been computed, sampling the warp $w$ associated with a new $\mathbf{C_2}$ is done by finding the corresponding TPS parameterization using (\ref{eq:tps}), and by sampling the deformations for each point $\mathbf{p}$ in a dense grid $\llbracket 1, H \rrbracket \stimes \llbracket 1, W \rrbracket$ using (\ref{eq:param}), where the spatial resolution $H \stimes W$ can be arbitrarily large.
Moreover, this process is fully differentiable with respect to $\mathbf{C_2}$ and each point $\mathbf{p}$ as it involves simple algebraic operations.
In the main paper, $\mathbf{C_1}$ and $\mathbf{C_2}$ correspond to the regular grid of control points $g^i$, associated with layer $i$, and its deformation $p_t^i$ at time step $t$ respectively. The warp $w_t^i$, also in the main paper, corresponds to the inverse transformation to the one presented here, and associates with every point $\mathbf{p_{12}}$ corresponding to a pixel of the image (right side in Figure~\ref{fig:tps}) the corresponding point $\mathbf{p}$ in object coordinates (left side).
However, computing the inverse transformation cannot be achieved by simply switching the roles of $\mathbf{C_1}$ and $\mathbf{C_2}$, since $\mathbf{C_1}$ have to be kept constant, which is why we resort to warp inversion whose simple formulation is detailed in the next section.

\section{Inverse warp computation}
\label{sec:inverse-warp}

Let $w$ be a warp in $\RR^{2 \times H \times W}$ (where $H \stimes W$ is a given spatial resolution), which we could also consider as a mapping from $\RR^2$ to $\RR^2$, where $w(p)$ is the geometric transformation of a point $p$ sampled on the grid $\llbracket 1, H \rrbracket \stimes \llbracket 1, W \rrbracket$.
Such a mapping is not surjective with respect to the grid, that is, all the cells in the grid are not necessarily reached by $w$.
As a result, we approximate the inverse warp $w^{-1}$ by the pixel-accurate inversion in cells for which such a direct mapping exists and use interpolation for filling others, starting from the cardinal neighbours of already inverted cells, and iteratively filling the remaining ones.
The warp $w^{-1}$ may need to point out of the grid for some cells, \eg, when an object moves out of the frame.
However, this cannot be extracted from $w$ which is only defined on the grid.
To avoid interpolating wrong values in these cells, we only invert $w$ in those which are close (as per a given threshold) to the ones initially reached by $w$, and make others point to an arbitrary position outside of the grid layout. Finally, the same reasons which justify that training errors can backpropagate through a spatial transformer~\citesupp{jaderberg2015spatial} also apply here.

\section{Semantics-aware refinement}
\label{sec:sem-ref}

We now describe in more details the semantics-refinement step introduced in the main body of the paper. Let $m_t$ be a soft mask in $[0,1]^{H\times W}$ at a given time step $t$ in $\llbracket 1, T\rrbracket$, and $c$ be a soft class assignment in $[0,1]^C$, both of them predicted for the same layer, and let $s_t$ be an input (soft) semantic map in $[0,1]^{C\times H\times W}$ also associated with time step $t$.
We denote by $\bar{c}$, the mean semantic class on the spatio-temporal tube defined by the masks, which, like $c$, is a vector in $[0,1]^C$, and writes:
\begin{equation}
    \bar{c}=\sum_{t,h,w} [m_{t} \odot \mathcal{F}(s_t, c)]_{(h,w)}/\sum_{t,h,w} [m_{t}]_{(h,w)},
\end{equation}
where $\odot$ is the element-wise product, and $\mathcal{F}$ is a class-filtering function parameterized by $c$ and applied to $s_t$, that is, at a given spatial location $(h,w)$, the result of kipping dominant classes as per $c$ in $s_t$:
\begin{equation}
    [\mathcal{F}(s_t, c)]_{(h,w)}=\frac{1}{1+k_c}\sum_{j}\langle[s_t]_{(h,w)},c+k_c\rangle,
\end{equation}
with $k_c$ a constant which defines the degree at which low scoring classes will be filtered out. %
One can set $k_c=0$ for full effect and greater values for filtering less.
In practice, we fix the value of $k_c$ to $0.1$.
Each mask $m_{t}$ is updated by computing the $L_1$ distance between the semantic map $s_t$ and the mean class $\bar{c}$ at every location $(h,w)$:
\begin{equation}
    [m_{t}]_{(h,w)}=(1-\|[s_t]_{(h,w)}-\bar{c}\|_1)[m_{t}]_{(h,w)}.
\end{equation}

\section{Layer occlusion model}
\label{sec:occ}

Our occlusion model is rather standard, but we include it here for completeness.
Ordering scores $o_t$ are used to filter non-visible parts in a layer $i$ due to the presence of another layer $j$ on top of it (\ie, when $o_t^i \ll o_t^j$). The transparency $m_t^i$ of layer $i$ at time step $t$ is updated as follows: %
\begin{equation}
\label{eq:ord}
    m^i_t=m^i_t\odot\prod_{j\neq i}\Big(\mathbf{1}-\frac{o_t^j}{o_t^i+o_t^j}m_t^j\Big), %
\end{equation}
where $\odot$ is the element-wise product whose right-hand side component has values between $0$ (occluded) and $1$ (visible).

\section{Stochastic extension of WALDO}
\label{sec:sto}
\setcounter{figure}{0}
\renewcommand{\thefigure}{E\arabic{figure}}

\begin{figure}[h]
    \centering
    \includegraphics[width=0.65\linewidth]{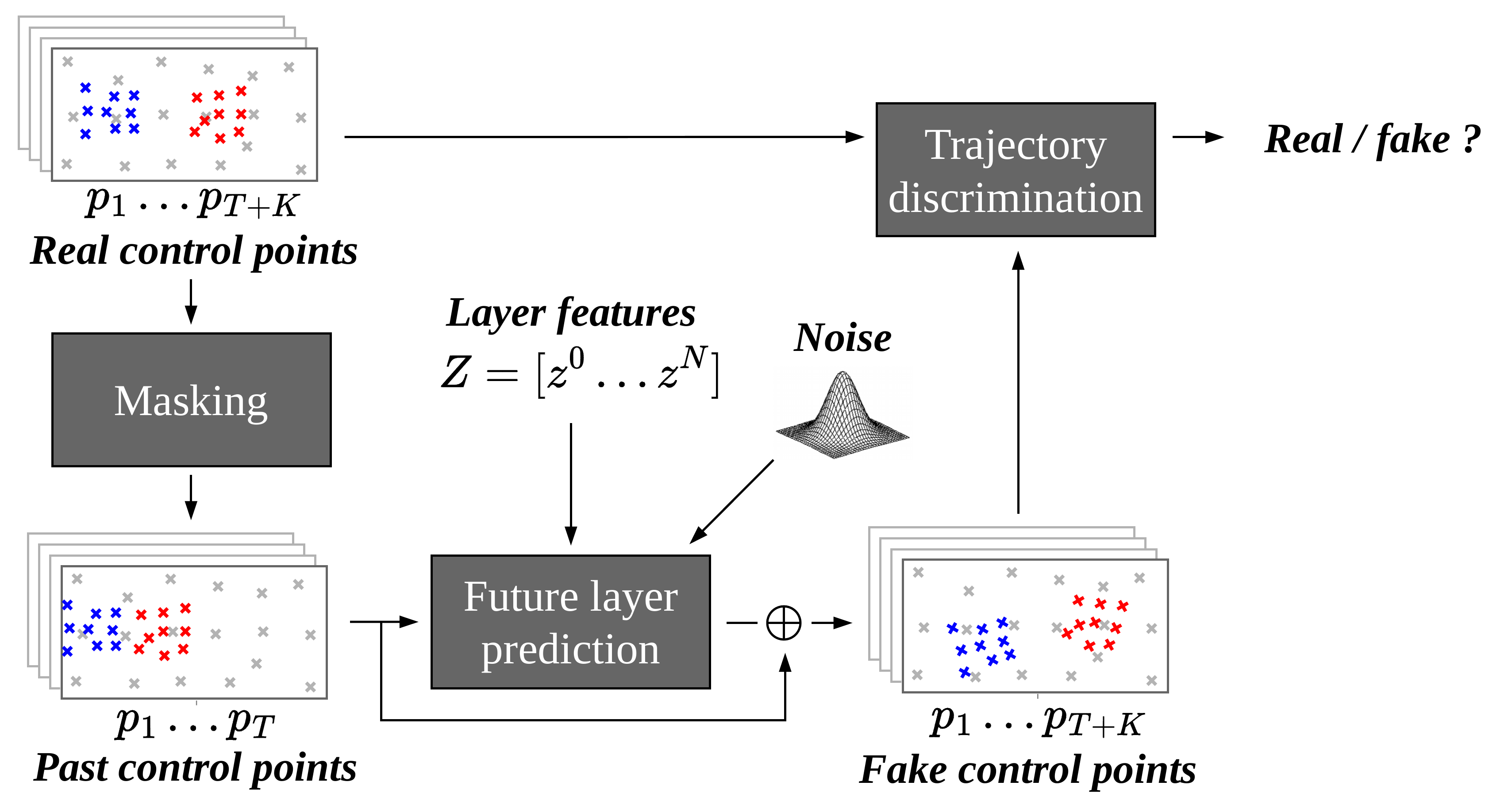}
	\caption{\textbf{Stochastic future prediction.} We extend the future layer prediction module of WALDO to allow the prediction of multiple futures. The module itself is not changed, except that it now uses noise (as input and in attention modules). The major difference is that a \emph{trajectory discrimination} module is used at train time to assist the model in producing \underline{a} realistic control point trajectory instead of \underline{the} mean trajectory. At inference, only the future layer prediction module is kept. See text for details.} %
	\label{fig:sto}
\end{figure}

In our original description, motions produced by WALDO are purely deterministic and they converge towards the mean of all possible future trajectories given the past. Although it is sufficient for short-term prediction, one could be interested in modeling different behaviours for longer future horizons. For this reason, we propose an extension of WALDO to account for the intrinsically uncertain nature of the future by allowing multiple predictions.

Our approach, illustrated in Figure~\ref{fig:sto}, builds upon generative adversarial networks (GANs)~\citesupp{goodfellow2014generative}. We consider the future layer prediction module as a generator which computes the future positions of control points given ones from the past. This generator is trained jointly with a discriminator which classifies trajectories as \emph{real} or \emph{fake}. Both are playing a minimax game, where the discriminator is taught to correctly distinguish real from fake trajectories, while the generator tries to fool the discriminator. Through this process we expect synthetic trajectories to gradually improve in realism and to capture multiple modes from the underlying data distribution.

We implement the discriminator with a transformer and use the WGAN loss introduced in~\citesupp{arjovsky2017wasserstein} for training. In addition to this loss, we find that keeping the initial reconstruction term ($\mathcal{L}_{{p}}$) is important for the stability of training. Moreover, we normalize gradients from both supervision signals (reconstruction and adversarial) so that they have matching contributions.

\section{Detailed architectures}
\label{sec:arch}
\setcounter{table}{0}
\renewcommand{\thetable}{F\arabic{table}}
We detail inner operations of each of WALDO's modules (for the dimensions used on Cityscapes~\citesupp{cordts2016cityscapes}).

\subsection{Layered video decomposition}\vspace{-2.5em}~\begin{wraptable}[17]{r}{6cm}
\setlength\heavyrulewidth{.25ex}
\aboverulesep=0ex
\belowrulesep=.3ex
\centering
\caption{Input encoding.}
\small

\begin{tabular}{@{}lccc@{}}
\toprule
Stage & Operation & On & Output size \\
\midrule
$s$ & - & - & $20\stimes 128 \stimes 256$ \\
$f$ & - & - & $2\stimes 128 \stimes 256$ \\
1 & Concat. & $s$, $f$ & $22\stimes 128 \stimes 256$ \\
\midrule
Conv$_1$ & $[3 \stimes 3]$ & 1 & $64\stimes 64 \stimes 128$ \\
LN$_1$  & Norm. & 1 & $64\stimes 64 \stimes 128$ \\
Act$_1$  & GELU & 1 & $64\stimes 64 \stimes 128$ \\
Conv$_2$  & $[3 \stimes 3]$ & 1 & $128\stimes 32 \stimes 64$ \\
LN$_2$ & Norm. & 1 & $128\stimes 32 \stimes 64$ \\
Act$_2$ & GELU & 1 & $128\stimes 32 \stimes 64$ \\
Conv$_3$ & $[3 \stimes 3]$ & 1 & $256\stimes 16 \stimes 32$ \\
LN$_3$ & Norm. & 1 & $256\stimes 16 \stimes 32$ \\
Act$_3$ & GELU & 1 & $256\stimes 16 \stimes 32$ \\
Conv$_4$ & $[3 \stimes 3]$ & 1 & $512\stimes 8 \stimes 16$ \\
\midrule
$y$ & - & 1 & $512\stimes 8 \stimes 16$ \\
\bottomrule
\end{tabular}

\label{tab:e}
\end{wraptable}
\hspace{-0.2em}\myparagraph{Input encoding.}  Semantic and flow maps ($s$ and $f$) are concatenated and go through a series of $3\stimes 3$ convolutional layers (Conv) with padding of $1$ and stride of $2$, each followed by layer norm (LN) and a GELU activation~\citesupp{hendrycks2016gelu}~(Act) to form downscaled features $y$ for a given time step.

\myparagraph{Layer feature extraction.} We combine features encoded from different time steps into~($Y$), and sum them with different embeddings corresponding to their temporal ordering and spatial positioning~(T and P) to form input~(1). Layer features $Z_{obj}$ and $Z_{bg}$ are computed from input (2), which is the concatenation of object and background embeddings (2a and 2b), themselves expressed as the sum of layer ordering and spatial positioning embeddings (S, L, O and B). The next operations consists in layer norm (LN), self-attention (Att) to update (2) by computing queries, keys and values for (1) but only keys and values for (2), and multi-layer perceptrons (MLP) with GELU activations~\citesupp{hendrycks2016gelu}.

\myparagraph{Control point positioning.} We apply similar operations to predict control points $(p_{obj},p_{bg})$ for object and background layers at a given time step from associated features $y$. A key difference with the previous module is that this one is time-independent, and that inputs (1) and (2) play the same role in the transformer blocks. A fully-connected layer (FC) outputs the 3D position of each control point in each of the $16+1$ layers.

\myparagraph{Object masking.} This module predicts an alpha-transparency mask $a$ for an object from its associated features $z$. It is the reverse process compared to the input encoding module, that is, we replace convolutions with transposed ones for progressively upscaling feature maps $z$, and we decrease the size of features at each step instead of increasing it.

\myparagraph{Object classification.} This module predicts for each object represented by $z$ a soft class assignment $c$, by first averaging $z$ over its spatial dimensions, and then applying layer norm (LN), a fully-connected layer (FC), and a softmax activation (Act) to output $c$ as a categorical distribution over classes.

\subsection{Future layer prediction}

\myparagraph{Past encoding.} Past control points corresponding to objects (resp. the background) are transformed into vectors, each of them paired with one layer and one time step, using a fully-connected layer FC1 (resp. FC2). We apply a pooling operation to layer representations ($Z_{obg}$ and $Z_{bg}$) to reduce them to a single vector representing each layer. All these vectors are concatenated, summed with the suitable embeddings (T and P), and passed through two vanilla transformer blocks to produce encoded features $E$.

\myparagraph{Future decoding.} Future control points are obtained by initializing future representations (2) with some embeddings (T and P), and then alternating between transformer blocks with self-attention on future representations (2), and cross-attention from past to future ones (1 and 2).

\subsection{Warping, inpainting and fusion}

The final component of our approach is a U-Net~\citesupp{ronneberger2015unet} composed of $6$ downscaling layers and $6$ upscaling ones, with as many skip connections between the two branches. Each downscaling (resp. upscaling) layer divides (resp. multiplies) by $2$ its input resolution using a $3 \stimes 3$ convolution (resp. transposed convolution) with a stride of $2$, and multiplies (resp. divides) by $2$ the size of features so that intermediate features (between the two branches) are of size $512$. This module outputs RGB values to update certain regions of an image (filling missing background or object parts, adapting light effects or shadows in other parts), a mask indicating these regions, a score (of confidence) at each pixel location to allow fusing multiple views corresponding to the same image.

\begin{minipage}[c]{0.49\textwidth}
\setlength\heavyrulewidth{.25ex}
\aboverulesep=0ex
\belowrulesep=.3ex
\centering
\small
\begin{table}[H]
\caption{Layer feature extraction.~~~~~~~~~~~~~~~~~~~~~~}
\begin{tabular}{@{}C{0.8cm}@{}C{2.2cm}@{}C{1.7cm}@{}C{2.1cm}@{}}
\toprule
Stage & Operation & On & Output size \\
\midrule
$Y$ & - & - & $4\stimes 512\stimes 8 \stimes 16$\\
(T) & Embed. & - & $4\stimes 512\stimes 1 \stimes 1$\\
(P) & Embed. & - & $1\stimes 512\stimes 8 \stimes 16$\\
1 & Sum/Reshape & $Y$, T, P & $512\stimes 512$\\
\midrule
(S) & Embed. & - & $1\stimes 512\stimes 4 \stimes 4$\\
(L) & Embed. & - & $1\stimes 512\stimes 8 \stimes 16$\\
(O) & Embed. & - & $16\stimes 512\stimes 1 \stimes 1$\\
(B) & Embed. & - & $1\stimes 512\stimes 1 \stimes 1$\\
2a & Sum/Reshape & S, O & $256\stimes 512$\\
2b & Sum/Reshape & L, B & $128\stimes 512$\\
2 & Concat. & 2a, 2b & $384\stimes 512$\\
\midrule
LN$_1$ & Norm. & 1 &  $512\stimes 512$ \\
LN$_2$ & Norm. & 2 &  $384\stimes 512$ \\
Att$_1$ & $[512]\stimes 3$ & 2/1 & $384\stimes 512$\\
LN$_3$ & Norm. & 2 &  $384\stimes 512$ \\
MLP$_1$ & $[2048, 512]$ & 2 &  $384\stimes 512$ \\
LN$_4$ & Norm. & 2 &  $384\stimes 512$ \\
Att$_2$ & $[512]\stimes 3$ & 2/1 & $384\stimes 512$\\
LN$_5$ & Norm. & 2 &  $384\stimes 512$ \\
MLP$_2$ & $[2048, 512]$ & 2 &  $384\stimes 512$ \\
\midrule
$Z_{obj}$ & Split/Reshape & 2 & $16\stimes 512\stimes 4 \stimes 4$\\
$Z_{bg}$ & Split/Reshape & 2 & $1\stimes 512\stimes 8 \stimes 16$\\
\bottomrule
\end{tabular}
\end{table}
\end{minipage}\hspace{2.1em}
\begin{minipage}[c]{0.49\textwidth}
\setlength\heavyrulewidth{.25ex}
\aboverulesep=0ex
\belowrulesep=.3ex
\centering
\small
\begin{table}[H]
\caption{Control point positioning.~~~~~~~~~~~~~~~~~~~~}
\begin{tabular}{@{}C{0.8cm}@{}C{2.2cm}@{}C{1.7cm}@{}C{2.1cm}@{}}
\toprule
Stage & Operation & On & Output size \\
\midrule
$y$ & - & - & $512\stimes 8 \stimes 16$\\
(P) & Embed. & - & $512\stimes 8 \stimes 16$\\
1 & Sum/Reshape & $y$, P & $128\stimes 512$\\
\midrule
$Z_{obj}$ & - & - & $16\stimes 512\stimes 4 \stimes 4$\\
$Z_{bg}$ & - & - & $1\stimes 512\stimes 8 \stimes 16$\\
(S) & Embed. & - & $1\stimes 512\stimes 4 \stimes 4$\\
(L) & Embed. & - & $1\stimes 512\stimes 8 \stimes 16$\\
(O) & Embed. & - & $16\stimes 512\stimes 1 \stimes 1$\\
(B)  & Embed. & - & $1\stimes 512\stimes 1 \stimes 1$\\
2a & Sum/Reshape & $Z_{obj}$, S, O & $16\stimes 512\stimes 4 \stimes 4$\\
2b & Sum/Reshape & $Z_{bg}$, L, B & $1\stimes 512\stimes 8 \stimes 16$\\
2 & Concat. & 2a, 2b & $384\stimes 512$\\
\midrule
3 & Concat. & 1, 2 & $512\stimes 512$\\
\midrule
LN$_1$ & Norm. & 3 &  $512\stimes 512$ \\
Att$_1$ & $[512]\stimes 3$ & 3 & $512\stimes 512$\\
LN$_2$ & Norm. & 3 &  $512\stimes 512$ \\
MLP$_1$ & $[2048, 512]$ & 3 &  $512\stimes 512$ \\
LN$_3$ & Norm. & 3 &  $512\stimes 512$ \\
Att$_2$ & $[512]\stimes 3$ & 3 & $512\stimes 512$\\
LN$_4$ & Norm. & 3 &  $512\stimes 512$ \\
MLP$_2$ & $[2048, 512]$ & 3 &  $512\stimes 512$ \\
2 & Split & 3 & $384\stimes 512$\\
FC & [3] & 2 &  $384\stimes 3$ \\
\midrule
$p_{obj}$ & Split/Reshape & 2 & $16\stimes 3\stimes 4 \stimes 4$\\
$p_{bg}$ & Split/Reshape & 2 & $1\stimes 3\stimes 8 \stimes 16$\\
\bottomrule
\end{tabular}
\end{table}
\end{minipage}

~

~

\begin{minipage}[c]{0.49\textwidth}
\setlength\heavyrulewidth{.25ex}
\aboverulesep=0ex
\belowrulesep=.3ex
\centering
\small
\begin{table}[H]
\caption{Object masking.~~~~~~~~~~~~~~~~~~~~~~~}
\begin{tabular}{@{}C{0.8cm}@{}C{2.2cm}@{}C{1.7cm}@{}C{2.1cm}@{}}
\toprule
Stage & Operation & On & Output size \\
\midrule
$z$ & - & - & $512\stimes 4 \stimes 4$ \\
1 & - & $z$ & $512\stimes 4 \stimes 4$ \\
\midrule
LN$_1$  & Norm. & 1 & $512\stimes 4 \stimes 4$ \\
TConv$_1$ & $[3 \stimes 3]$ & 1 & $256\stimes 8 \stimes 8$ \\
LN$_2$  & Norm. & 1 & $256\stimes 8 \stimes 8$ \\
Act$_1$  & GELU & 1 & $256\stimes 8 \stimes 8$ \\
TConv$_2$  & $[3 \stimes 3]$ & 1 & $128\stimes 16 \stimes 16$ \\
LN$_3$ & Norm. & 1 & $128\stimes 16 \stimes 16$ \\
Act$_2$ & GELU & 1 & $128\stimes 16 \stimes 16$ \\
TConv$_3$ & $[3 \stimes 3]$ & 1 & $64\stimes 32 \stimes 32$ \\
LN$_4$ & Norm. & 1 & $64\stimes 32 \stimes 32$ \\
Act$_3$ & GELU & 1 & $64\stimes 32 \stimes 32$ \\
TConv$_4$ & $[3 \stimes 3]$ & 1 & $1\stimes 64 \stimes 64$ \\
Act$_4$ & Sigmoid & 1 & $1\stimes 64 \stimes 64$ \\
\midrule
$a$ & - & 1 & $1\stimes 64 \stimes 64$ \\
\bottomrule
\end{tabular}
\end{table}
\end{minipage}\hspace{2.1em}
\begin{minipage}[c]{0.49\textwidth}
\setlength\heavyrulewidth{.25ex}
\aboverulesep=0ex
\belowrulesep=.3ex
\centering
\small
\begin{table}[H]
\caption{Object classification.~~~~~~~~~~~~~~~~~~~~~~}
\begin{tabular}{@{}C{0.8cm}@{}C{2.2cm}@{}C{1.7cm}@{}C{2.1cm}@{}}
\toprule
Stage & Operation & On & Output size \\
\midrule
$z$ & - & - & $512\stimes 4 \stimes 4$ \\
1 & Spatial mean & $z$ & $512$ \\
\midrule
LN  & Norm. & 1 & $512$ \\
FC & $[20]$ & 1 & $20$ \\
Act & Softmax & 1 & $20$ \\
\midrule
$c$ & - & 1 & $20$ \\
\bottomrule
\end{tabular}
\end{table}
\end{minipage}

\begin{minipage}[c]{0.49\textwidth}
\setlength\heavyrulewidth{.25ex}
\aboverulesep=0ex
\belowrulesep=.3ex
\centering
\small
\begin{table}[H]
\caption{Past encoding.~~~~~~~~~~~~~~~~~~~~~~}
\begin{tabular}{@{}C{0.8cm}@{}C{2.2cm}@{}C{1.7cm}@{}C{2.1cm}@{}}
\toprule
Stage & Operation & On & Output size \\
\midrule
$P_{obj}$ & - & - & $4 \stimes 16\stimes 3\stimes 4 \stimes 4$\\
$P_{bg}$ & - & - & $4\stimes 1\stimes 3\stimes 8 \stimes 16$\\
1a & Reshape & $P_{obj}$ & $4 \stimes 16\stimes 48$\\
1b & Reshape & $P_{bg}$ & $4 \stimes 1\stimes 128$\\
FC1 & [512] & 1a & $4 \stimes 16\stimes 512$\\
FC2 & [512] & 1b & $4 \stimes 1\stimes 512$\\
\midrule
$Z_{obj}$ & - & - & $16\stimes 512\stimes 4 \stimes 4$\\
$Z_{bg}$ & - & - & $1\stimes 512\stimes 8 \stimes 16$\\
2a & Pool/Reshape & $Z_{obj}$ & $1\stimes 16\stimes 512$\\
2b & Pool/Reshape & $Z_{bg}$ & $1\stimes 1\stimes 512$\\
\midrule
3 & Concat. & 1a, 1b, 2a, 2b & $5\stimes 17\stimes 512$\\
(T) & Embed. & - & $5\stimes 1 \stimes 512$\\
(P) & Embed. & - & $1\stimes 17 \stimes 512$\\
3 & Sum/Reshape & 3, T, P & $85\stimes 512$\\
\midrule
LN$_1$ & Norm. & 3 &  $85\stimes 512$ \\
Att$_1$ & $[512]\stimes 3$ & 3 & $85\stimes 512$\\
LN$_2$ & Norm. & 3 &  $85\stimes 512$ \\
MLP$_1$ & $[2048, 512]$ & 3 &  $85\stimes 512$ \\
LN$_3$ & Norm. & 3 &  $85\stimes 512$ \\
Att$_2$ & $[512]\stimes 3$ & 3 & $85\stimes 512$\\
LN$_4$ & Norm. & 3 &  $85\stimes 512$ \\
MLP$_2$ & $[2048, 512]$ & 3 &  $85\stimes 512$ \\
\midrule
$E$ & Reshape & 3 & $5\stimes 17\stimes 512$\\
\bottomrule
\end{tabular}
\end{table}
\end{minipage}\hspace{2.1em}
\begin{minipage}[c]{0.49\textwidth}
\setlength\heavyrulewidth{.25ex}
\aboverulesep=0ex
\belowrulesep=.3ex
\centering
\small
\begin{table}[H]
\caption{Future decoding.~~~~~~~~~~~~~~~~~~~~~~}
\begin{tabular}{@{}C{0.8cm}@{}C{2.2cm}@{}C{1.7cm}@{}C{2.1cm}@{}}
\toprule
Stage & Operation & On & Output size \\
\midrule
$E$ & - & - & $5\stimes 17\stimes 512$\\
1 & Reshape & $E$ & $85\stimes 512$\\
\midrule
(T) & Embed. & - & $10\stimes 1 \stimes 512$\\
(P) & Embed. & - & $1\stimes 17 \stimes 512$\\
2 & Sum/Reshape & T, P & $170\stimes 512$\\
\midrule
LN$_1$ & Norm. & 1 &  $170\stimes 512$ \\
LN$_2$ & Norm. & 2 &  $170\stimes 512$ \\
Att$_1$ & $[512]\stimes 3$ & 2 & $170\stimes 512$\\
LN$_3$ & Norm. & 2 &  $170\stimes 512$ \\
MLP$_1$ & $[2048, 512]$ & 2 &  $170\stimes 512$ \\
LN$_4$ & Norm. & 2 &  $170\stimes 512$ \\
Att$_2$ & $[512]\stimes 3$ & 2/1 & $170\stimes 512$\\
LN$_5$ & Norm. & 2 &  $170\stimes 512$ \\
MLP$_2$ & $[2048, 512]$ & 2 &  $170\stimes 512$ \\
LN$_6$ & Norm. & 2 &  $170\stimes 512$ \\
Att$_3$ & $[512]\stimes 3$ & 2 & $170\stimes 512$\\
LN$_7$ & Norm. & 2 &  $170\stimes 512$ \\
MLP$_3$ & $[2048, 512]$ & 2 &  $170\stimes 512$ \\
LN$_8$ & Norm. & 2 &  $170\stimes 512$ \\
Att$_4$ & $[512]\stimes 3$ & 2/1 & $170\stimes 512$\\
LN$_9$ & Norm. & 2 &  $170\stimes 512$ \\
MLP$_4$ & $[2048, 512]$ & 2 &  $170\stimes 512$ \\
LN$_{10}$ & Norm. & 2 &  $170\stimes 512$ \\
2a & Split & 2 & $160\stimes 512$\\
2b & Split & 2 & $10\stimes 512$\\
FC1 & [48] & 2a &  $160\stimes 48$ \\
FC2 & [384] & 2b &  $10\stimes 384$ \\
\midrule
$P_{obj}$ & Reshape & 2a & $10\stimes 16\stimes 3\stimes 4 \stimes 4$\\
$P_{bg}$ & Reshape & 2b & $10\stimes1\stimes 3\stimes 8 \stimes 16$\\
\bottomrule
\end{tabular}
\end{table}
\end{minipage}

~

\section{Additional visual results on nonrigid scenes}
\label{sec:taichi}
\setcounter{table}{0}
\renewcommand{\thetable}{G\arabic{table}}
\renewcommand{\thefigure}{G\arabic{figure}}

\begin{figure}[H]
\small
\centering
\setlength\tabcolsep{0.7pt}
\renewcommand{\arraystretch}{0.5}
\begin{tabular}{@{}cccccc@{}}
\croppedgraphicstaichi{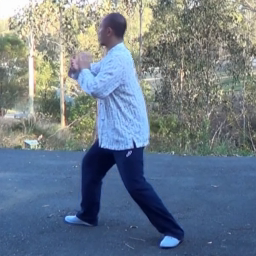} &
\croppedgraphicstaichi{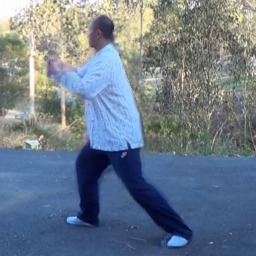} &
\croppedgraphicstaichi{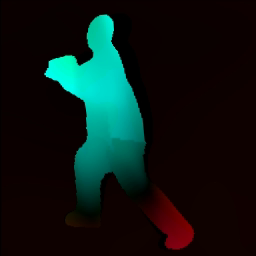} &
\croppedgraphicstaichi{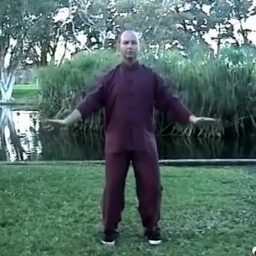} &
\croppedgraphicstaichi{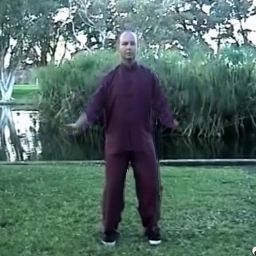} &
\croppedgraphicstaichi{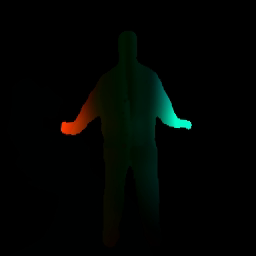} \\

\croppedgraphicstaichi{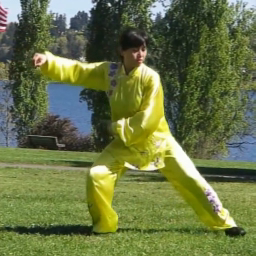} &
\croppedgraphicstaichi{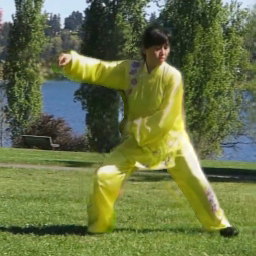} &
\croppedgraphicstaichi{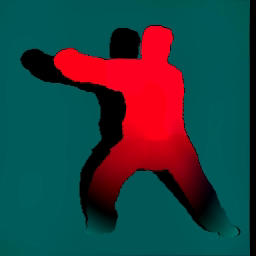} &
\croppedgraphicstaichi{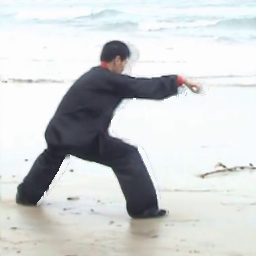} &
\croppedgraphicstaichi{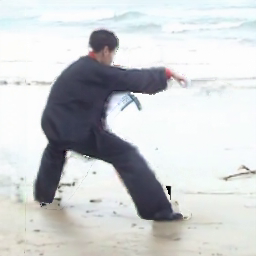} &
\croppedgraphicstaichi{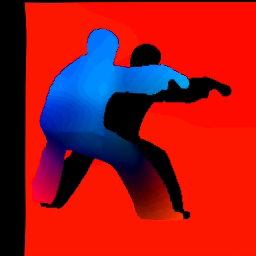} \\

{$T{+}1$} & {$T{+}10$} & {Warp} & {$T{+}1$} & {$T{+}10$} & {Warp}\\

\end{tabular}   

\caption{\blue{Future prediction from $T{=}4$ frames on Taichi-HD ({\tiny $256\stimes256$}). Nonrigid motions can be visualized by the associated warps, predicted from the control points between $T$ and $T{+}10$ (colors represent different directions).}}
\label{fig:taichi}
\end{figure}

We further illustrate (Figure~\ref{fig:taichi}) that the motion representation proposed in WALDO allows us to represent nonrigid object deformations by training on Taichi-HD~\cite{siarohin2019first} dataset.
Our approach can handle complex human motions such as leaning forward / backward, moving one leg while keeping the other on the ground, raising individual arms... 

\section{Influence of the choice of the pretrained segmentation and optical flow models}
\label{sec:setup}
\setcounter{table}{0}
\renewcommand{\thetable}{H\arabic{table}}

The use of pretrained networks in prior works vary widely according to the level of information required by each method, and depending on what was available at the time to extract these information. We summarize the differences in Table~\ref{tab:input}. 

\begin{table}[h]
\setlength\heavyrulewidth{.25ex}
\aboverulesep=0ex
\belowrulesep=.3ex
\centering
\caption{Pretrained models used by different methods.}
\small

\begin{tabular}{@{}L{2cm}@{}>{\centering}m{2.35cm}@{}>{\centering}m{2.35cm}@{}>{\centering}m{2.35cm}@{}>{\centering}m{2.35cm}@{}>{\centering}m{2.35cm}@{}>{\centering\arraybackslash}m{2.35cm}@{}}
\toprule
{\normalsize Method} & {Optical flow estimation} & {Semantic segmentation} & {Instance segmentation} & {Depth estimation} & {Object tracking} & {Video Frame Interpolation} \\
\midrule
VPVFI~\citesupp{wu2022optimizing} & RAFT~\citesupp{teed2020raft} & - & - & - & - & RIFE~\citesupp{huang2022rife} \\
VPCL~\citesupp{geng2022comparing} & RAFT~\citesupp{teed2020raft} & - & - & - & - & - \\
Vid2vid~\citesupp{wang2018video} & FlowNet2~\citesupp{ilg2017flownet} & DeepLabV3~\citesupp{chen2018encoder} & - & - & - & - \\
OMP~\citesupp{wu2020future} & PWCNet~\citesupp{sun2018pwc} & VPLR~\citesupp{zhu2019improving} & UPSNet~\citesupp{xiong2019upsnet} & GeoNet~\citesupp{yin2018geonet} & SiamRPN++\citesupp{li2019siamrpn} & - \\
SADM~\citesupp{bei2021learning} & PWCNet~\citesupp{sun2018pwc} & DeepLabV3~\citesupp{chen2018encoder} & - & - & - & - \\
\midrule
WALDO & RAFT~\citesupp{teed2020raft} & DeepLabV3~\citesupp{chen2018encoder} & - & - & - & - \\
\bottomrule 
\end{tabular}

\label{tab:input}
\end{table}

We thus evaluate the influence of the choice of the pretrained segmentation and optical flow models to WALDO's performance. Results on Cityscapes and KITTI test set, obtained by substituting segmentation model DeepLabV3~\citesupp{chen2018encoder} with MobileNetV2~\citesupp{sandler2018mobilenetv2} or ViT-Adapter~\citesupp{chen2022vitadapter}, and optical flow model RAFT~\citesupp{teed2020raft} with PWCNet~\citesupp{sun2018pwc}, are presented in Table~\ref{tab:model}.

\begin{table}[h]
\setlength\heavyrulewidth{.25ex}
\aboverulesep=0ex
\belowrulesep=.3ex
\centering
\caption{Ablation studies of optical flow estimation and semantic segmentation methods on the Cityscapes and KITTI test sets. Like in the main paper, we compute multi-scale SSIM ({\small$\times 10^3$}) and LPIPS ({\small$\times 10^3$}) for the $k^\text{th}$ frame and report the average for $k$ in $\llbracket 1, K \rrbracket$.}
\small

\begin{tabular}{@{}c@{\hskip 0.55em}c@{\hskip 0.55em}c@{}}

\begin{tabular}{@{}C{.95cm}C{.95cm}@{}C{.95cm}@{}C{.95cm}@{}C{.95cm}@{}}
\\
\toprule
\multicolumn{2}{@{}c@{}}{Flow} & \multicolumn{3}{@{}c@{}}{Segmentation} \\
\cmidrule(r){1-2}\cmidrule(){3-5}
\citesupp{sun2018pwc} & \citesupp{teed2020raft} & \citesupp{sandler2018mobilenetv2} & \citesupp{chen2018encoder} & \citesupp{chen2022vitadapter} \\
\midrule
\checkmark & & & \checkmark & \\
& \checkmark & \checkmark &  & \\
& \checkmark & & \checkmark & \\
& \checkmark & & & \checkmark \\
\bottomrule 
\end{tabular} &

\begin{tabular}{@{}C{.95cm}@{}C{.95cm}@{}C{.95cm}@{}C{.95cm}@{}C{.95cm}@{}C{.95cm}@{}}
\multicolumn{6}{@{}c@{}}{(a) Cityscapes} \\
\toprule
\multicolumn{2}{@{}c@{}}{$K=1$} & \multicolumn{2}{@{}c@{}}{$K=5$} & \multicolumn{2}{@{}c@{}}{$K=10$} \\
\cmidrule(r){1-2}\cmidrule(r){3-4}\cmidrule(){5-6}
{\scriptsize SSIM $\uparrow$} & {\scriptsize LPIPS $\downarrow$} & {\scriptsize SSIM $\uparrow$} & {\scriptsize LPIPS $\downarrow$} & {\scriptsize SSIM $\uparrow$} & {\scriptsize LPIPS $\downarrow$} \\
\midrule
947 & {062} & {836} & {122} & {753} & {175} \\
954 & 055 & 849 & 111 & 768 & 165 \\
$\textbf{957}$ & $\textbf{049}$ & $\textbf{854}$ & $\textbf{105}$ & $\textbf{771}$ & $\textbf{158}$ \\
$\textbf{957}$ & $\underline{050}$ & $\underline{853}$ & $\textbf{105}$ & $\underline{770}$ & $\underline{159}$ \\
\bottomrule 
\end{tabular} &

\begin{tabular}{@{}C{.95cm}@{}C{.95cm}@{}C{.95cm}@{}C{.95cm}@{}C{.95cm}@{}C{.95cm}@{}}
\multicolumn{6}{@{}c@{}}{(b) KITTI} \\
\toprule
\multicolumn{2}{@{}c@{}}{$K=1$} & \multicolumn{2}{@{}c@{}}{$K=3$} & \multicolumn{2}{@{}c@{}}{$K=5$} \\
\cmidrule(r){1-2}\cmidrule(r){3-4}\cmidrule(){5-6}
{\scriptsize SSIM $\uparrow$} & {\scriptsize LPIPS $\downarrow$} & {\scriptsize SSIM $\uparrow$} & {\scriptsize LPIPS $\downarrow$} & {\scriptsize SSIM $\uparrow$} & {\scriptsize LPIPS $\downarrow$} \\
\midrule
{856} & {116} & {760} & {171} & {697} & {214} \\
859 & 112 & 756 & 166 & 692 & 209 \\
$\textbf{867}$ & $\textbf{108}$ & $\underline{766}$ & $\textbf{163}$ & $\underline{702}$ & $\underline{206}$ \\
$\underline{866}$ & $\underline{109}$ & $\textbf{767}$ & $\textbf{163}$ & $\textbf{703}$ & $\textbf{205}$ \\
\bottomrule 
\end{tabular}

\end{tabular}
\\

\label{tab:model}
\end{table}

Despite PWCNet~\citesupp{sun2018pwc} having approximately twice the average end-point-error of RAFT~\citesupp{teed2020raft}, using it instead of RAFT only results in a small performance drop for video prediction with WALDO. This is in line with the conclusions of Geng~\etal in~\citesupp{geng2022comparing} who conducted a similar comparison. Moreover, replacing DeepLabV3~\citesupp{chen2018encoder} with MobileNetV2~\citesupp{sandler2018mobilenetv2}, with respective segmentation performance of $81$ and $75$ in terms of test set mIoU on Cityscapes, yields only little loss of video prediction quality. Conversely, using a more advanced segmentation model like ViT-Adapter~\citesupp{chen2022vitadapter}, with $85$ test set mIoU on Cityscapes, does not change the results substantially. Our interpretation is that the segmentation quality is not the limiting factor for WALDO's performance once it has reached a sufficient level. To conclude, WALDO is quite robust to the choice of the pretrained segmentation and optical flow models.

\section{Ablation study of the inpainting strategy}
\label{sec:inpabl}
\setcounter{table}{0}
\renewcommand{\thetable}{I\arabic{table}}
\renewcommand{\thefigure}{I\arabic{figure}}

\begin{table}[H]
\setlength\heavyrulewidth{.25ex}
\aboverulesep=0ex
\belowrulesep=.3ex
\centering
\caption{Ablation of our inpainting strategy on the Cityscapes test set.}
\small

\begin{tabular}{@{}C{3.5cm}@{}C{3.5cm}@{}L{0.90cm}@{}L{1.00cm}@{}L{0.90cm}@{}L{1.00cm}@{}L{0.90cm}@{}L{1.00cm}@{}L{0.90cm}@{}}
\toprule
\multirow{2}{*}{Adversarial inpainting~\citesupp{li2022mat}} &
\multirow{2}{*}{Temporal consistency} &
\multicolumn{2}{@{}c@{}}{$K=1$} & \multicolumn{2}{@{}c@{}}{$K=5$} & \multicolumn{3}{@{}c@{}}{$K=10$} \\
\cmidrule(r){3-4}\cmidrule(r){5-6}\cmidrule(){7-9}
& & {\footnotesize SSIM $\uparrow$} & {\footnotesize LPIPS $\downarrow$} & {\footnotesize SSIM $\uparrow$} & {\footnotesize LPIPS $\downarrow$} & {\footnotesize SSIM $\uparrow$} & {\footnotesize LPIPS $\downarrow$} & {\footnotesize FVD $\downarrow$} \\
\midrule
& & ~~$\textbf{957}$ & ~~$\textbf{049}$ & ~~${853}$ & ~~$\textbf{105}$ & ~~${770}$ & ~~$\textbf{158}$ & ~~{061}\\
\checkmark & & ~~$\textbf{957}$ & ~~$\textbf{049}$ & ~~$\textbf{854}$ & ~~$\textbf{105}$ & ~~$\textbf{771}$ & ~~$\textbf{158}$ & ~~$\underline{057}$ \\
\checkmark & \checkmark & ~~$\textbf{957}$ & ~~$\textbf{049}$ & ~~$\textbf{854}$ & ~~$\textbf{105}$ & ~~$\textbf{771}$ & ~~$\textbf{158}$ & ~~$\textbf{055}$ \\
\bottomrule 
\end{tabular} 
\vspace{-0.5em}
\label{tab:inp}
\end{table}
\begin{figure}[H]
	\setlength\tabcolsep{1.5pt}
	\renewcommand{\arraystretch}{1.5}
	\small
    \centering
	\begin{tabular}{C{0.8cm}cccccc}
 		
 	      \raisebox{3.2\normalbaselineskip}[0pt][0pt]{\rotatebox[origin=c]{90}{$\varnothing$}} &
 		\includegraphics[width=.150\linewidth]{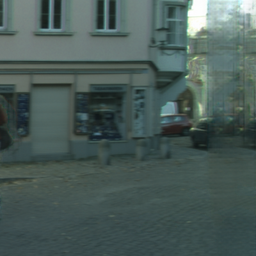} &
 		\includegraphics[width=.150\linewidth]{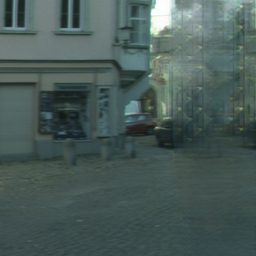} &
 		\includegraphics[width=.150\linewidth]{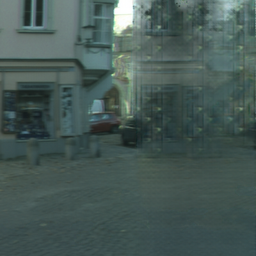} &
 		\includegraphics[width=.150\linewidth]{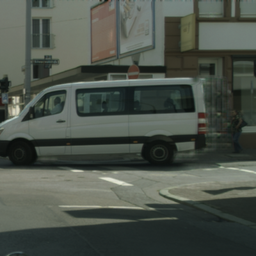} &
 		\includegraphics[width=.150\linewidth]{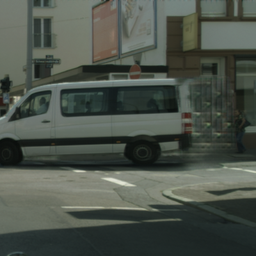} &
 		\includegraphics[width=.150\linewidth]{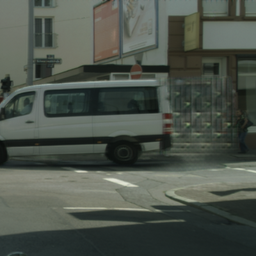} \\[-0.12cm]

        \raisebox{3.2\normalbaselineskip}[0pt][0pt]{\rotatebox[origin=c]{90}{Adv.\ inp.~\cite{li2022mat}}} &
 		\includegraphics[width=.150\linewidth]{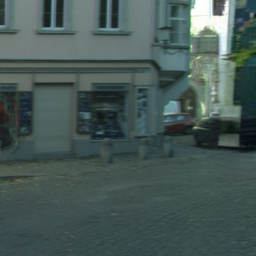} &
 		\includegraphics[width=.150\linewidth]{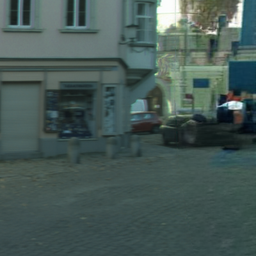} &
 		\includegraphics[width=.150\linewidth]{files/figures/inp/1/013_adv.png} &
 		\includegraphics[width=.150\linewidth]{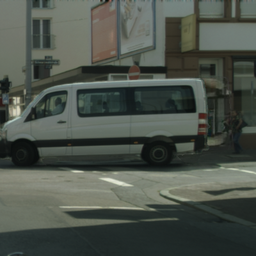} &
 		\includegraphics[width=.150\linewidth]{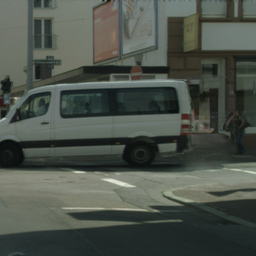} &
 		\includegraphics[width=.150\linewidth]{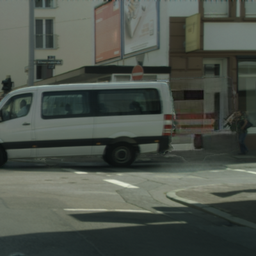} \\[-0.12cm]

        \raisebox{3.2\normalbaselineskip}[0pt][0pt]{\rotatebox[origin=c]{90}{\parbox{4cm}{\centering Adv.\ inp.~\cite{li2022mat} \\ + Temp.\ cons.}}} &
 		\includegraphics[width=.150\linewidth]{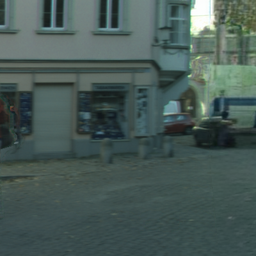} &
 		\includegraphics[width=.150\linewidth]{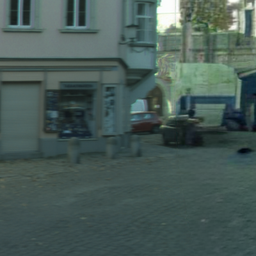} &
 		\includegraphics[width=.150\linewidth]{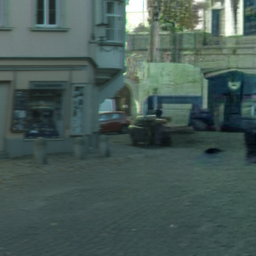} &
 		\includegraphics[width=.150\linewidth]{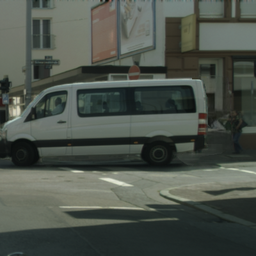} &
 		\includegraphics[width=.150\linewidth]{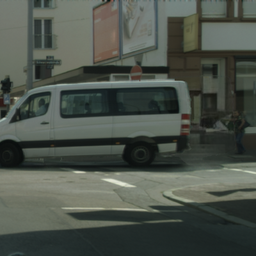} &
 		\includegraphics[width=.150\linewidth]{files/figures/inp/2/013_adv_cons.png} \\[-0.12cm]

        \raisebox{3.2\normalbaselineskip}[0pt][0pt]{\rotatebox[origin=c]{90}{\parbox{4cm}{\centering Disoccluded \\ regions}}} &
 		\includegraphics[width=.150\linewidth]{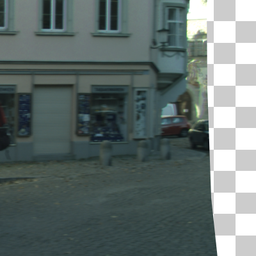} &
 		\includegraphics[width=.150\linewidth]{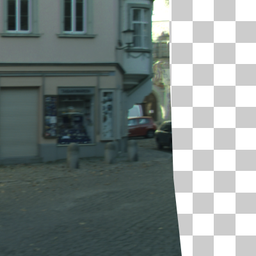} &
 		\includegraphics[width=.150\linewidth]{files/figures/inp/1/013_gtc.png} &
 		\includegraphics[width=.150\linewidth]{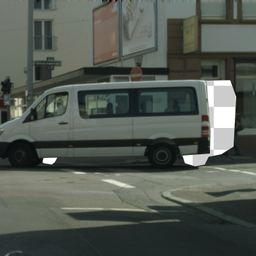} &
 		\includegraphics[width=.150\linewidth]{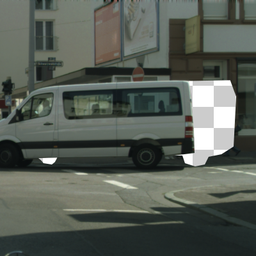} &
 		\includegraphics[width=.150\linewidth]{files/figures/inp/2/013_gtc.png} \\[-0.12cm]

   & $T+4$ & $T+7$ & $T+10$ & $T+4$ & $T+7$ & $T+10$ \\
	\end{tabular}
	\caption{Visual ablation of our inpainting strategy. We start off with a method that fills in empty regions with the sole objective to minimize the reconstruction error ($\varnothing$), we then propose to leverage an off-the-shelf adversarial inpainting (adv.\ inp.) method~\cite{li2022mat} to improve realism, and we finally illustrate our full strategy where we use the predicted flow to ensure the temporal consistency (temp.\ const.) of inpainted regions. For clarity, these regions are indicated in the last row. Please zoom in for details.}
	\label{fig:qualinp}
\end{figure}

Given that we use an off-the-shelf inpainting method~\cite{li2022mat} trained on external data~\cite{zhou2017places}, we assess the impact of the use of such a model in our approach on quantitative measurements.

Results are presented in Table~\ref{tab:inp} and show that although gains in SSIM and FVD are possible, these gains remain small. In addition, no perceptible change is observed on LPIPS. Our temporal consistency strategy, which consists in filling frames one by one and using the predicted flow to propagate new contents into subsequent frames, allow small extra gains on the FVD metric.

The benefits from our inpainting strategy are more visible in qualitative samples presented in Figure~\ref{fig:qualinp}. Without adversarial inpainting ($\varnothing$), empty image regions are filled using a model trained to minimize the reconstruction error. We observe that this results in blurry image parts with important artefacts. Using adversarial inpainting produces much more realistic images when considering frames individually, but filling each of them independently is still not very natural (best viewed in the videos included in the \proj{}). Our approach for producing temporally-consistent outputs is able to solve this issue. For example, in the left-most sample sequence of the third row of Figure~\ref{fig:qualinp}, we see that inpainted image parts match between different time steps although the camera is moving. %

\section{Further implementation details}
\label{sec:imp}

We follow~\citesupp{wu2020future} and group semantic classes which form a consistent entity together, \eg, riders with their bicycle, traffic lights/signs with the poles, which allows us to represent those within a single object layer.
We use horizontal flips, cropping, and color jittering as data augmentation.
Although we encounter signs of over-fitting in some experiments, validation curves do not increase during training, nor after convergence.
So best model selection is not necessary, and we always save the last checkpoint.
We find that a good initialization of object regions helps the layer decomposition module to reach a better optimum and to converge faster.
Hence, we add a warmup period during which the module is trained without flow reconstruction ($\lambda_f=0$), and then progressively increase the associated parameter during training ($\lambda_f>0$) once we start having good object proposals.
Without doing so, the module tends to rely on the background layer alone to reconstruct the scene motion, which, as a result, leads to an under-use of the object layers.
We also find that removing data-augmentation in the last few epochs when training the warping, inpainting and fusing module slightly improves the performance at inference time.

\section{Societal impact}
\label{sec:state}
The total cost for this project, including architecture and hyperparameter search, training, testing and comparisons with baselines has been around 25K GPU hours, with an associated environmental cost of course. On the other hand, we strive to minimize this cost by decomposing video prediction into efficient lightweight modules, and our approach will hopefully contribute to eventually improve the safety of autonomous vehicles, by, say, predicting the motion of nearby agents.

\section{Qualitative study of WALDO}
\label{sec:qual-abl}
\setcounter{figure}{0}
\renewcommand{\thefigure}{L\arabic{figure}}

In this section, we provide qualitative samples for each of the three modules which compose WALDO.

\begin{figure}[h]
    \centering
	\setlength\tabcolsep{1.0pt}
	\renewcommand{\arraystretch}{1.5}
	\small
	\begin{tabular}{L{0.6cm}cccccc}
	& $t=1$ & $t=7$ & $t=14$ & $t=1$ & $t=7$ & $t=14$ \\
 		
 	\raisebox{1.5\normalbaselineskip}[0pt][0pt]{\rotatebox[origin=c]{90}{\parbox{4cm}{\centering Real \\ \vspace{-0.3em}frame}}} &
 	\includegraphics[width=.155\linewidth]{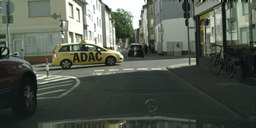} &
 	\includegraphics[width=.155\linewidth]{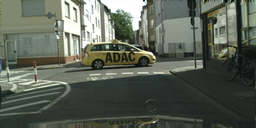} &
 	\includegraphics[width=.155\linewidth]{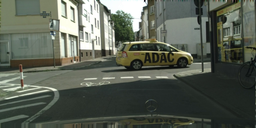} & \includegraphics[width=.155\linewidth]{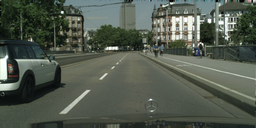} &
 	\includegraphics[width=.155\linewidth]{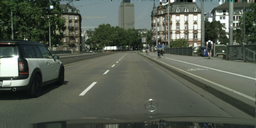} &
 	\includegraphics[width=.155\linewidth]{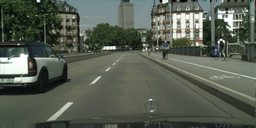}\\
 	
 	\raisebox{1.5\normalbaselineskip}[0pt][0pt]{\rotatebox[origin=c]{90}{\parbox{4cm}{\centering Real seg. \\ \vspace{-0.3em}map}}} &
 	\includegraphics[width=.155\linewidth]{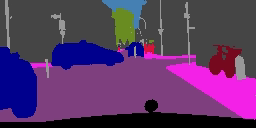} &
 	\includegraphics[width=.155\linewidth]{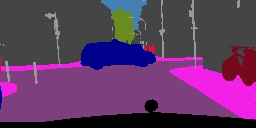} &
 	\includegraphics[width=.155\linewidth]{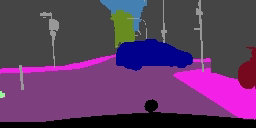} &
 	\includegraphics[width=.155\linewidth]{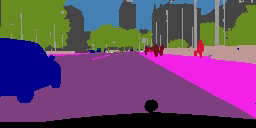} &
 	\includegraphics[width=.155\linewidth]{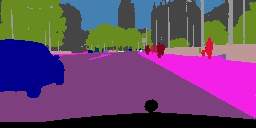} &
 	\includegraphics[width=.155\linewidth]{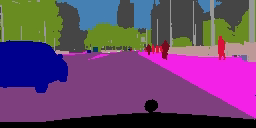}\\
 	
 	\raisebox{1.5\normalbaselineskip}[0pt][0pt]{\rotatebox[origin=c]{90}{\parbox{4cm}{\centering Real flow \\ \vspace{-0.3em}map}}} &
 	\includegraphics[width=.155\linewidth]{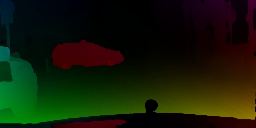} &
 	\includegraphics[width=.155\linewidth]{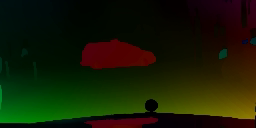} &
 	\includegraphics[width=.155\linewidth]{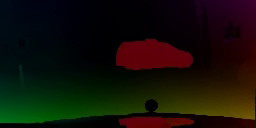} &
 	\includegraphics[width=.155\linewidth]{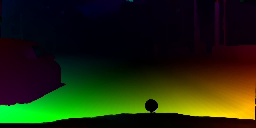} &
 	\includegraphics[width=.155\linewidth]{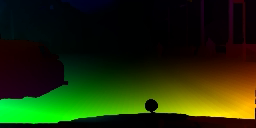} &
 	\includegraphics[width=.155\linewidth]{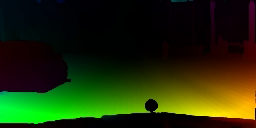}\\
 	
 	\raisebox{1.5\normalbaselineskip}[0pt][0pt]{\rotatebox[origin=c]{90}{\parbox{4cm}{\centering Pseudo \\ \vspace{-0.3em}objects}}} &
 	\includegraphics[width=.155\linewidth]{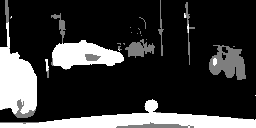} &
 	\includegraphics[width=.155\linewidth]{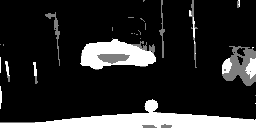} &
 	\includegraphics[width=.155\linewidth]{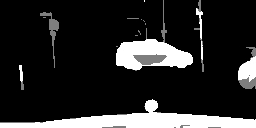} &
 	\includegraphics[width=.155\linewidth]{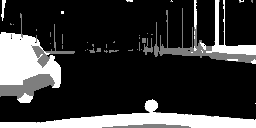} &
 	\includegraphics[width=.155\linewidth]{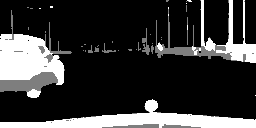} &
 	\includegraphics[width=.155\linewidth]{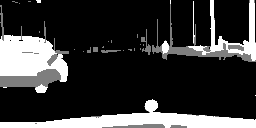}\\
 	
 	\raisebox{1.5\normalbaselineskip}[0pt][0pt]{\rotatebox[origin=c]{90}{\parbox{4cm}{\centering Rec. \\ \vspace{-0.3em}objects}}} &
 	\includegraphics[width=.155\linewidth]{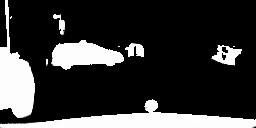} &
 	\includegraphics[width=.155\linewidth]{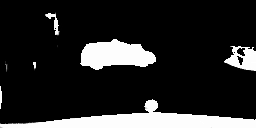} &
 	\includegraphics[width=.155\linewidth]{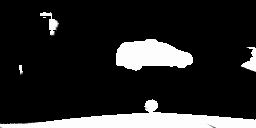} &
 	\includegraphics[width=.155\linewidth]{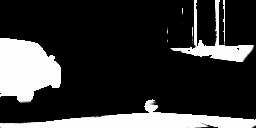} &
 	\includegraphics[width=.155\linewidth]{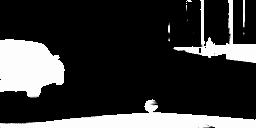} &
 	\includegraphics[width=.155\linewidth]{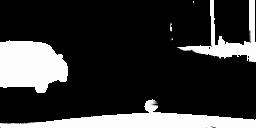}\\
 	
 	\raisebox{1.5\normalbaselineskip}[0pt][0pt]{\rotatebox[origin=c]{90}{\parbox{4cm}{\centering Layer \\ \vspace{-0.3em}masks}}} &
 	\includegraphics[width=.155\linewidth]{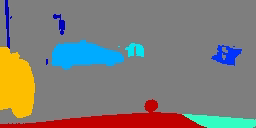} &
 	\includegraphics[width=.155\linewidth]{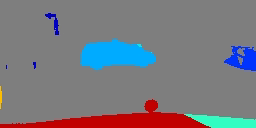} &
 	\includegraphics[width=.155\linewidth]{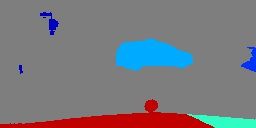} &
 	\includegraphics[width=.155\linewidth]{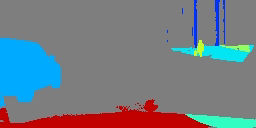} &
 	\includegraphics[width=.155\linewidth]{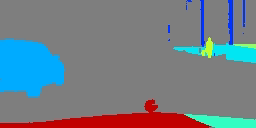} &
 	\includegraphics[width=.155\linewidth]{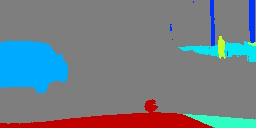}\\
 	
 	\raisebox{1.5\normalbaselineskip}[0pt][0pt]{\rotatebox[origin=c]{90}{\parbox{4cm}{\centering Rec. \\ \vspace{-0.3em}motion}}} &
 	\includegraphics[width=.155\linewidth]{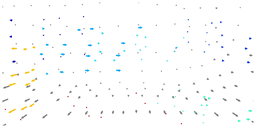} &
 	\includegraphics[width=.155\linewidth]{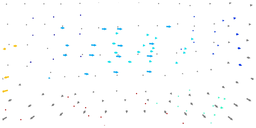} &
 	\includegraphics[width=.155\linewidth]{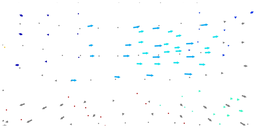} &
 	\includegraphics[width=.155\linewidth]{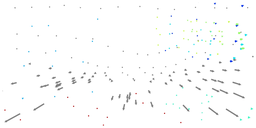} &
 	\includegraphics[width=.155\linewidth]{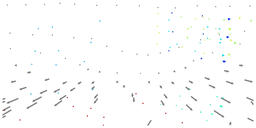} &
 	\includegraphics[width=.155\linewidth]{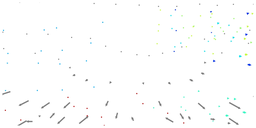}\\
 	
 	\raisebox{1.5\normalbaselineskip}[0pt][0pt]{\rotatebox[origin=c]{90}{\parbox{4cm}{\centering Rec. \\ \vspace{-0.3em}flow}}} &
 	\includegraphics[width=.155\linewidth]{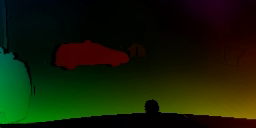} &
 	\includegraphics[width=.155\linewidth]{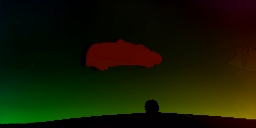} &
 	\includegraphics[width=.155\linewidth]{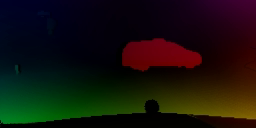} &
 	\includegraphics[width=.155\linewidth]{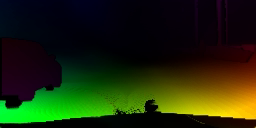} &
 	\includegraphics[width=.155\linewidth]{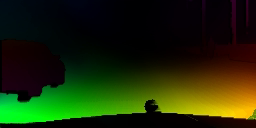} &
 	\includegraphics[width=.155\linewidth]{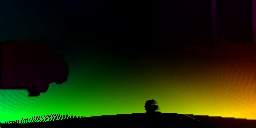}\\

	\end{tabular}
	\vspace{-0.5em}
	\caption{Visualization of the layered video decomposition. Semantic segmentation maps and optical flow maps are extracted from RGB frames using off-the-shelf methods~\cite{chen2018encoder, teed2020raft}.  We combine both to construct pseudo ground truths for object discovery: in {\setlength{\fboxsep}{1.5pt}\fbox{white}}, moving foreground regions towards which objects are attracted; in {\setlength{\fboxsep}{1.5pt}\colorbox{black!50}{\textcolor{white}{grey}}}, static foreground ones which remain neutral; and in {\setlength{\fboxsep}{1.5pt}\colorbox{black}{\textcolor{white}{black}}}, the background which repulses objects. We then show predicted object regions and their decomposition into layers. Each layer is tracked over time using a small set of control points. We compute motion vectors between points in pairs of frames (here, between consecutive time steps), and reconstruct from these and the layer masks the complex scene flow. Please zoom in for details.}
	\label{fig:lvd-1}
\end{figure}

\myparagraph{Layered video decomposition.} 
We illustrate in Figure~\ref{fig:lvd-1} our strategy to decompose videos into layers as a way to build inter-frame connections using a compact representation of motion from which we recover the dense scene flow.

We observe that objects are predicted in regions which match our pseudo ground truth, constructed from input segmentation and flow maps, even in difficult regions like the poles.
Although they share the same semantic class, the three cars in the left-most example in Figure~\ref{fig:lvd-1} are correctly segmented into different objects.
Still, it may happen that multiple objects are merged into the same layer, or that an object is over-segmented into multiple layers (like the ego vehicle in these examples).
This is due, in part, to the limitations of our approach which indicates regions of interest for the objects, but does not impose how they should be split among the different layers.
When computing video decompositions, we also position a set of control points associated with each layer.
The delta of control points between pairs of time steps produces sparse motion vectors for the background and the objects.
We show that, although we use a small number of points, we are able to accurately recover the dense scene flow using TPS transformations, and that motion discontinuities occur at layer boundaries as expected.

\begin{figure}[h]
	\setlength\tabcolsep{1.0pt}
	\renewcommand{\arraystretch}{1.0}
	\small
	\begin{tabular}{L{0.6cm}cccccc}
	& $T+1$ & $T+5$ & $T+10$ & $T+1$ & $T+5$ & $T+10$ \\

 	\raisebox{1.5\normalbaselineskip}[0pt][0pt]{\rotatebox[origin=c]{90}{\parbox{4cm}{\centering Real \\ \vspace{-0.3em}frame}}} &
 	\includegraphics[width=.155\linewidth]{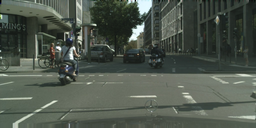} &
 	\includegraphics[width=.155\linewidth]{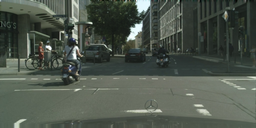} &
 	\includegraphics[width=.155\linewidth]{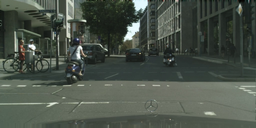} &
 	\includegraphics[width=.155\linewidth]{files/figures/sup/flp/vid_00009_5.png} &
 	\includegraphics[width=.155\linewidth]{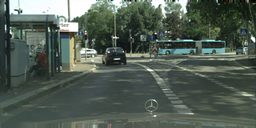} &
 	\includegraphics[width=.155\linewidth]{files/figures/sup/flp/vid_00009_14.png}\\
 	
 	\raisebox{1.5\normalbaselineskip}[0pt][0pt]{\rotatebox[origin=c]{90}{\parbox{4cm}{\centering Rec. \\ \vspace{-0.3em}motion}}} &
 	\includegraphics[width=.155\linewidth]{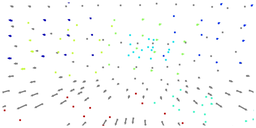} &
 	\includegraphics[width=.155\linewidth]{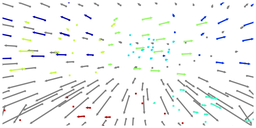} &
 	\includegraphics[width=.155\linewidth]{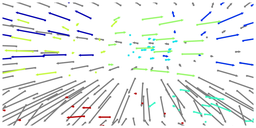} &
 	\includegraphics[width=.155\linewidth]{files/figures/sup/flp/rdm_00009_1.png} &
 	\includegraphics[width=.155\linewidth]{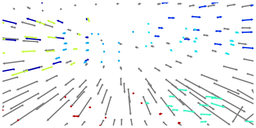} &
 	\includegraphics[width=.155\linewidth]{files/figures/sup/flp/rdm_00009_10.png}\\
 	
 	\raisebox{1.5\normalbaselineskip}[0pt][0pt]{\rotatebox[origin=c]{90}{\parbox{4cm}{\centering Pred. \\ \vspace{-0.3em}motion}}} &
 	\includegraphics[width=.155\linewidth]{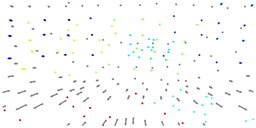} &
 	\includegraphics[width=.155\linewidth]{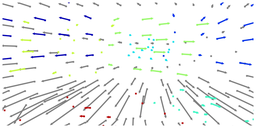} &
 	\includegraphics[width=.155\linewidth]{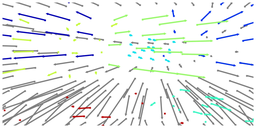} &
 	\includegraphics[width=.155\linewidth]{files/figures/sup/flp/pdm_00009_1.png} &
 	\includegraphics[width=.155\linewidth]{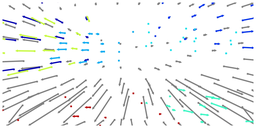} &
 	\includegraphics[width=.155\linewidth]{files/figures/sup/flp/pdm_00009_10.png}\\
 	
 	~ \\
 	
 	\raisebox{1.5\normalbaselineskip}[0pt][0pt]{\rotatebox[origin=c]{90}{\parbox{4cm}{\centering Real \\ \vspace{-0.3em}frame}}} &
 	\includegraphics[width=.155\linewidth]{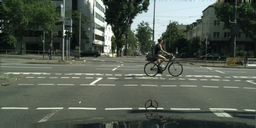} &
 	\includegraphics[width=.155\linewidth]{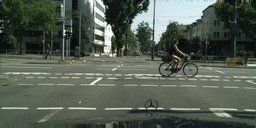} &
 	\includegraphics[width=.155\linewidth]{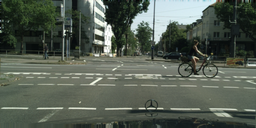} &
 	\includegraphics[width=.155\linewidth]{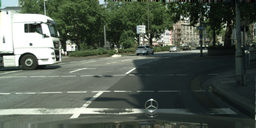} &
 	\includegraphics[width=.155\linewidth]{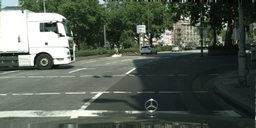} &
 	\includegraphics[width=.155\linewidth]{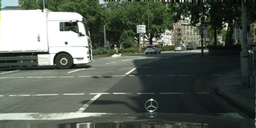}\\
 	
 	\raisebox{1.5\normalbaselineskip}[0pt][0pt]{\rotatebox[origin=c]{90}{\parbox{4cm}{\centering Rec. \\ \vspace{-0.3em}motion}}} &
 	\includegraphics[width=.155\linewidth]{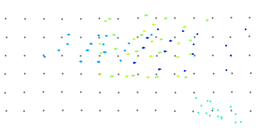} &
 	\includegraphics[width=.155\linewidth]{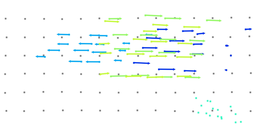} &
 	\includegraphics[width=.155\linewidth]{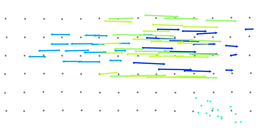} &
 	\includegraphics[width=.155\linewidth]{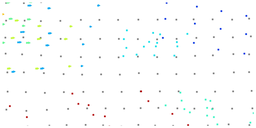} &
 	\includegraphics[width=.155\linewidth]{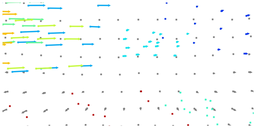} &
 	\includegraphics[width=.155\linewidth]{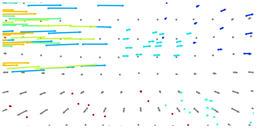}\\
 	
 	\raisebox{1.5\normalbaselineskip}[0pt][0pt]{\rotatebox[origin=c]{90}{\parbox{4cm}{\centering Pred. \\ \vspace{-0.3em}motion}}} &
 	\includegraphics[width=.155\linewidth]{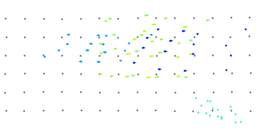} &
 	\includegraphics[width=.155\linewidth]{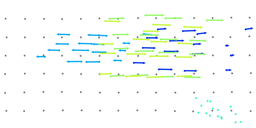} &
 	\includegraphics[width=.155\linewidth]{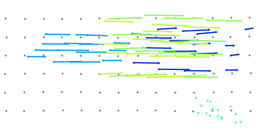} &
 	\includegraphics[width=.155\linewidth]{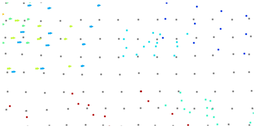} &
 	\includegraphics[width=.155\linewidth]{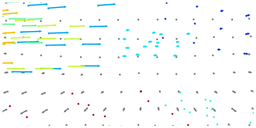} &
 	\includegraphics[width=.155\linewidth]{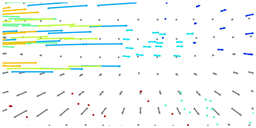}\\
 	
 	~ \\
 	
 	\raisebox{1.5\normalbaselineskip}[0pt][0pt]{\rotatebox[origin=c]{90}{\parbox{4cm}{\centering Real \\ \vspace{-0.3em}frame}}} &
 	\includegraphics[width=.155\linewidth]{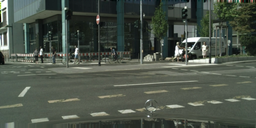} &
 	\includegraphics[width=.155\linewidth]{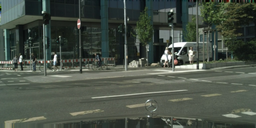} &
 	\includegraphics[width=.155\linewidth]{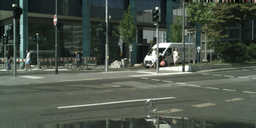} &
 	\includegraphics[width=.155\linewidth]{files/figures/sup/flp/vid_00074_5.png} &
 	\includegraphics[width=.155\linewidth]{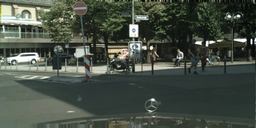} &
 	\includegraphics[width=.155\linewidth]{files/figures/sup/flp/vid_00074_14.png}\\
 	
 	\raisebox{1.5\normalbaselineskip}[0pt][0pt]{\rotatebox[origin=c]{90}{\parbox{4cm}{\centering Rec. \\ \vspace{-0.3em}motion}}} &
 	\includegraphics[width=.155\linewidth]{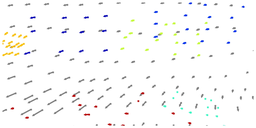} &
 	\includegraphics[width=.155\linewidth]{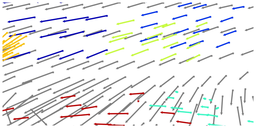} &
 	\includegraphics[width=.155\linewidth]{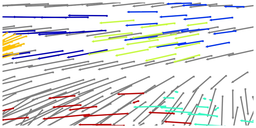} &
 	\includegraphics[width=.155\linewidth]{files/figures/sup/flp/rdm_00074_1.png} &
 	\includegraphics[width=.155\linewidth]{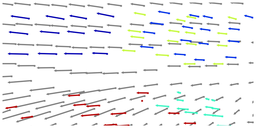} &
 	\includegraphics[width=.155\linewidth]{files/figures/sup/flp/rdm_00074_10.png}\\
 	
 	\raisebox{1.5\normalbaselineskip}[0pt][0pt]{\rotatebox[origin=c]{90}{\parbox{4cm}{\centering Pred. \\ \vspace{-0.3em}motion}}} &
 	\includegraphics[width=.155\linewidth]{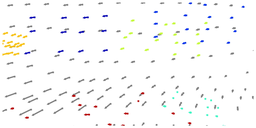} &
 	\includegraphics[width=.155\linewidth]{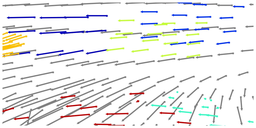} &
 	\includegraphics[width=.155\linewidth]{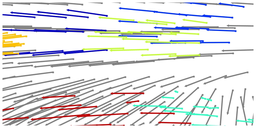} &
 	\includegraphics[width=.155\linewidth]{files/figures/sup/flp/pdm_00074_1.png} &
 	\includegraphics[width=.155\linewidth]{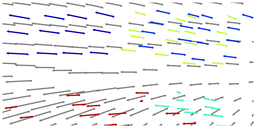} &
 	\includegraphics[width=.155\linewidth]{files/figures/sup/flp/pdm_00074_10.png}\\
    
	\end{tabular}
	\vspace{-0.5em}
	\caption{Visualization of future layer prediction. We use control points from the layered video decomposition as supervision. We compare motion vectors reconstructed from these points to the ones predicted for up to time step $T+10$ from a context of $T{=}4$ past frames. The motion vectors are computed between time step $T$ and time step $t$ in $\{T+1,T+10\}$. Different colors correspond to different layers. Please zoom in for details.}
	\label{fig:flp-1}
\end{figure}

\myparagraph{Future layer prediction.}
We compare in Figure~\ref{fig:flp-1} the scene dynamics extracted from our decomposition strategy to the one inferred via the future layer prediction module.

We accurately reconstruct complex motions under various scenarios: when the background is static, moves towards the camera due to the ego motion, or sideways when the car is turning; in the presence of different kinds of objects such as trucks, cars, bikes or various road elements; and whether these objects move in the same direction or not.

\begin{figure}[h]
	\setlength\tabcolsep{1.0pt}
	\renewcommand{\arraystretch}{1.0}
	\small
	\begin{tabular}{C{0.6cm}cccccc}
	& $T+1$ & $T+5$ & $T+10$ & $T+1$ & $T+5$ & $T+10$ \\

 	\raisebox{1.5\normalbaselineskip}[0pt][0pt]{\rotatebox[origin=c]{90}{Real frame}} &
 	\includegraphics[width=.155\linewidth]{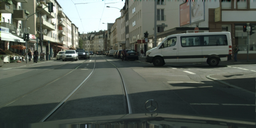} &
 	\includegraphics[width=.155\linewidth]{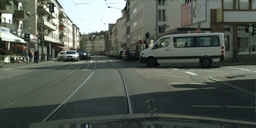} &
 	\includegraphics[width=.155\linewidth]{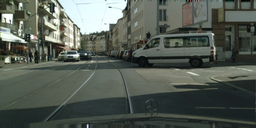} &
 	\includegraphics[width=.155\linewidth]{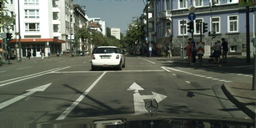} &
 	\includegraphics[width=.155\linewidth]{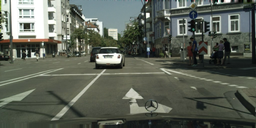} &
 	\includegraphics[width=.155\linewidth]{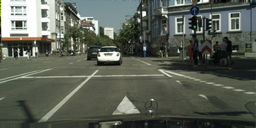}\\
 	
 	\raisebox{1.5\normalbaselineskip}[0pt][0pt]{\rotatebox[origin=c]{90}{\parbox{4cm}{\centering Warped \\ \vspace{-0.3em}frame}}} &
 	\includegraphics[width=.155\linewidth]{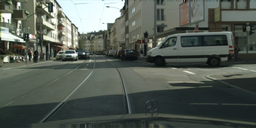} &
 	\includegraphics[width=.155\linewidth]{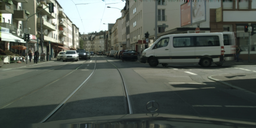} &
 	\includegraphics[width=.155\linewidth]{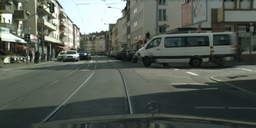} &
 	\includegraphics[width=.155\linewidth]{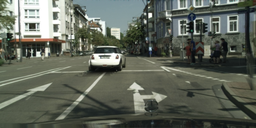} &
 	\includegraphics[width=.155\linewidth]{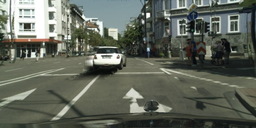} &
 	\includegraphics[width=.155\linewidth]{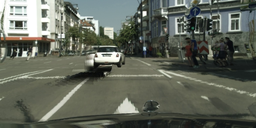}\\
 	
 	\raisebox{1.5\normalbaselineskip}[0pt][0pt]{\rotatebox[origin=c]{90}{Inp. frame}} &
 	\includegraphics[width=.155\linewidth]{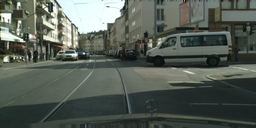} &
 	\includegraphics[width=.155\linewidth]{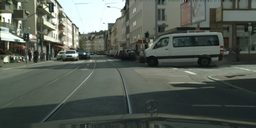} &
 	\includegraphics[width=.155\linewidth]{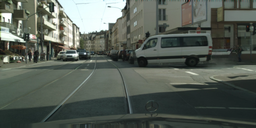} &
 	\includegraphics[width=.155\linewidth]{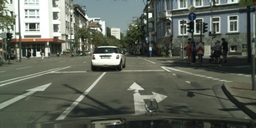} &
 	\includegraphics[width=.155\linewidth]{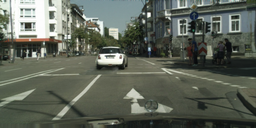} &
 	\includegraphics[width=.155\linewidth]{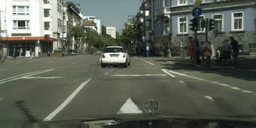}\\
 	
 	\raisebox{1.5\normalbaselineskip}[0pt][0pt]{\rotatebox[origin=c]{90}{Disocc.}} &
 	\includegraphics[width=.155\linewidth]{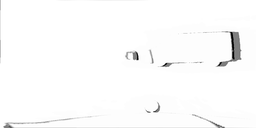} &
 	\includegraphics[width=.155\linewidth]{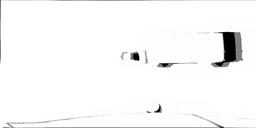} &
 	\includegraphics[width=.155\linewidth]{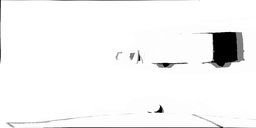} &
 	\includegraphics[width=.155\linewidth]{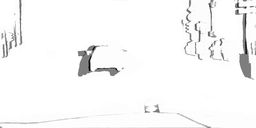} &
 	\includegraphics[width=.155\linewidth]{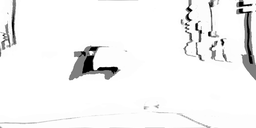} &
 	\includegraphics[width=.155\linewidth]{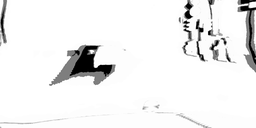}\\
 	
 	~\\
 	
 	\raisebox{1.5\normalbaselineskip}[0pt][0pt]{\rotatebox[origin=c]{90}{Real frame}} &
 	\includegraphics[width=.155\linewidth]{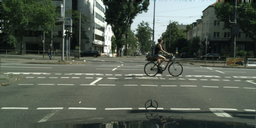} &
 	\includegraphics[width=.155\linewidth]{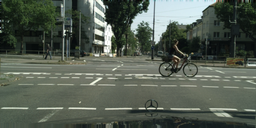} &
 	\includegraphics[width=.155\linewidth]{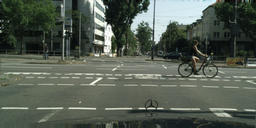} &
 	\includegraphics[width=.155\linewidth]{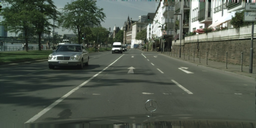} &
 	\includegraphics[width=.155\linewidth]{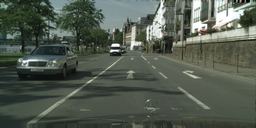} &
 	\includegraphics[width=.155\linewidth]{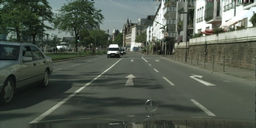}\\
 	
 	\raisebox{1.5\normalbaselineskip}[0pt][0pt]{\rotatebox[origin=c]{90}{\parbox{4cm}{\centering Warped \\ \vspace{-0.3em}frame}}} &
 	\includegraphics[width=.155\linewidth]{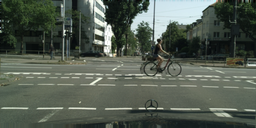} &
 	\includegraphics[width=.155\linewidth]{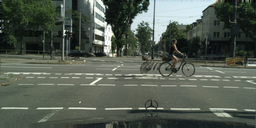} &
 	\includegraphics[width=.155\linewidth]{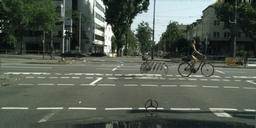} &
 	\includegraphics[width=.155\linewidth]{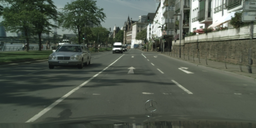} &
 	\includegraphics[width=.155\linewidth]{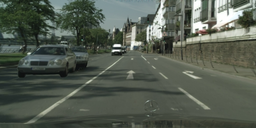} &
 	\includegraphics[width=.155\linewidth]{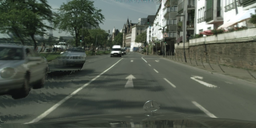}\\
 	
 	\raisebox{1.5\normalbaselineskip}[0pt][0pt]{\rotatebox[origin=c]{90}{Inp. frame}} &
 	\includegraphics[width=.155\linewidth]{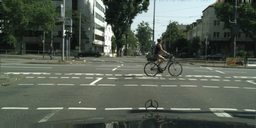} &
 	\includegraphics[width=.155\linewidth]{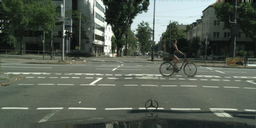} &
 	\includegraphics[width=.155\linewidth]{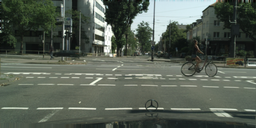} &
 	\includegraphics[width=.155\linewidth]{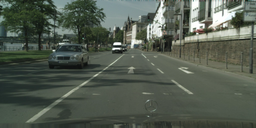} &
 	\includegraphics[width=.155\linewidth]{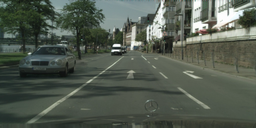} &
 	\includegraphics[width=.155\linewidth]{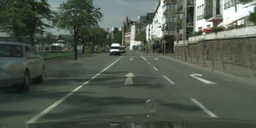}\\
 	
 	\raisebox{1.5\normalbaselineskip}[0pt][0pt]{\rotatebox[origin=c]{90}{Disocc.}} &
 	\includegraphics[width=.155\linewidth]{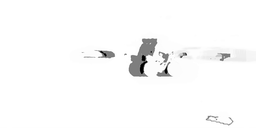} &
 	\includegraphics[width=.155\linewidth]{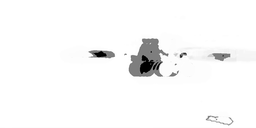} &
 	\includegraphics[width=.155\linewidth]{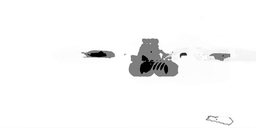} &
 	\includegraphics[width=.155\linewidth]{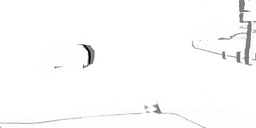} &
 	\includegraphics[width=.155\linewidth]{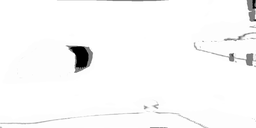} &
 	\includegraphics[width=.155\linewidth]{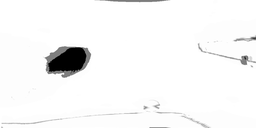}\\
    
	\end{tabular}
	\vspace{-0.5em}
	\caption{Visualization of the warping, fusion and inpainting module. We use the layer decomposition computed on a whole video to reconstruct the last $10$ frames using the $T=4$ first ones as context. This is done by warping, inpainting and fusing different views from the context. We compare real frames, warped ones, and fused/inpainted ones. We also illustrate the effect of disocclusion, by showing in the last row, in {\setlength{\fboxsep}{1.5pt}\colorbox{black}{\textcolor{white}{black}}}, regions which are not visible in any frame of the context, and, in {\setlength{\fboxsep}{1.5pt}\colorbox{black!50}{\textcolor{white}{grey}}}, those which are not visible in some frames but visible in others. Please zoom in for details.}
	\label{fig:wif-1}
\end{figure}

\myparagraph{Warping, fusion and inpainting.}
We illustrate in Figure~\ref{fig:wif-1} how future frames are finally synthesized.

Warping past frames to obtain future ones is not enough, as some regions may not be recovered from the past.
In particular, we see that shadows may not always be consistent with the new position of objects, \eg, for the vehicle in the bottom-right example in Figure~\ref{fig:wif-1}.
Our fusion and inpainting strategy is able to fill empty regions with realistic and temporal-consistent content (see also Sec~\ref{sec:inpabl}), and handles shadows reasonably well.
Finally, we show that by fusing multiple views from the past context, WALDO is able to reduce disocclusions significantly (from the {\setlength{\fboxsep}{1.5pt}\colorbox{black!50}{\textcolor{white}{grey}}} + {\setlength{\fboxsep}{1.5pt}\colorbox{black}{\textcolor{white}{black}}} to only {\setlength{\fboxsep}{1.5pt}\colorbox{black}{\textcolor{white}{black}}} regions in the last row of the figure).

\end{document}